\documentclass{article}

\PassOptionsToPackage{numbers,square,sort&compress}{natbib}

\usepackage[final]{neurips_2020}

\usepackage[utf8]{inputenc}
\usepackage[T1]{fontenc}
\usepackage[colorlinks=true,citecolor=black,urlcolor=black,linkcolor=black]{hyperref}
\usepackage{url}
\usepackage{booktabs}
\usepackage{amsfonts}
\usepackage{nicefrac}
\usepackage{microtype}

\usepackage{graphicx}
\usepackage{subcaption}

\renewcommand{\paragraph}[1]{{\bf #1}~~}

\usepackage{amsmath,amsfonts,bm}

\def\1{\bm{1}}

\def\vzero{{\bm{0}}}

\def\vtheta{{\bm{\theta}}}

\def\vg{{\bm{g}}}

\def\vm{{\bm{m}}}

\def\vs{{\bm{s}}}

\def\vx{{\bm{x}}}

\def\vz{{\bm{z}}}

\DeclareMathAlphabet{\mathsfit}{\encodingdefault}{\sfdefault}{m}{sl}
\SetMathAlphabet{\mathsfit}{bold}{\encodingdefault}{\sfdefault}{bx}{n}

\newcommand{\R}{\mathbb{R}}

\usepackage{url}
\usepackage{xspace}
\usepackage{amssymb}
\usepackage{wrapfig}
\usepackage[capitalize,nameinlink]{cleveref}
\usepackage{colortbl}

\usepackage{tikz,pgfplots}
\usetikzlibrary{plotmarks,shapes,shapes.arrows,positioning,calc,spy,intersections}
\pgfplotsset{compat=newest} 
\pgfkeys{/pgf/number format/.cd,1000 sep={}}

\newlength\figureheight
\newlength\figurewidth

\pgfplotsset{every axis/.append style={
  grid style={line width=0.6pt,dotted,gray}}}
  
\definecolor{mycolor0}{rgb}{0.2667,0.4471,0.7098}
\definecolor{mycolor1}{rgb}{0.1647,0.6706,0.3804}
\definecolor{mycolor2}{rgb}{0.8275,0.2627,0.3059}
\definecolor{mycolor3}{rgb}{0.5216,0.4392,0.7176}
\definecolor{mycolor4}{rgb}{0.8118,0.7255,0.4118}
\definecolor{mycolor5}{rgb}{0.2745,0.7176,0.8157}

\definecolor{mylcolor0}{rgb}{0.6902,0.7686,0.8863}
\definecolor{mylcolor1}{rgb}{0.5451,0.8902,0.6941}
\definecolor{mylcolor2}{rgb}{0.9412,0.7490,0.7647}
\definecolor{mylcolor3}{rgb}{0.8627,0.8392,0.9176}
\definecolor{mylcolor4}{rgb}{0.9569,0.9373,0.8667}
\definecolor{mylcolor5}{rgb}{0.7529,0.9020,0.9373}
\definecolor{mylcolor6}{rgb}{0.8750,0.8750,0.8750}

\newcommand{\eg}{\textit{e.g.}}
\newcommand{\ie}{\textit{i.e.}}

\newcommand{\etc}{\textit{etc.}}

\newcommand{\mathbold}[1]{\bm{#1}}
\newcommand{\mbf}[1]{\mathbf{#1}}
\newcommand{\vect}[1]{\mathbf{#1}}

\crefname{section}{Sec.}{Secs.}
\crefname{proposition}{Prop.}{Props.}
\crefname{lemma}{Lem.}{Lems.}
\crefname{model}{Mod.}{Mods.}
\crefname{appendix}{App.}{Apps.}

\newcommand{\valpha}[0]{\mathbold{\alpha}}

\newcommand{\vchi}[0]{\mathbold{\chi}}

\newcommand{\vxi}[0]{\mathbold{\xi}}
\newcommand{\vphi}[0]{\mathbold{\phi}}

\renewcommand{\mid}{\,|\,}

\newcommand{\MI}{\mbf{I}}

\newcommand{\CelebAHQ}{\textsc{CelebA-HQ}\xspace}

\newcommand{\LSUN}{\textsc{LSUN}\xspace}
\newcommand{\PIONEER}{\textsc{Pioneer}\xspace}

\newcommand{\todo}[1]{\textcolor{blue}{\textbf{[#1]}}}

\renewcommand{\cite}{\todo{DO NOT USE THE CITE COMMAND, use citet or citep!}}

\newcommand{\toptitlebar}{
  \hrule height 4pt
  \vskip 0.25in
  \vskip -\parskip%
}
\newcommand{\bottomtitlebar}{
  \vskip 0.29in
  \vskip -\parskip
  \hrule height 1pt
  \vskip 0.09in%
}
\newcommand{\nipstitle}[1]{{\phantomsection\hsize\textwidth\linewidth\hsize %
  \vskip 0.1in \toptitlebar\begin{minipage}{\textwidth}\centering{\LARGE\bf #1\par}\end{minipage}\bottomtitlebar%
  \addcontentsline{toc}{section}{#1}}}

\title{Deep Automodulators}

\author{
Ari Heljakka$^{1, 2}$ \qquad Yuxin Hou$^{1}$  \qquad Juho Kannala$^1$  \qquad Arno Solin$^1$ \\
$^1$Aalto University \qquad $^2$GenMind 
\\  {\small \texttt{\{ari.heljakka, yuxin.hou, juho.kannala, arno.solin\}@aalto.fi}}
}

\begin{document}

\maketitle

\begin{abstract}
  We introduce a new category of generative autoencoders called automodulators. These networks can faithfully reproduce individual real-world input images like regular autoencoders, but also generate a fused sample from an arbitrary combination of several such images, allowing instantaneous `style-mixing' and other new applications. An automodulator decouples the data flow of decoder operations from statistical properties thereof and uses the latent vector to modulate the former by the latter, with a principled approach for mutual disentanglement of decoder layers. Prior work has explored similar decoder architecture with GANs, but their focus has been on random sampling. A corresponding autoencoder could operate on real input images. For the first time, we show how to train such a general-purpose model with sharp outputs in high resolution, using novel training techniques, demonstrated on four image data sets. Besides style-mixing, we show state-of-the-art results in autoencoder comparison, and visual image quality nearly indistinguishable from state-of-the-art GANs. We expect the automodulator variants to become a useful building block for image applications and other data domains.
\end{abstract}

\section{Introduction}
\begin{figure*}
  \centering\scriptsize
  \resizebox{\textwidth}{!}{%
  \begin{tikzpicture}[outer sep=0]

    \newcommand{\insquare}[4]{\node [minimum width=1.5cm,minimum height=1.5cm, rounded corners=3pt,path picture={\node at (path picture bounding box.center){\includegraphics[width=1.5cm]{#3}};}] (#4) at (#1,#2) {};};

    \def\ida{31D}
    \def\id{31D}

    \insquare{-2.5}{1}{fig/sysmix/\ida/source-1.jpg}{in1}
    \insquare{-2.5}{-1}{fig/sysmix/\ida/source-4.jpg}{in2}

    \insquare{0}{1}{fig/sysmix/\ida/mix_1_1_0--1.jpg}{rec1}    
    \insquare{0}{-1}{fig/sysmix/\ida/mix_4_4_0--1.jpg}{rec2}

    \insquare{2.5}{1}{fig/sysmix/\ida/mix_4_1_2--1.jpg}{mod1}

    \insquare{2.5}{-1}{fig/sysmix/\id/rand-3.jpg}{samp}

    \insquare{5}{0}{fig/sysmix/\id/rmix_3_1_2-4.jpg}{mod2}

    \insquare{7.5}{1}{fig/sysmix/\id/r2mix_0_3_2-4.jpg}{csamp0}
    \insquare{7.5}{-1}{fig/sysmix/\id/r2mix_1_3_2-4.jpg}{csamp1}    
    \insquare{9.25}{1}{fig/sysmix/\id/r2mix_2_3_2-4.jpg}{csamp2}
    \insquare{9.25}{-1}{fig/sysmix/\id/r2mix_3_3_2-4.jpg}{csamp3}  
    \insquare{11.0}{1}{fig/sysmix/\id/r2mix_4_3_2-4.jpg}{csamp4}
    \insquare{11.0}{-1}{fig/sysmix/\id/r2mix_5_3_2-4.jpg}{csamp5}

    \tikzstyle{lab} = [text width=2cm, align=center]
    \node[lab,above of=in1] {\bf Input};
    \node[lab,below of=in2] {\bf Input};    
    \node[lab,below of=in1] {Real input images};      
    \node[lab,above of=rec1] {\bf Reconstruction};
    \node[lab,below of=rec1] {is modulated  by};
    \node[lab,below of=rec2] {\bf Reconstruction};    
    \node[lab,above of=mod1] {\bf Modulated output};
    \node[lab,below of=mod1] {is modulated  by};     
    \node[lab,below of=samp] {\bf Random sample};
    \node[lab,above of=mod2] {\bf Modulated output};    
    \node[lab,below of=mod2, yshift=-2mm] {`Coarse' features modulated};

    \node[lab,below of=csamp3, text width=6cm, yshift=-2mm] {Conditional samples given `coarse' and `fine' features from input};

    \tikzstyle{arr} = [->,black,thick]
    \draw[arr] (in1) to[out=0,in=180] (rec1);
    \draw[arr] (in2) to[out=0,in=180] (rec2);
    \draw[arr] (rec1) to[out=0,in=180] (mod1);
    \draw[arr] (rec2) to[out=0,in=180] (mod1); 
    \draw[arr] (mod1) to[out=0,in=180] (mod2);    
    \draw[arr] (samp) to[out=0,in=180] (mod2);
    \draw[arr] (mod2) to[out=0,in=180] (csamp0);
    \draw[arr] (mod2) to[out=0,in=180] (csamp1);
            
  \end{tikzpicture}}
  \caption{Illustration of some automodulator capabilities. The model can directly encode real (unseen) input images (left). Inputs can be mixed by modulating one with another or with a randomly drawn sample, at desired scales (center); \eg, `coarse' scales affect pose and gender \etc\ Finally, taking random modulations for certain scales produces novel samples conditioned on the input image (right).}
  \label{fig:teaser}

\end{figure*}
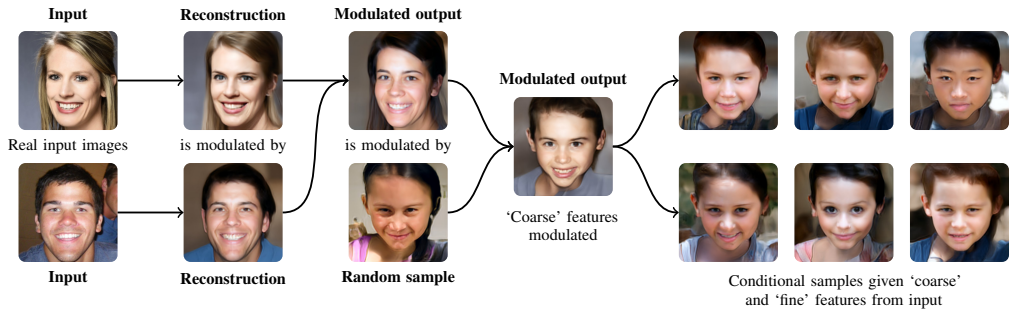

This paper introduces a new category of generative autoencoders for learning representations of image data sets, capable of not only reconstructing real-world input images, but also of arbitrarily combining their latent codes to generate fused images.
\cref{fig:teaser} illustrates the rationale: The same model can encode input images (far-left), mix their features (middle), generate novel ones (middle), and sample new variants of an image (conditional sampling, far-right). Without discriminator networks, training such an autoencoder for sharp high resolution images is challenging. For the first time, we show a way to achieve this.

Recently, impressive results have been achieved in random image generation (\eg, by GANs \citep[]{goodfellow2014,karras2018style,brock2018}). However, in order to manipulate a real input image, an `encoder' must first infer the correct representation of it. This means simultaneously requiring {sufficient} output image quality and the ability for reconstruction and feature extraction, which then allow semantic editing.
Deep generative autoencoders provide a principled approach for this. Building on the \PIONEER autoencoder \citep{heljakka2019towards}, we proceed to show that modulation of decoder layers by leveraging adaptive instance normalization (AdaIn,  \citep{UlyanovVL16,dumoulin2016learned,HuangB17}) further improves these capabilities. It also yields representations that are less entangled, a property here broadly defined as something that allows for fine and independent control of one semantic (image) sample attribute at a time. Here, the inductive bias is to assume each such attribute to only affect certain scales, allowing disentanglement \citep{locatello2018}. Unlike \citep{HuangB17}, previous GAN-based works on AdaIn \citep{karras2018style,chen2018self} have no built-in encoder for new input images.

In a typical autoencoder, input images are encoded into a latent space, and the information of the latent variables is then passed through successive layers of decoding until a reconstructed image has been formed. In our model, the latent vector independently {\em modulates} the statistics of each layer of the decoder so that  the output of layer $n$ is no longer solely determined by the input from layer $n-1$.

A key idea in our work is to reduce the mutual entanglement of decoder layers. For robustness, the samples once encoded and reconstructed by the autoencoder could be re-introduced to the encoder, repeating the process, and we could require consistency between the passes. In comparison to stochastic models such as VAEs \citep{kingma2014,rezende2014}, our deterministic model is better suited to take advantage of this. We can take the latent codes of two separate samples, drive certain layers (scales) of the decoder with one and the rest with the other, and then separately measure whether the information contained in each latent is conserved during the full decode--encode cycle. This enforces disentanglement of layer-specific properties, because we can ensure that the latent code introduced to affect only certain scales on the 1\textsuperscript{st} pass should not affect the other layers on the 2\textsuperscript{nd} pass, either.

In comparison to implicit (GAN) methods, regular image autoencoders such as VAEs tend to have poor output image quality.
In contrast, our model {\em {simultaneously}} balances sharp image outputs with the capability to encode and arbitrarily mix latent representations of real input images.

The contributions of this paper are as follows. 
  {\em (i)}~We provide techniques for stable {\em fully unsupervised} training of a high-resolution {\em automodulator}, a new form of an autoencoder with powerful properties not found in regular autoencoders, including scale-specific style transfer \citep{gatys2016image}. In contrast to architecturally similar `style'-based GANs, the automodulator can directly encode and manipulate new inputs.
  {\em (ii)}~We shift the way of thinking about autoencoders by presenting a novel disentanglement loss that further helps to learn more disentangled representations than regular autoencoders, a principled approach for incorporating scale-specific prior information in training, and a clean scale-specific approach to attribute modification.
  {\em (iii)}~We demonstrate promising qualitative and quantitative performance and applications on FFHQ, \CelebAHQ, and \LSUN Bedrooms and Cars data sets.

\section{Related Work}
\label{sec:related-work}

Our work builds upon several lines of previous work in unsupervised representation learning. The most relevant concepts are variational autoencoders (VAEs,  \citep{kingma2014,rezende2014}) and generative adversarial networks (GANs,  \citep{goodfellow2014}).
In VAEs, an encoder maps data points to a lower dimensional latent space and a decoder maps the latent representations back to the data space. The model is learnt by minimizing the reconstruction error, under a regularization term that encourages the distribution of latents to match a predefined prior. Latent representations often provide useful features for applications (\eg, image analysis and manipulation), and allow data synthesis by random sampling from the prior. However, with images, the samples are often blurry and not photorealistic, with imperfect reconstructions.

Current state-of-the-art in generative image modeling is represented by GAN models \citep{brock2018,karras2018style,2019arXiv191204958K} which achieve higher image quality than VAE-based models. Nevertheless, these GANs lack an encoder for obtaining the latent representation for a given image, limiting their usefulness. In some cases, a given image can be semantically mapped to the latent space via generator inversion but this iterative process is prohibitively slow for many applications (see comparison in \cref{app:comparison-to-gan-inv}), and the result may depend on initialization \citep{image2stylegan, creswell18}.

Bidirectional mapping has been targeted by VAE-GAN hybrids \citep{makhzani2015,larsen2015,mescheder2017,DBLP:journals/corr/abs-1805-09804}, and adversarial models \citep{Donahue2017,dumoulin2016D}. These models learn mappings between the data space and latent space using combinations of encoders, generators, and discriminators. However, even the latest state-of-the-art variant BigBiGAN \citep{DBLP:journals/corr/abs-1907-02544} focuses on random sampling and downstream classification performance, not on faithfulness of reconstructions. InfoGAN \citep{NIPS2016_6399,DBLP:journals/corr/abs-1906-06034} uses an encoder to constrain sampling but not for full reconstruction. IntroVAE \citep{huang2018} and Adversarial Generator Encoder (AGE, \citep{ulyanov2017}) {\em only comprise an encoder and a decoder}, adversarially related. \PIONEER scales AGE to high resolutions \citep{heljakka2018,heljakka2019towards}. VQ-VAE  \citep[]{NIPS2017_7210,VQVAE2} achieves high sample quality with a discrete latent space, but such space cannot, \eg, be interpolated, which hinders semantic image manipulation and prevents direct comparison.

Architecturally, our decoder and use of AdaIn are similar to the recent StyleGAN \citep{karras2018style} generator (without the `mapping network' $f$), but having a built-in encoder instead of the disposable discriminator leads to fundamental differences.
AdaIn-based skip connections are different from regular (non-modulating) 1-to-many skip connections from latent space to decoder layers, such as, \eg, 
in BEGAN \citep{berthelot2017,li2018}. Those skip connections have not been shown to allow `mixing' multiple latent codes, but merely to map the one and the same code to many layers, for the purpose of improving the reconstruction quality. Besides the AGE-based training \citep{ulyanov2017}, we can, \eg, also recirculate style-mixed reconstructions as `second-pass' inputs to further encourage the independence and disentanglement of emerging styles and conservation of layer-specific information. The biologically motivated recirculation idea is conceptually related to many works, going back to at least 1988 \citep{hinton1988learning}. Utilizing the outputs of the model as inputs for the next iteration has been shown to benefit, \eg, image classification \citep{zamir2017}, and is used extensively in RNN-based methods \citep{rezende2016one,gregor2015draw,gregor2016towards}.

\section{Methods}
\label{sec:methods}
We begin with the primary underlying techniques used to construct the automodulator: the progressive growing of the architecture necessary for high-resolution images and the AGE-like adversarial training as combined in the \PIONEER \citep{heljakka2018,heljakka2019towards}, but now with an architecturally different decoder to enable `modulation' by AdaIn \citep{UlyanovVL16,dumoulin2016learned,HuangB17,karras2018style} (\cref{sec:architecture}). The statistics modulation allows for multiple latent vectors to contribute to the output, which we leverage for an improved unsupervised loss function in \cref{sec:strong-conservation}. We then introduce an optional method for weakly supervised training setup, applicable when there are known scale-specific invariances in the training data itself \cref{sec:invariances}.

\subsection{Automodulator Components}
\label{sec:architecture}

Our overall scheme starts from unsupervised training of a symmetric convolution--deconvolution autoencoder-like model. Input images $\vx$ are fed through an encoder $\vphi$ to form a low-dimensional latent space representation $\vz$ (we use $\vz \in \R^{512}$, normalized to unity). This representation can then be decoded back into an image $\hat{\vx}$ through a decoder $\vtheta$.

\paragraph{Adversarial generator encoder loss}
To utilize adversarial training, the automodulator training builds upon AGE and \PIONEER. The encoder $\vphi$ and the decoder $\theta$ are trained on separate steps, where $\vphi$ attempts to push the latent codes of training images towards a unit Gaussian distribution $ \mathrm{N}(\vect{0},\MI)$, and the codes of random generated images away from it. $\vtheta$ attempts to produce random samples with the opposite goal. In consecutive steps, one optimizes  loss $\mathcal{L}_\phi$ and $\mathcal{L}_\theta$ \citep{ulyanov2017}, with margin $M_\mathrm{gap}$ for $\mathcal{L}_\phi$ \citep{heljakka2019towards} (negative KL term of $\mathcal{L}_\theta$ dropped, as customary \citep{ulyanovGithub,heljakka2018}), defined as
\hspace*{0.05cm}\vbox{%
\begin{align}
  \hspace*{-.2cm}
  \mathcal{L}_\phi &{=} \max(-M_\mathrm{gap}, \mathrm{D_{KL}}[q_\phi(\vz\mid\vx) \,\|\, \mathrm{N}(\vect{0},\MI)]    {-} \mathrm{D_{KL}}[q_\phi(\vz\mid\hat{\vx}) \,\|\, \mathrm{N}(\vect{0},\MI)]) {+} \lambda_\mathcal{X} \, d_\mathcal{X}(\vx, \vtheta(\vphi(\vx))), \label{eq:loss-orig-phi} \\
  \hspace*{-.2cm}
  \mathcal{L}_\theta &{=} \mathrm{D_{KL}}[q_\phi(\vz\mid\hat{\vx}) \,\|\, \mathrm{N}(\vect{0},\MI)] {+} \lambda_\mathcal{Z}\,d_{\mathrm{cos}}(\vz, \vphi(\vtheta(\vz))), \label{eq:loss-orig-theta-phi}
\end{align}}
where
 ${\vx}$ is sampled from the training set, $\hat{\vx} \sim q_\theta(\vx \mid \vz)$, $\vz \sim \mathrm{N}(\vect{0},\MI)$, $d_\mathcal{X}$ is L1 or L2 distance, and $d_{\mathrm{cos}}$ is the cosine distance.
 The KL divergence can be calculated from empirical distributions of $q_\phi(\vz\mid\hat{\vx})$ and $q_\phi(\vz\mid{\vx})$. Still, the model inference is deterministic, so we could retain, in principle, the full information contained in the image, at every stage of the processing.
For any latent vector $\vz$, decoded back to image space as $\hat{\vx}$, and re-encoded as a latent $\vz'$, it is possible and desirable to require that $\vz$ is as close to $\vz'$ as possible, yielding the latent reconstruction error $d_{\mathrm{cos}}(\vz, \vphi(\vtheta(\vz)))$. We will generalize this term in \ref{sec:strong-conservation}.
  
\paragraph{Progressively growing autoencoder architecture}
To make the AGE-like training stable in high resolution, we build up the architecture and increase image resolution progressively during training, starting from tiny images and gradually growing them, making the learning task harder (see \citep{karras2017,heljakka2018} and Supplement \cref{fig:architecture}).
The convolutional layers of the symmetric encoder and decoder are faded in gradually during the training, in tandem with the resolution of training images and generated images (\cref{fig:architecture}).

\begin{figure*}[!t]
  \centering

    \tikzstyle{block} = [rounded corners=0.5pt,minimum width=1.5mm,minimum height=1.5mm,inner sep=0,draw=black!50!black,fill=white]

    \newcommand{\encoder}[4]{%
      \foreach \j [count=\i] in {4,8,16,32,64,128,256} 
        \node[block,minimum width={0.25cm+0.2cm*\i}] (#4-\i) at (#1,#2-.3*\i) {\scalebox{.5}{$\j{\times}\j$}};
    }   
  \begin{subfigure}[b]{.28\textwidth}
    \centering
    \scalebox{0.8}{%
    \begin{tikzpicture}[inner sep=0]

    \node [minimum width=3mm,minimum height=3mm] (zA) at (1.5,.25) {$\vz$};

    \node [minimum width=3mm] (xi0) at (3,.25) {$\vxi^{(0)}$};

    \node at (0,-3.2) (in) {\includegraphics[width=1.5cm]{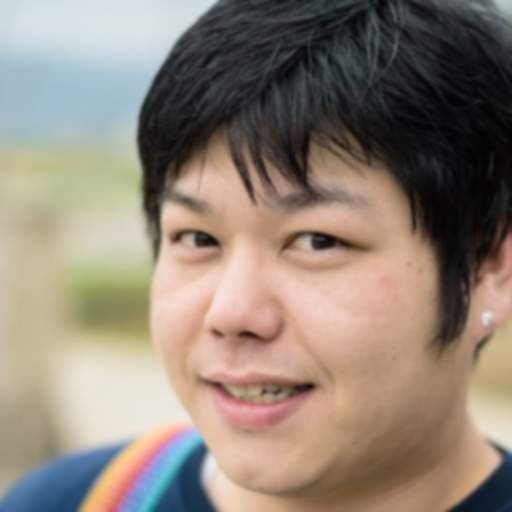}};
    \node[minimum width=7mm,minimum height=3.5mm,fill=white,opacity=1,draw=black,rounded corners=0.5pt] at (0,-3.2-.75) {};
    \node at (0,-3.2-.75) {$\vphantom{\hat{\vx}}\vx$};

    \node (out) at (3,-3.2) {\includegraphics[width=1.5cm]{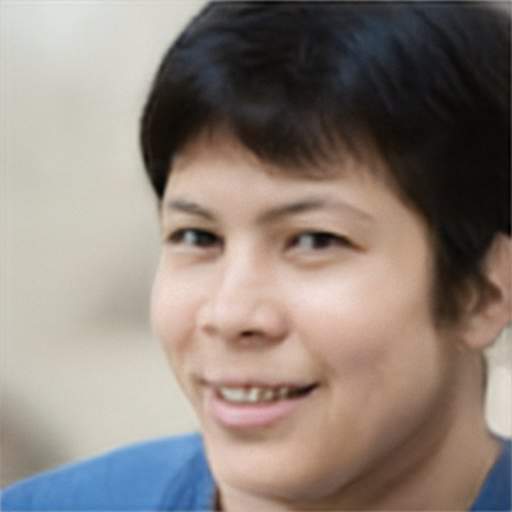}};
    \node[minimum width=7mm,minimum height=3.5mm,fill=white,opacity=1,draw=black,rounded corners=0.5pt] at (3,-3.2-.75) {};
    \node at (3,-3.2-.75) {$\hat{\vx}$};

    \draw [->, rounded corners=3mm, thick] (in) |- (zA);
    \draw [->, rounded corners=3mm, thick] (xi0) -- (out);

    \encoder{0}{0}{4}{enc}
    \encoder{3}{0}{4}{dec}

    \draw [->, rounded corners=3mm, thick] (zA) |- (dec-1);
    \draw [->, rounded corners=3mm, thick] (zA) |- (dec-2);
    \draw [->, rounded corners=3mm, thick] (zA) |- (dec-3);
    \draw [->, rounded corners=3mm, thick] (zA) |- (dec-4);
    \draw [->, rounded corners=3mm, thick] (zA) |- (dec-5);
    \draw [->, rounded corners=3mm, thick] (zA) |- (dec-6);
    \draw [->, rounded corners=3mm, thick] (zA) |- (dec-7);

    \node[rotate=90] at (-.75,-.5) {\scriptsize Encoder, $\vphi$};
    \node[rotate=-90] at (3.75,-.5) {\scriptsize Decoder, $\vtheta$};
    \node[above of=zA,yshift=-3mm,text width=1cm, align=center] {\scriptsize Latent\\[-3pt] encoding};        
    \node[above of=xi0,yshift=-3mm,text width=1cm, align=center] {\scriptsize Canvas\\[-3pt] variable};

    \foreach \i in {1,2,3,4,5,6,7} 
      \node[block,minimum width={0.2},shape=circle,draw=black,fill=white] at (1.9,-0.3*\i) {};            
            
    \end{tikzpicture}}
    \caption{Architecture}   
    \label[figure]{fig:arch}     
  \end{subfigure}
  \hspace{\fill}
  \begin{subfigure}[b]{.28\textwidth}
    \centering  
    \scalebox{0.8}{%
    \begin{tikzpicture}[inner sep=0]

    \node [minimum width=3mm,minimum height=3mm] (zA) at (1.5,.25) {$\vz_A$};
    \node [minimum width=3mm,minimum height=3mm] (zB) at (1.5,-2.75) {$\vz_B$};

    \node [minimum width=3mm] (xi0) at (3,.25) {$\vxi^{(0)}$};

    \node at (0,-3.2) (in) {\includegraphics[width=1.5cm]{./fig/sysmix8/42420N2/source-5.jpg}};
    \node[minimum width=7mm,minimum height=3.5mm,fill=white,opacity=1,draw=black,rounded corners=0.5pt] at (0,-3.2-.75) {};
    \node at (0,-3.2-.75) {$\vphantom{\hat{\vx}_{AB}}\vx_A$};

    \node (out) at (3,-3.2) {\includegraphics[width=1.5cm]{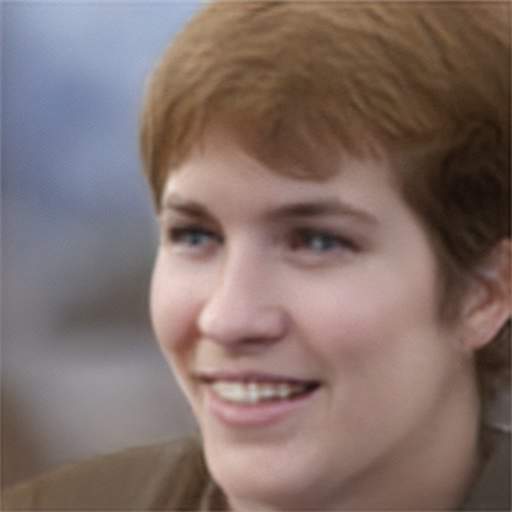}};
    \node[minimum width=7mm,minimum height=3.5mm,fill=white,opacity=1,draw=black,rounded corners=0.5pt] at (3,-3.2-.75) {};
    \node at (3,-3.2-.75) {$\hat{\vx}_{AB}$};

    \draw [->, rounded corners=3mm, thick] (in) |- (zA);
    \draw [->, rounded corners=3mm, thick] (xi0) -- (out);

    \encoder{0}{0}{4}{enc}
    \encoder{3}{0}{4}{dec}

    \draw [->, rounded corners=3mm, thick] (zA) |- (dec-1);
    \draw [->, rounded corners=3mm, thick] (zA) |- (dec-2);
    \draw [->, rounded corners=3mm, thick] (zB) |- (dec-3);
    \draw [->, rounded corners=3mm, thick] (zB) |- (dec-4);
    \draw [->, rounded corners=3mm, thick] (zB) |- (dec-5);
    \draw [->, rounded corners=3mm, thick] (zB) |- (dec-6);
    \draw [->, rounded corners=3mm, thick] (zB) |- (dec-7);

    \node[rotate=90] at (-.75,-.5) {\scriptsize Encoder, $\vphi$};
    \node[rotate=-90] at (3.75,-.5) {\scriptsize Decoder, $\vtheta$};

    \foreach \i in {1,2,3,4,5,6,7} 
      \node[block,minimum width={0.2},shape=circle,draw=black,fill=white] at (1.9,-0.3*\i) {};            
            
    \end{tikzpicture}}
    \caption{1-pass flow}   
    \label{fig:1-pass}     
  \end{subfigure}
  \hspace{\fill}
  \begin{subfigure}[b]{.28\textwidth}
    \centering  
    \hspace*{-1cm}
    \scalebox{0.8}{%
    \begin{tikzpicture}[inner sep=0, spy using outlines={circle}]

    \node [minimum width=3mm,minimum height=3mm] (zA) at (1.5,.25) {$\vz_A$};
    \node [minimum width=3mm,minimum height=3mm] (zB) at (1.5,-2.75) {$\vz_B$};
    \node [minimum width=3mm,minimum height=3mm] (zAB) at (2.0,.75) {\textcolor{blue!50!black}{$\hat{\vz}_{AB}$}};

    \node [minimum width=3mm] (xi0) at (3,.25) {$\vxi^{(0)}$};

    \node[opacity=.5] at (0,-3.2) (in) {\includegraphics[width=1.5cm]{./fig/sysmix8/42420N2/source-5.jpg}};
    \node[minimum width=7mm,minimum height=3.5mm,fill=white,opacity=1,draw=black,rounded corners=0.5pt] at (0,-3.2-.75) {};
    \node[opacity=.5] at (0,-3.2-.75) {$\vphantom{\hat{\vx}_{AB}}\vx_A$};

    \coordinate (in2) at (.5,-2.7);

    \node (out) at (3,-3.2) {\includegraphics[width=1.5cm]{./fig/sysmix8/42420N2/mix-2_5_0-2.jpg}};
    \node[minimum width=7mm,minimum height=3.5mm,fill=white,opacity=1,draw=black,rounded corners=0.5pt] at (3,-3.2-.75) {};
    \node at (3,-3.2-.75) {$\hat{\vx}_{AB}$};

    \draw [->, rounded corners=3mm, thick, opacity=.5] (in) |- (zA);
    \draw [rounded corners=3mm, thick] (xi0) -- (dec-1);
    \draw [->, rounded corners=3mm, thick] (dec-3) -- (out);
    
    \draw [rounded corners=3mm, thick, black] ([xshift=-0.5mm]dec-1.center) -- ([xshift=-0.5mm]dec-3.center);
    \draw [rounded corners=3mm, double, blue!50!black] ([xshift=0.5mm]dec-1.center) -- ([xshift=0.5mm]dec-3.center);    

    \encoder{3}{0}{4}{dec}

    \tikzstyle{block} = [rounded corners=0.5pt,minimum width=1.5mm,minimum height=1.5mm,inner sep=0,draw=black!50!black,fill=white,opacity=.5]

    \encoder{0}{0}{4}{enc}

    \draw [->, rounded corners=3mm, thick] (zA) |- (dec-1);
    \draw [->, rounded corners=3mm, thick] (zA) |- (dec-2);
    \draw [->, rounded corners=3mm, thick] (zB) |- (dec-3);
    \draw [->, rounded corners=3mm, thick] (zB) |- (dec-4);
    \draw [->, rounded corners=3mm, thick] (zB) |- (dec-5);
    \draw [->, rounded corners=3mm, thick] (zB) |- (dec-6);
    \draw [->, rounded corners=3mm, thick] (zB) |- (dec-7);

    \node[rotate=90,opacity=.5] at (-.75,-.5) {\scriptsize Encoder, $\vphi$};
    \node[rotate=-90] at (3.75,-.5) {\scriptsize Decoder, $\vtheta$};

    \foreach \i in {1,2,3,4,5,6,7} 
      \node[block,minimum width={0.2},shape=circle,draw=black,fill=white,opacity=1] at (1.9,-0.3*\i) {};

    \draw [->, rounded corners=3mm, double, blue!50!black] (zAB) |- (dec-1);
    \draw [->, rounded corners=3mm, double, blue!50!black] (zAB) |- (dec-2);

    \draw [rounded corners=3mm, double, blue!50!black] (out) -| (in2);
    \draw [->, rounded corners=3mm, double, blue!50!black] (in2) |- (zAB);

    \spy [circle, red!50!black, draw, height=2cm, width=2cm, magnification=2.5,
      connect spies] on (3,-.75) in node [fill=white, opacity=.75] at (-1.1,-.75);  
    \node at (3.01,-.75) {\scalebox{.23}{$\vxi^{(2)}_A ~~\sim~~ \textcolor{blue!50!black}{\vxi^{(2)}_{AB}}$}};  
    
    \end{tikzpicture}}
    \caption{2-pass flow}    
    \label{fig:2-pass}
  \end{subfigure}
  \caption{(a)~The autoencoder-like usage of the model. (b)~Modulations in the decoder can come from different latent vectors. This can be leveraged in feature/style mixing, conditional sampling, and during the model training (first pass). (c)~The second pass during training, yielding $\mathcal{L}_j$.}
  \label{fig:model} 

\end{figure*}

\paragraph{Automodulation}
To build a separate pathway for modulation of decoder layer statistics, we need to integrate the AdaIn operation for each layer (following \citep{karras2018style}). In order to generate an image, a traditional image decoder would start by mapping the latent code to the first deconvolutional layer to form a small-resolution image ($\vtheta_0(\vz)$) and expand the image layer by layer ($\vtheta_1(\vtheta_0(\vz))$ \etc) until the full image is formed. In contrast, our decoder is composed of layer-wise functions $\vtheta_i(\vxi^{(i-1)}, \vz)$ that separately take a `canvas' variable $\vxi^{(i-1)}$ denoting the output of the preceding layer (see \cref{fig:architecture,fig:arch}), and the actual (shared) latent code $\vz$. First, for each feature map ~\#$j$ of the deconvolutional layer~\#$i$, we compute the activations ~$\vchi_{ij}$ from $\vxi^{(i-1)}$ as in traditional decoders.
But now, we modulate (\ie, re-scale) ~$\vchi_{ij}$ into having a new mean $m_{ij}$ and standard deviation $s_{ij}$, based on $\vz$ (\eg, a block of four layers with 16 channels uses $4{\times}16{\times}2$ scalars). 
To do this, we need to learn a mapping $\vg_i: \vz \mapsto (\vm_i, \vs_i)$.
We arrive at the AdaIn normalization (also see \cref{sec:archdetail}):
\begin{equation}\label{eq:adain}
  \mathrm{AdaIn}(\vchi_{ij}, \vg_i(\vz)) = s_{ij} \left( \frac{\vchi_{ij}-\mu(\vchi_{ij})}{\sigma(\vchi_{ij})} \right) + m_{ij}.
\end{equation}
We implement $\vg_i$ as a fully connected linear layer (in $\vtheta$), with output size $2\,C_i$ for $C_i$ channels. 
 Layer~\#$1$ starts from a constant input $\vxi^{(0)} \in \R^{4{\times}4}$. Without loss of generality, here we focus on pyramidal decoders with monotonically increasing resolution and decreasing number of channels.

\subsection{Conserving Scale-specific Information Over Cycles}
\label{sec:strong-conservation}
We now proceed to generalize the reconstruction losses in a way that specifically benefits from the automodulator architecture. We encourage the latent space to become hierarchically disentangled with respect to the levels of image detail, allowing one to separately retrieve `coarse' vs.\ `fine' aspects of a latent code. This enables, \eg, conditional sampling by fixing the latent code at specific decoder layers, or mixing the scale-specific features of multiple input images---impossible feats for a traditional autoencoder with mutually entangled decoder layers.

First, reinterpret the latent reconstruction error $d_{\mathrm{cos}}(\vz, \vphi(\vtheta(\vz)))$ in \cref{eq:loss-orig-theta-phi} as `reconstruction at decoder layer \#0'. One can then trivially generalize it to any layer \#$i$ of $\vtheta$ by measuring differences in $\vxi^{(i)}$, instead. We simply pick a layer of measurement, record $\vxi^{(i)}_1$, pass the sample through a full encoder--decoder cycle, and compare the new $\vxi^{(i)}_2$. But now, in the automodulator, different latent codes can be introduced on a per-layer basis, enabling us to measure how much information about a {\em specific} latent code is conserved at a specific layer, after one more full cycle. Without loss of generality, here we only consider mixtures of two codes. We can present the output of a decoder (\cref{fig:1-pass}) with $N$ layers, split after the $j$\textsuperscript{th} one, as a composition $\hat{\vx}_{AB} = \vtheta_{j+1:N}(\vtheta_{1:j}(\vxi^{(0)}, \vz_A), \vz_B)$. Crucially, we can choose $\vz_A \neq \vz_B$ (extending the method of \citep{karras2018style}), such as $\vz_A = \vphi(\vx_A)$ and $\vz_B = \vphi(\vx_B)$ for (image) inputs $\vx_A \neq \vx_B$. Because the earlier layers~\#$1{:}j$ operate on image content at lower (`coarse') resolutions, the fusion image $\hat{\vx}_{AB}$ has the `coarse' features of $\vz_A$ and the `fine' features of $\vz_B$.
Now, any $\vz$ holds feature information at different levels of detail, some empirically known to be mutually independent (\eg, skin color and pose), and we want them separately retrievable, \ie, to keep them `disentangled' in $\vz$. Hence, when we {\textit {re-encode}} $\hat{\vx}_{AB}$ into $\hat{\vz}_{AB} = \vphi(\hat{\vx}_{AB})$, then $\vtheta_{1:j}(\vxi^{(0)}, \hat{\vz}_{AB})$ should extract the same output as $\vtheta_{1:j}(\vxi^{(0)}, {\vz}_{A})$, unaffected by $\vz_B$.

This motivates us to minimize the layer disentanglement loss
\begin{equation}\label{eq:Lj}
  \mathcal{L}_j = d(\vtheta_{1:j}(\mathrm{\vxi^{(0)}}, \hat{\vz}_{AB}), \vtheta_{1:j}(\mathrm{\vxi^{(0)}}, \vz_{A}))
\end{equation}
for some distance function $d$ (here, L2 norm), with $\vz_{A}, \vz_{B} \sim \mathrm{N}(\vect{0},\MI)$, for each $j$.
In other words, the fusion image can be encoded into a new latent vector
\begin{equation}\label{eq:zAB}
\hat{\vz}_{AB} \sim q_{\phi}(\vz \mid \vx) \, q_{\theta_{j+1:N}}(\vx \mid \vxi^{(j)}, \vz_B) \, q_{\theta_{1:j}}(\vxi^{(j)} \mid \vxi^{(0)}, \vz_A),
\end{equation}
in such a way that, at each layer, the decoder will treat the new code similarly to whichever of the original two separate latent codes was originally used there (see \cref{fig:2-pass}). For a perfect network, $\mathcal{L}_j$ can be viewed as a `layer entanglement error'. Randomizing $j$ during the training, we can measure $\mathcal{L}_j$ for any layers of the decoder. A similar loss for the later stage $\vtheta_{j:N}(\vxi^{(j)}, \vz_B)$ is also possible, but due to more compounded noise and computation cost (longer cycle), was omitted for now.

\begin{wrapfigure}[28]{r}{0.4\textwidth}
  \centering
  \vspace*{-1em}

    \tikzstyle{block} = [rounded corners=0.5pt,minimum width=1.5mm,minimum height=1.5mm,inner sep=0,draw=black!50!black,fill=white]

          \newcommand{\encoder}[4]{%
          \foreach \j [count=\i] in {{\vphi}, {\vtheta}} 
            \node[block,minimum width={0.25cm+0.4cm},minimum height=0.5cm] (#4-\i) at (#1*\i-3.5,#2) {\scalebox{.8}{$\j$}};
        }   
        \newcommand{\encodert}[4]{%
        \foreach \j [count=\i] in {{\vphi}, {\vtheta}} 
          \node[block,minimum width={0.8cm},minimum height=0.5cm] (#4-\i) at (#1*\i-2.0,#2) {\scalebox{.8}{$\j$}};
      }   
      \newcommand{\recoh}[4]{%
        \draw [thick] (#1-0.25,#2-0.25) -- (#1-0.25,#2+0.25);
        \draw [thick] (#1+0.25,#2-0.25) -- (#1+0.25,#2+0.25);
        \draw [->,thick] (#1-0.25,#2-0.2) -- (#1-0.25+0.15,#2-0.2);
        \draw [->,thick] (#1+0.25,#2+0.2) -- (#1+0.25-0.15,#2+0.2);
        }
        \newcommand{\recov}[4]{%
        \draw [thick] (#1-0.25,#2-0.25) -- (#1+0.25,#2-0.25);
        \draw [thick] (#1-0.25,#2+0.25) -- (#1+0.25,#2+0.25);
        \draw [->,thick] (#1-0.2,#2-0.25) -- (#1-0.2,#2-0.25+0.15);
        \draw [->,thick] (#1+0.2,#2+0.25) -- (#1+0.2,#2+0.25-0.15);
        }
        \newcommand{\recovc}[4]{%
        \draw [thick] (#1-0.25,#2-0.25) -- (#1+0.25,#2-0.25);
        \draw [thick] (#1-0.25,#2+0.25) -- (#1+0.25,#2+0.25);
        \draw [->,thick] (#1-0.2,#2-0.25) -- (#1-0.2,#2-0.25-0.15);
        \draw [->,thick] (#1+0.2,#2+0.25) -- (#1+0.2,#2+0.25+0.15);
        }

  \begin{subfigure}[b]{.4\textwidth}
    \centering

    \tikzstyle{block} = [rounded corners=0.5pt,minimum width=1.5mm,minimum height=1.5mm,inner sep=0,draw=black!50!black,fill=white]
          
    \scalebox{0.75}{%
    \begin{tikzpicture}[inner sep=0]

    \node at (1.75-2.75,-3.2) (xin) {\includegraphics[width=1.5cm]{./fig/sysmix8/42420N2/source-5.jpg}};
    \node at (1.75+2.75,-3.2) (xout) {\includegraphics[width=1.5cm]{./fig/sysmix8/42420N2/mix-5_5_0--1.jpg}};

    \node at (1.75,-3.2) (in) {\includegraphics[width=1.5cm]{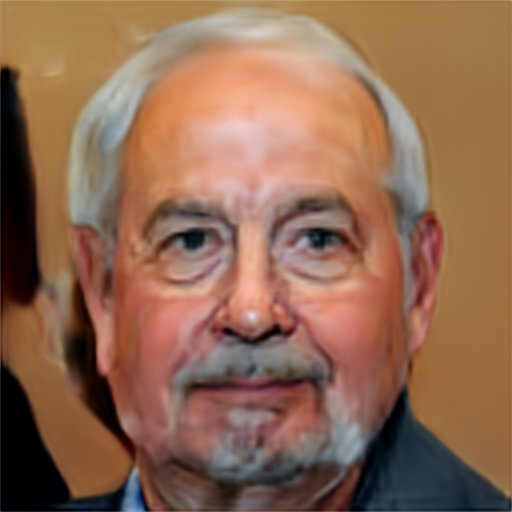}};

    \node [minimum width=3mm,minimum height=3mm] at (1.3,-1.6) (zhat) {$\hat{\vz}$};
    \node [minimum width=3mm,minimum height=3mm] at (2.2,-1.6) (z) {$\vz$};
    \node [minimum width=3mm,minimum height=3mm] at (1.3,1.8) (z2) {$\vz$};

    \node[minimum width=7mm,minimum height=6mm,fill=white,opacity=1,draw=black,rounded corners=0.1pt] at (1.3,0.2) (norm1box) {};
    \node[minimum width=7mm,minimum height=6mm,fill=white,opacity=1,draw=black,rounded corners=0.1pt] at (2.2,0.2) (norm2box) {};
    \node at (1.3,0.2) (norm-1) {\includegraphics[width=0.6cm]{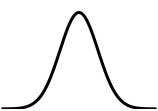}};
    \node at (2.2,0.2) (norm-2) {\includegraphics[width=0.6cm]{./fig/architecture2/norm.jpg}};
    \node[minimum width=0.6cm,fill=white,opacity=0,draw=black,rounded corners=0.5pt] at (2.8,0.0) (norm2bottom) {};

    \node[minimum width=0.65cm,fill=white,opacity=0,draw=black,rounded corners=0.5pt] at (3.5,-1.9) (thetatop) {};
    \node[minimum width=0.65cm,fill=white,opacity=0,draw=black,rounded corners=0.5pt] at (3.5,-2.0) (thetamid) {};
    \node[minimum width=0.65cm,fill=white,opacity=0,draw=black,rounded corners=0.5pt] at (3.5,-2.1) (thetabottom) {};

    \node[minimum width=7mm,minimum height=3.5mm,fill=white,opacity=1,draw=black,rounded corners=0.5pt] at (1.75,-3.2-.75) {};
    \node at (1.75,-3.2-.75) {$\hat{\vx}$};

    \node[minimum width=7mm,minimum height=3.5mm,fill=white,opacity=1,draw=black,rounded corners=0.5pt] at (-1,-3.2-.75) (xt) {};
    \node at (-1,-3.2-.75) {${\vx}$};
    \node[minimum width=7mm,minimum height=3.5mm,fill=white,opacity=1,draw=black,rounded corners=0.5pt] at (4.5,-3.2-.75) (xhatt) {};
    \node at (4.5,-3.2-.75) {$\tilde{\vx}$};

    \encodert{2.5}{-2}{4}{dec}

    \node[minimum width=0.65cm,fill=white,opacity=0,draw=black,rounded corners=0.5pt] at (4.0,-1.5) (zgateway) {};

    \draw [->, rounded corners=3mm, thick] (xin) -| ([xshift=-2mm]dec-1.south);
    \draw [->, rounded corners=3mm, thick] ([xshift=2mm]dec-2.south) |- (xout.west);
    \draw [->, rounded corners=3mm, thick] ([xshift=-2mm]dec-1.north) |- (z2);

    \draw [->, rounded corners=3mm, thick] (in) -| ([xshift=2mm]dec-1.south);
    \draw [->, rounded corners=3mm, thick] ([xshift=-1.8mm]dec-2.south) |- (in.east);
    \draw [->, rounded corners=1mm, thick] ([xshift=2mm]dec-1.north) |- (zhat);
    \draw [->, rounded corners=3mm, thick] (z) |- (dec-2.west);
    \draw [->, rounded corners=3mm, thick] (norm2box.south) -- (z);
    \draw [rounded corners=3mm, thick] (z2) -| (zgateway.north);
    \draw [->, rounded corners=3mm, thick] (zgateway.north) |- (dec-2.east);

    \draw [rounded corners=3mm, thick] (xhatt.south) |- (2.0,-4.55);
    \draw [rounded corners=3mm, thick] (1.5,-4.55) -| (xt.south);

    \draw [rounded corners=3mm, thick,lightgray] ([xshift=-1.8mm]dec-2.south) to[out=90,in=0 ] (dec-2.west);
    \draw [thick,lightgray] ([xshift=2mm]dec-2.south) to[out=90,in=180](dec-2.east);
    \draw [thick,lightgray] ([xshift=2mm]dec-1.south) -- ([xshift=2mm]dec-1.north);
    \draw [thick,lightgray] ([xshift=-2mm]dec-1.south) -- ([xshift=-2mm]dec-1.north);

    \recov{1.3}{1.05}{0}{0}
    \recovc{1.3}{-0.75}{0}{0}
    \recoh{1.75}{-4.55}{0}{0}

    \draw [thick] (z2) -- (1.3,1.55-0.25);
    \draw [thick] (norm1box.north) -- (1.3,0.8);
  \draw [thick] (norm1box.south) -- (1.3,-0.5);
  \draw [thick] (zhat) -- (1.3,-1.00);

    \node[block,minimum width={0.25},shape=circle,draw=black,fill=white] (n4) at (1.75,-4.55) {3};    
    \node[block,minimum width={0.25},shape=circle,draw=black,fill=white] (n6) at (1.3,1.05) {1};
    \node[block,minimum width={0.25},shape=circle,draw=black,fill=white] (n7) at (1.3,-0.75) {2};    
            
    \end{tikzpicture}}
    \caption{Encoder loss}
    \label{fig:encoder-loss}     
  \end{subfigure}\\[1em]
  \begin{subfigure}[b]{.4\textwidth}
      \centering

        \tikzstyle{block} = [rounded corners=0.5pt,minimum width=1.5mm,minimum height=1.5mm,inner sep=0,draw=black!50!black,fill=white]
              
        \scalebox{0.75}{%
        \begin{tikzpicture}[inner sep=0]

        \node at (1.75,-3.2) (in) {\includegraphics[width=1.5cm]{./fig/ffhq_sample/42420N/42420001_1000.jpg}};

        \node [minimum width=3mm,minimum height=3mm] at (0.8,-1.6) (zhat) {$\hat{\vz}$};
        \node [minimum width=3mm,minimum height=3mm] at (2.7,-1.6) (z) {$\vz$};
        \node [minimum width=3mm,minimum height=3mm] at (5.0,-0.8) (zA) {${\vz}_A$};
        \node [minimum width=3mm,minimum height=3mm] at (4.5,-1.5) (zB) {${\vz}_B$};

        \node[minimum width=7mm,minimum height=6mm,fill=white,opacity=1,draw=black,rounded corners=0.1pt] at (0.8,0.2) (norm1box) {};
        \node[minimum width=7mm,minimum height=6mm,fill=white,opacity=1,draw=black,rounded corners=0.1pt] at (2.7,0.2) (norm2box) {};
        \node at (0.8,0.2) (norm-1) {\includegraphics[width=0.6cm]{./fig/architecture2/norm.jpg}};
        \node at (2.7,0.2) (norm-2) {\includegraphics[width=0.6cm]{./fig/architecture2/norm.jpg}};
        \node[minimum width=0.6cm,fill=white,opacity=0,draw=black,rounded corners=0.5pt] at (2.8,0.0) (norm2bottom) {};
    
        \node[minimum width=0.65cm,fill=white,opacity=0,draw=black,rounded corners=0.5pt] at (3.5,-1.9) (thetatop) {};
        \node[minimum width=0.65cm,fill=white,opacity=0,draw=black,rounded corners=0.5pt] at (3.5,-2.0) (thetamid) {};
        \node[minimum width=0.65cm,fill=white,opacity=0,draw=black,rounded corners=0.5pt] at (3.5,-2.1) (thetabottom) {};

        \node[minimum width=7mm,minimum height=3.5mm,fill=white,opacity=1,draw=black,rounded corners=0.5pt] at (1.75,-3.2-.75) {};
        \node at (1.75,-3.2-.75) {$\hat{\vx}$};
    
        \encoder{3.5}{-2}{4}{dec}

        \draw [->, rounded corners=3mm, thick] (in) -| (dec-1);
        \draw [->, rounded corners=3mm, thick] (dec-2) |- (in);
        \draw [->, rounded corners=1mm, thick] (dec-1) |- (zhat);
        \draw [->, rounded corners=3mm, thick] (z) |- (thetamid);
        \draw [->, rounded corners=3mm, thick] (zA) |- (thetabottom);
        \draw [->, rounded corners=3mm, thick] (zB) |- (thetatop);

        \draw [->, rounded corners=3mm, thick] (norm2box.south) -- (z);
        \draw [->, rounded corners=3mm, thick] ([yshift=1mm]norm2box.east) -| (zA);
        \draw [->, rounded corners=3mm, thick] ([yshift=-1mm]norm2box.east) -| (zB);
    
        \recoh{1.75}{-1.6}{0}{0}
        \recov{0.8}{-0.75}{0}{0}
        \draw [thick] (zhat) -- (1.5,-1.6);

        \draw [thick] (z) -- (2.0,-1.6);

      \draw [thick] (zhat) -- (0.8,-1.00);
      \draw [thick] (norm1box.south) -- (0.8,-0.6);

        \node[block,minimum width={0.25},shape=circle,draw=black,fill=white] (n4) at (2.7,-0.7) {4};
        \node[block,minimum width={0.25},shape=circle,draw=black,fill=white] (n5) at (3.3,0.2) {5};
        \node[block,minimum width={0.25},shape=circle,draw=black,fill=white] (n6) at (0.8,-0.75) {6};
        \node[block,minimum width={0.25},shape=circle,draw=black,fill=white] (n7) at (1.75,-1.6) {7};

        \end{tikzpicture}}
        \caption{Decoder loss}
        \label[figure]{fig:decoder-loss}

  \end{subfigure}
  \caption{Breakdown of the 1-pass flow loss terms.}

  \label{fig:lossterms} 

\end{wrapfigure}
\textbf{Full unsupervised loss~~}
We expect the fusion images to increase the number  of outliers during training. To manage this, we replace L1/L2 in \cref{eq:loss-orig-phi} by a robust
 loss $d_{\rho}$ \citep{Barron17}. $d_{\rho}$ generalizes various norms
  via an explicit parameter vector $\valpha$. Thus, $\mathcal{L}_\phi$ remains as in \cref{eq:loss-orig-phi} but with $d_\mathcal{X} = d_{\rho}$, and
\begin{multline}
  \mathcal{L}_\theta = \mathrm{D_{KL}}[q_\phi(\vz\mid\hat{\vx}) \,\|\, \mathrm{N}(\vect{0},\MI)] \\ + \lambda_\mathcal{Z}\,d_{\mathrm{cos}}(\vz, \vphi(\vtheta(\vz))) + \mathcal{L}_j, \label{eq:loss-theta-phi}
\end{multline}
where $\hat{\vx}_{1:\frac{3}{4}M} \sim q_\theta(\vx \mid \vz)$ with $\vz \sim \mathrm{N}(\vect{0},\MI)$, and $\hat{\vx}_{\frac{3}{4}M:M} \sim q_\theta(\vx \mid \hat{\vz}_{AB})$, with a set 3:4 ratio of regular and mixed samples for batch size $M$, $j \sim \mathrm{U}\{1,N\}$, and $\hat{\vz}_{AB}$ from \cref{eq:zAB}. Margin $M_\mathrm{gap} = 0.5$, except for \CelebAHQ and Bedrooms $128{\times}128$ ($M_\mathrm{gap} = 0.2$) and  \CelebAHQ $256{\times}256$ ($M_\mathrm{gap} = 0.4$).
To avoid discontinuities in $\valpha$, we introduce a progressively-growing variation of $d_{\rho}$, where we first learn the $\valpha$ in the lowest resolution (\eg, $4 {\times} 4$). There, each $\alpha_i$ corresponds to one pixel $p_{x,y}$. Then, whenever doubling the resolution, we initialize the new---now four times as large---$\valpha$ in the higher resolution by replicating each $\alpha_i$ to cover the new $\alpha_j^{1\times{4}}$ that now corresponds to $p_{x,y}$, $p_{x+1,y}$, $p_{x,y+1}$ and $p_{x+1,y+1}$, in the higher resolution.

\newcommand{\ball}[1]{\scalebox{.75}{\tikz[baseline]\node[circle,draw,inner sep=1pt,anchor=base,yshift=1pt]{#1};}}
We summarize the final loss computation as follows. At the encoder training step (\cref{fig:encoder-loss})\citep{ulyanov2017}, we compute $\mathcal{L}_\phi$ by first encoding training samples $\vx$ into latents $\vz$, minimizing the KL divergence between the distribution of $\vz$ and $\mathrm{N}(\vect{0},\MI)$ \ball{1}. Simultaneously, we encode randomly generated samples $\hat{\vx}$ into $\hat{\vz}$, maximizing their corresponding divergence from $\mathrm{N}(\vect{0},\MI)$ \ball{2}. We also decode each $\vz$ into $\tilde{\vx}$, with the reconstruction error $d_\mathcal{X}(\vx, \tilde{\vx})$ \ball{3}. At the decoder training step, we first compute the 1-pass terms of $\mathcal{L}_{\theta}$ (\cref{fig:decoder-loss}) by generating random samples $\hat{\vx}$, each decoded from either  a single ${\vz}$~ \ball{4} or a mixture pair $({\vz}_A,{\vz}_B)$ ~\ball{5} drawn from $\mathrm{N}(\vect{0},\MI)$. We encode each $\hat{\vx}$ into $\hat{\vz}$ and minimize the KL divergence between their distribution and $\mathrm{N}(\vect{0},\MI)$ \ball{6}. We compute the latent reconstruction error $d_{\mathrm{cos}}$ between each $\vz$ and its re-encoded counterpart $\hat{\vz}$ \ball{7}. Finally, for $({\vz}_A,{\vz}_B)$, we do the second pass, adding the term $\mathcal{L}_j$ (see \cref{fig:2-pass}).

\subsection{Enforcing Known Invariances at Specific Layers}
\label{sec:invariances}
As an extension to the main approach described so far, one can independently consider the following. The architecture and the cyclic training method also allow for a novel principled approach to leverage known scale-specific invariances in training data. 
Assume that images $x_1$ and $x_2$ have identical characteristics at some scales, but differ on others, with this information further encoded into $z_1$ and $z_2$, correspondingly. In the automodulator, we could try to have the shared information affect only the decoder layers \#$j{:}k$. For any $\vxi^{(j-1)}$, we then must have $\theta_{j:k}(\vxi^{(j-1)}, \vz_1) = \theta_{j:k}(\vxi^{(j-1)}, \vz_2)$. Assume that it is possible to represent the rest of the information in the images of that data set in layers \#$1{:}(j-1)$ and \#$(k+1){:}N$. This situation occurs, \eg, when two images are known to differ only in high-frequency properties, representable in the `fine' layers. By mutual independence of layers, our goal is to have $\vz_1$ and $\vz_2$ interchangeable at the middle:
\begin{align}
  \vtheta_{k+1:N}(\vtheta_{j:k}(\vtheta_{1:j-1}(\vxi^{(0)}, \vz_2), \vz_1), \vz_2) 
  &= \vtheta_{k+1:N}(\vtheta_{j:k}(\vtheta_{1:j-1}(\vxi^{(0)}, \vz_2), \vz_2), \vz_2) \nonumber \\ &= \vtheta_{1:N}(\mathrm{\vxi^{(0)}}, \vz_{2}) = \vtheta(\vphi(\vx_2)),
\end{align}
which turns into the optimization target (for some distance function $d$)
\begin{align}
  d(\vtheta(\vphi(\vx_2)), \vtheta_{k+1:N}(\vtheta_{j:k}(\vtheta_{1:j-1}(\vxi^{(0)}, \vz_2), \vz_1), \vz_2)) \label{eq:loss-inv0}.
\end{align}
By construction of $\vphi$ and $\vtheta$, this is equivalent to directly minimizing
\begin{align}
  \mathcal{L}_\mathrm{inv} &= d(\vx_2, \vtheta_{k+1:N}(\vtheta_{j:k}(\vtheta_{1:j-1}(\vxi^{(0)}, \vz_2), \vz_1), \vz_2)), \label{eq:loss-inv}
\end{align}
where $\vz_1 = \vphi(\vx_1)$ and $\vz_2 = \vphi(\vx_2)$. By symmetry, the complement term $\mathcal{L}_\mathrm{inv}'$ can be constructed by swapping $z_1$ with $z_2$ and $x_1$ with $x_2$. For each known invariant pair $\vx_1$ and $\vx_2$ of the minibatch, you can now add the terms $\mathcal{L}_\mathrm{inv} + \mathcal{L}_\mathrm{inv}'$ to $\mathcal{L}_\phi$ of \cref{eq:loss-theta-phi}. Note that in the case of $\vz_1 = \vz_2$, $\mathcal{L}_\mathrm{inv}$ reduces to the regular sample reconstruction loss, revealing our formulation as a generalization thereof.

As we push the invariant information to layers \#$j{:}k$, and the other information \textit{away} from them, there are less layers available for the rest of the image information. Thus, we may need to add extra layers to retain the overall decoder capacity. Note that in a pyramidal deconvolutional stack where the resolution increases monotonically, if the layers \#$j{:}k$ span more than two consecutive levels of detail, the scales in-between cannot be extended in that manner.

\section{Experiments}
\label{sec:exper}
Since automodulators offer more applications than either typical autoencoders or GANs without an encoder, we strive for reasonable performance across experiments, rather than beating any specific metric.  (Experiment details in \cref{sec:trainingdetails}.)
Any generative model can be evaluated in terms of sample quality and diversity. To measure them, we use Fr\'echet inception distance (FID) \citep{heusel2017}, which is comparable across models when sample size is fixed \citep{binkowski2018}, though notably uninformative about the ratio of precision and recall \citep{kynkaanniemi2019improved}. Encoder--decoder models can further be evaluated in terms of their ability to reconstruct new test inputs, which underlies their ability to perform more interesting applications such as latent space interpolation and, in our case, mixing of latent codes. For a similarity metric between original and reconstructed face images (center-cropped), we use LPIPS \citep{zhang2018}, a metric with better correspondence to human evaluation than, \eg, traditional L2 norm.

The degree of latent space disentanglement is often considered the key property of a latent variable model. Qualitatively, it is the necessary condition for, \eg, style mixing capabilities. Quantitatively, one could expect that, for a constant-length step in the latent space, the less entangled the model, the smaller is the overall perceptual change. The extent of this change, measured by LPIPS, is the basis of measuring disentanglement as Perceptual Path Length (PPL) \citep{karras2018style}.
\begin{wraptable}{r}{7cm}
  \vspace*{-4pt}
  \caption{Effect of loss terms on \CelebAHQ  at $256 {\times} 256$ with 40M seen samples (50k FID batch) before applying layer noise.}

  \label{tbl:abl}
  \centering
  {\noindent\scriptsize
   \setlength{\tabcolsep}{4pt}
  \begin{tabular}{lccc} 
    \toprule
     & FID & FID (mix) & PPL \\ 
    \midrule
    Automodulator architecture         &$ 45.25$ & $ 52.83 $ & $206.3$\\
    + Loss $\mathcal{L}_j$                &$44.06$ & $ 47.74 $ & ${210.0}$\\
    + Loss $d_\rho$ replacing L1              &${\bf 36.20}$ & $ { 43.53 }$ & $217.3$\\
    + Loss $\mathcal{L}_j$ + $d_\rho$ replacing L1              &$37.95$ & $ {\bf 40.90 }$ & ${\bf 201.8 }$\\

    \bottomrule
  \end{tabular}}
\end{wraptable}
We justify our choice of a loss function in \cref{eq:loss-theta-phi}, compare to baselines on relevant measures, demonstrate the style-mixing capabilities specific to automodulators, and show a proof-of-concept for leveraging scale-specific invariances (see \cref{sec:invariances}). 
In the following, we use \cref{eq:loss-orig-phi,eq:loss-theta-phi}, and polish the output by adding a source of unit Gaussian noise with a learnable scaling factor before the activation in each decoder layer, as in StyleGAN \citep{karras2018style}, also improving FID.

\paragraph{Ablation study for the loss metric}
\label{sec:ablstudy}
In \cref{tbl:abl}, we illustrate the contribution of the layer disentanglement loss $\mathcal{L}_j$ and the robust loss $d_{\rho}$ on the FID for regular and mixed samples from the model at $256{\times}256$ resolution, as well as PPL. We train the model variants on \CelebAHQ \citep{karras2017} data set to convergence (40M seen samples) and choose the best of three restarts with different random seeds.
Our hypothesis was that $\mathcal{L}_j$ improves the FID of mixed samples and that replacing L1 sample reconstruction loss with $d_{\rho}$ improves FID further and makes training more stable. The results confirm this. Given the improvement from $d_{\rho}$ also for the mixed samples, we separately tested the effect of $d_{\rho}$ without $\mathcal{L}_j$ and find that it produces even slightly better FID for regular samples but then considerably worse FID for the mixed ones, due to, presumably, more mutually entangled layers. For ablation of the $M_\mathrm{gap}$ term, see \citep{heljakka2019towards}. The effect of the term $d_{\mathrm{cos}}$ was studied in \citep{ulyanov2017} (for $cos$ instead of L2, see \citep{ulyanovGithub}).

\begin{table*}[!b]
   \caption{Performance in \CelebAHQ (CAHQ), FFHQ, and LSUN Bedrooms and Cars. We measure LPIPS, Fr\'echet Inception Distance (FID), and perceptual path length (PPL). Resolution is $256{\times}256$, except *$128{\times}128$. For all numbers, \textbf{smaller is better}. Only the `-AdaIn' architectures are functionally equivalent to the automodulator (encoding and latent mixing). \protect\tikz[baseline]\protect\node[fill=gray!10,anchor=base]{GANs in gray.};}
  \begin{subfigure}[t]{.39\textwidth} 
   \caption{Encoder--decoder comparison}
   \label{tbl:results}
   \centering
   {\noindent\scriptsize
   \setlength{\tabcolsep}{3pt}
   \begin{tabular*}{\textwidth}{lcccccc}
   \toprule
     & LPIPS & FID & PPL \\
     & (CAHQ$^*$) & (CAHQ$^*$) & (CAHQ$^*$) \\
   \midrule
   B-\PIONEER & $0.092$ &  {$\bf 19.61$}  & $ 92.8 $\\
   WAE-AdaIn  & $0.165$ & $99.81$ & $\bf 62.2 \bf$ \\
   WAE-classic  & $0.162$ & $112.06$ & $236.8 $ \\
   VAE-AdaIn  & $0.267$ & $114.05$ & $83.5$ \\
   VAE-classic  & $0.291$ & $173.81$ & $71.7$ \\
   Automodulator & $\bf0.083$ &  $27.00$ & $\bf 62.3$\\

   \bottomrule
   \end{tabular*}}

  \end{subfigure}
  \hfill
  \begin{subfigure}[t]{.59\textwidth}

   \caption{Generative models comparison}
   \label{tbl:results2}
   \centering
   {\noindent\scriptsize%
   \setlength{\tabcolsep}{3.3pt}
   \begin{tabular*}{\textwidth}{lcccccc}
   \toprule
     & FID  & FID & FID & FID & PPL & PPL \\
     & (CAHQ) & (FFHQ) & (Bedrooms) & (Cars) & (CAHQ) & (FFHQ) \\
   \midrule
   \rowcolor{gray!10}
   StyleGAN  & $\bf 5.17$ & $ 4.68 $ & $\bf 2.65$ & $3.23 $ & $ 179.8  $ & $234.0$ \\
   \rowcolor{gray!10} 
   StyleGAN2  & --- & $\bf 3.11$ & --- & $\bf 5.64$ & ---  & $\bf 129.4$ \\
   \rowcolor{gray!10}
   PGGAN  & $7.79$ & $8.04$ & $8.34$ & $8.36$ & $229.2$ & $412.0$ \\
   GLOW  & $68.93$ & --- & --- & --- & $219.6$ & --- \\
   B-\PIONEER &  {$25.25$} &  $61.35$ & $21.52$ &$42.81$  & $\bf 146.2$ & $160.0$\\
   Automodulator &   {$29.13$} &  $31.64$ & $25.53$ & $19.82$  & $ 203.8 $ & $250.2$ \\

   \bottomrule
   \end{tabular*}}
  \end{subfigure}   
  \vspace*{-1em}
\end{table*}

\paragraph{Encoding, decoding, and random sampling}
\label{sec:mainexp}
To compare encoding, decoding, and random sampling performance, autoencoders are more appropriate baselines than GANs without an encoder, since the latter tend to have higher quality samples, but are more limited
since they cannot manipulate real input samples. However, we do also require reasonable sampling performance from our model, and hence separately compare to non-autoencoders.
In \cref{tbl:results}, we compare to autoencoders: Balanced \PIONEER \citep{heljakka2019towards}, a vanilla VAE, and a more recent Wasserstein Autoencoder (WAE) \citep{tolstikhin2018}. We train on $128{\times}128$ \CelebAHQ, with our proposed architecture (`AdaIn') and the regular one (`classic'). We measure LPIPS, FID (50k batch of generated samples compared to training samples, STD over 3~runs $< 1$ for all models) and PPL. Our method has the best LPIPS and PPL.

\begin{figure}[!t]
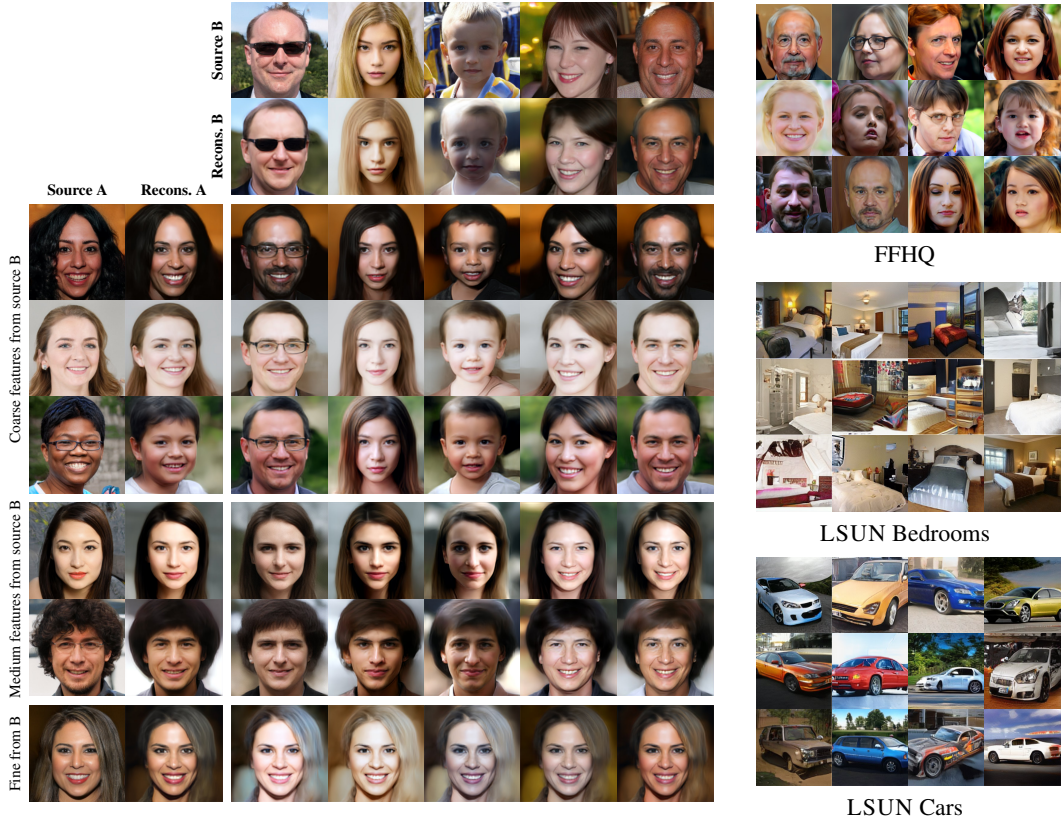

  \begin{subfigure}{.67\textwidth}
  \vspace*{-6pt}
  \setlength{\figurewidth}{.09\textwidth}
  \setlength{\figurewidth}{.14\textwidth}
  \setlength{\figureheight}{\figurewidth}
  \centering\tiny
  \resizebox{\textwidth}{!}{%
  \begin{tikzpicture}[inner sep=0]

  \def\id{42420N}

  \tikzstyle{arrow} = [draw=black!10, single arrow, minimum height=10mm, minimum width=3mm, single arrow head extend=1mm, fill=black!10, anchor=center, rotate=-90, inner sep=2pt]

  \foreach \x [count=\i] in {0,1,2,3,4} {
    \node[] at ({\figurewidth*\i},{1.1*\figureheight}) {\includegraphics[width=\figurewidth]{./fig/fig_mix/\id/source-B-\x.jpg}};
  }

  \foreach \x [count=\i] in {0,1,2,3,4} {
    \node[] at ({\figurewidth*\i},{.1*\figureheight}) {\includegraphics[width=\figurewidth]{./fig/fig_mix/\id/source-B-\x-reco.jpg}};
  }

  \def\gap{0}
  \foreach \y [count=\j] in {0,1,2,3,4,5} {
    \ifnum \y>2
      \def\gap{.1}
    \fi
    \ifnum \y>4
      \def\gap{.2}
    \fi    
    \node[] at ({-1.1*\figurewidth},{-\figureheight*\j-\gap*\figureheight}) {\includegraphics[width=\figurewidth]{./fig/fig_mix/\id/source-A-\y.jpg}};
  }

  \def\gap{0}
  \foreach \y [count=\j] in {0,1,2,3,4,5} {
    \ifnum \y>2
      \def\gap{.1}
    \fi
    \ifnum \y>4
      \def\gap{.2}
    \fi    
    \node[] at ({-.1*\figurewidth},{-\figureheight*\j-\gap*\figureheight}) {\includegraphics[width=\figurewidth]{./fig/fig_mix/\id/source-A-\y-reco.jpg}};
  }

  \def\gap{0}
  \foreach \y [count=\j] in {0,1,2,3,4,5} {
    \ifnum \y>2
      \def\gap{.1}
    \fi
    \ifnum \y>4
      \def\gap{.2}
    \fi   
    \foreach \x [count=\i] in {0,1,2,3,4} {
      \node[] at ({\figurewidth*\i},{-\figureheight*\j-\gap*\figureheight}) {\includegraphics[width=\figurewidth]{./fig/fig_mix/\id/mix-\y-\x.jpg}};
    }
  }

  \node[align=center] at (-1.1*\figureheight,-.35\figurewidth) {\bf Source A};
  \node[align=center] at (-.1*\figureheight,-.35\figurewidth) {\bf Recons.\ A};  
  \node[align=center,rotate=90] at (.35\figurewidth,1.1\figureheight) {\bf Source B};
  \node[align=center,rotate=90] at (.35\figurewidth,.1\figureheight) {\bf Recons.\ B};  
  \node[align=center,rotate=90] at (-1.75\figurewidth,-2\figureheight) {Coarse features from source B};
  \node[align=center,rotate=90] at (-1.75\figurewidth,-4.6\figureheight) {Medium features from source B};
  \node[align=center,rotate=90] at (-1.75\figurewidth,-6.2\figureheight) {Fine from B};
        
  \end{tikzpicture}}
  \end{subfigure}
  \hfill  
  \begin{subfigure}{.29\textwidth}
  \begin{subfigure}[b]{\textwidth}
    \centering
    \setlength{\figurewidth}{.25\textwidth}
    \setlength{\figureheight}{\figurewidth}
    \begin{tikzpicture}[inner sep=0]
      \foreach \x [count=\i] in {0,...,9}
        \node[draw=white,fill=black!20,minimum size=\figurewidth,inner sep=0pt]
          (\i) at ({\figurewidth*mod(\i-1,4)},{-\figureheight*int((\i-1)/4)})
          {\includegraphics[width=\figurewidth]{./fig/ffhq_sample/42420N/42420001_100\x.jpg}};
      \foreach \x [count=\i] in {0,...,1}
          \node[draw=white,fill=black!20,minimum size=\figurewidth,inner sep=0pt]
            (\i) at ({\figurewidth*mod(\i+9,4)},{-\figureheight*int((\i+9)/4)})
            {\includegraphics[width=\figurewidth]{./fig/ffhq_sample/42420N/42420001_101\x.jpg}};
      \end{tikzpicture}
    {\footnotesize FFHQ}
    \label{fig:sample-faces}    
  \end{subfigure}\\[5pt]
  \begin{subfigure}[b]{\textwidth}
    \centering
    \setlength{\figurewidth}{.25\textwidth}
    \setlength{\figureheight}{\figurewidth}
    \begin{tikzpicture}[inner sep=0]
      \foreach \x [count=\i] in {0,...,9}
        \node[draw=white,fill=black!20,minimum size=\figurewidth,inner sep=0pt]
          (\i) at ({\figurewidth*mod(\i-1,4)},{-\figureheight*int((\i-1)/4)})
          {\includegraphics[width=\figurewidth]{./fig/bedroom_sample/27572k/27574285_100\x.jpg}};
      \foreach \x [count=\i] in {0,...,1}
          \node[draw=white,fill=black!20,minimum size=\figurewidth,inner sep=0pt]
            (\i) at ({\figurewidth*mod(\i+9,4)},{-\figureheight*int((\i+9)/4)})
            {\includegraphics[width=\figurewidth]{./fig/bedroom_sample/27572k/27574285_101\x.jpg}};
      \end{tikzpicture}
    {\footnotesize \LSUN Bedrooms}
    \label{fig:sample-bedrooms}    
  \end{subfigure}\\[5pt]
  \begin{subfigure}[b]{\textwidth}
    \centering 
    \setlength{\figurewidth}{.25\textwidth}
    \setlength{\figureheight}{\figurewidth}    
    \begin{tikzpicture}[inner sep=0]
      \foreach \x [count=\i] in {0,...,11}
        \node[draw=white,fill=black!20,minimum size=\figurewidth,inner sep=0pt]
          (\i) at ({\figurewidth*mod(\i-1,4)},{-\figureheight*int((\i-1)/4)})
          {\includegraphics[width=\figurewidth]{./fig/car_sample/32057141_\x.jpg}};
    \end{tikzpicture}
    {\footnotesize \LSUN Cars}
    \label{fig:sample-cars}
  \end{subfigure}  
  \end{subfigure}    
  \caption{(Left): Feeding the random fake source images in \citet{karras2018style} into our model as `real' inputs, reconstructing at $512{\times}512$ and mixing at three scales. (The same for real faces, see Supplement.) (Right): Uncurated random samples of $512{\times}512$ FFHQ and $256{\times}256$ \LSUN.}
  \label{fig:stylegan-mixing}

\end{figure}

In \cref{tbl:results2}, we compare to non-autoencoders: StyleGAN, Progressively Growing GAN (PGGAN) \citep{karras2017}, and GLOW \citep{Kingma+Dhariwal:2018}. To show that our model can reasonably perform for many data sets, we train at $256{\times}256$ on \CelebAHQ, FFHQ \citep{karras2018style}, \LSUN Bedrooms and \LSUN Cars \citep{Yu:2015}. We measure PPL and FID (uncurated samples in \cref{fig:stylegan-mixing} (right), STD of FID over 3~runs $<.1$).
The performance of the automodulator is comparable to the Balanced \PIONEER on most data sets. GANs have clearly best FID results on all data sets (NB: a hyper-parameter search with various schemes was used in \citep{karras2018style} to achieve their high PPL values). We train on the actual 60k training set of FFHQ only (StyleGAN trained on all 70k images). We also tested what will happen if we try invert the StyleGAN by finding a latent code for an image by an optimization process. Though this can be done, the inference is over 1000 times slower to meet and exceed the automodulator LPIPS score (see \cref{app:comparison-to-gan-inv,fig:stylegan-comp}).
We also evaluate the 4-way image interpolation capabilities in unseen FFHQ test images (\cref{fig:interpolation} in the supplement) and observe smooth transitions. Note that in GANs without an encoder, one can only interpolate between the codes of {\em random} samples, revealing little about the recall ability of the model.

\paragraph{Style mixing}
\label{sec:stylemixexp}
The key benefit of the automodulators over regular autoencoders is the style-mixing capability (\cref{fig:1-pass}), and the key benefit over style-based GANs is that `real' unseen test images can be instantly style-mixed. We demonstrate both in \cref{fig:stylegan-mixing}. For comparison with prior work, we use the randomly generated source images from the StyleGAN paper \citep{karras2018style}. Importantly, for our model, they appear as unseen `real' test images. Performance in mixing real-world images is similar (Supplementary \cref{fig:face-mixing,fig:car-mixing}). In \cref{fig:stylegan-mixing}, we mix specific input faces (from source~A and B) so that the `coarse' (latent resolutions $4 {\times} 4$ -- $8 {\times} 8$), `intermediate' ($16 {\times} 16$ -- $32 {\times} 32$) or `fine' ($64 {\times} 64$ -- $512 {\times} 512$) layers of the decoder use one input, and the rest of the layers use the other.

\begin{figure*}[t!]
  \centering\scriptsize
  \setlength{\figurewidth}{.1\textwidth}
  \setlength{\figureheight}{\figurewidth}  
  \begin{subfigure}[t]{.58\textwidth}
      \centering
      \begin{tikzpicture}[inner sep=0]

      \foreach \x [count=\i] in {0,1,3,5} {
        \node[] at ({\figurewidth*\i-1*\figureheight},{0*\figureheight}) {\includegraphics[width=\figurewidth]{./fig/fig_rotation_invariance/128/source-\x.jpg}};
      }

    \node[] at ({-1.1\figurewidth},{-1.6\figureheight}) {\includegraphics[width=\figurewidth]{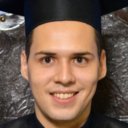}};
  \foreach \x [count=\i] in {0,1,3,5} {
             \node[] at ({\figurewidth*\i-1*\figurewidth},{-1.1\figureheight}) {\includegraphics[width=\figurewidth]{./fig/fig_rotation_invariance/128/mix_\x_2_0-2.jpg}};
             \node[] at ({\figurewidth*\i-1*\figurewidth},{-2.1\figureheight}) {\includegraphics[width=\figurewidth]{./fig/fig_rotation_invariance/128/mix_\x_5_0-2s.jpg}};}

             \node[align=center,rotate=90] at (-1.75*\figureheight,-1.6*\figurewidth) {\bf Source A};

             \node[align=center,rotate=90] at (-.65\figurewidth,0\figureheight) {\bf Source B};
             \node[align=center] at (-1.15*\figurewidth,-.75*\figureheight) {\bf Enforced id. \\ \bf invariance $\rightarrow$};
             \node[align=center] at (-1.15*\figurewidth,-2.35*\figureheight) {\bf Regular $\rightarrow$};

      \end{tikzpicture}
      \caption{Training with enforced identity invariance}
      \label{fig:invariances}
  \end{subfigure}
  \hfill
  \begin{subfigure}[t]{0.4\textwidth}
      \centering

      \begin{tikzpicture}[inner sep=0]
        \node[] at ({-1*\figurewidth},{-1.1\figureheight}) {\includegraphics[width=\figurewidth]{./fig/fig_rotation_invariance/128/source-2B.jpg}};
        \node[] at ({-1*\figurewidth},{-2.1\figureheight}) {\includegraphics[width=\figurewidth]{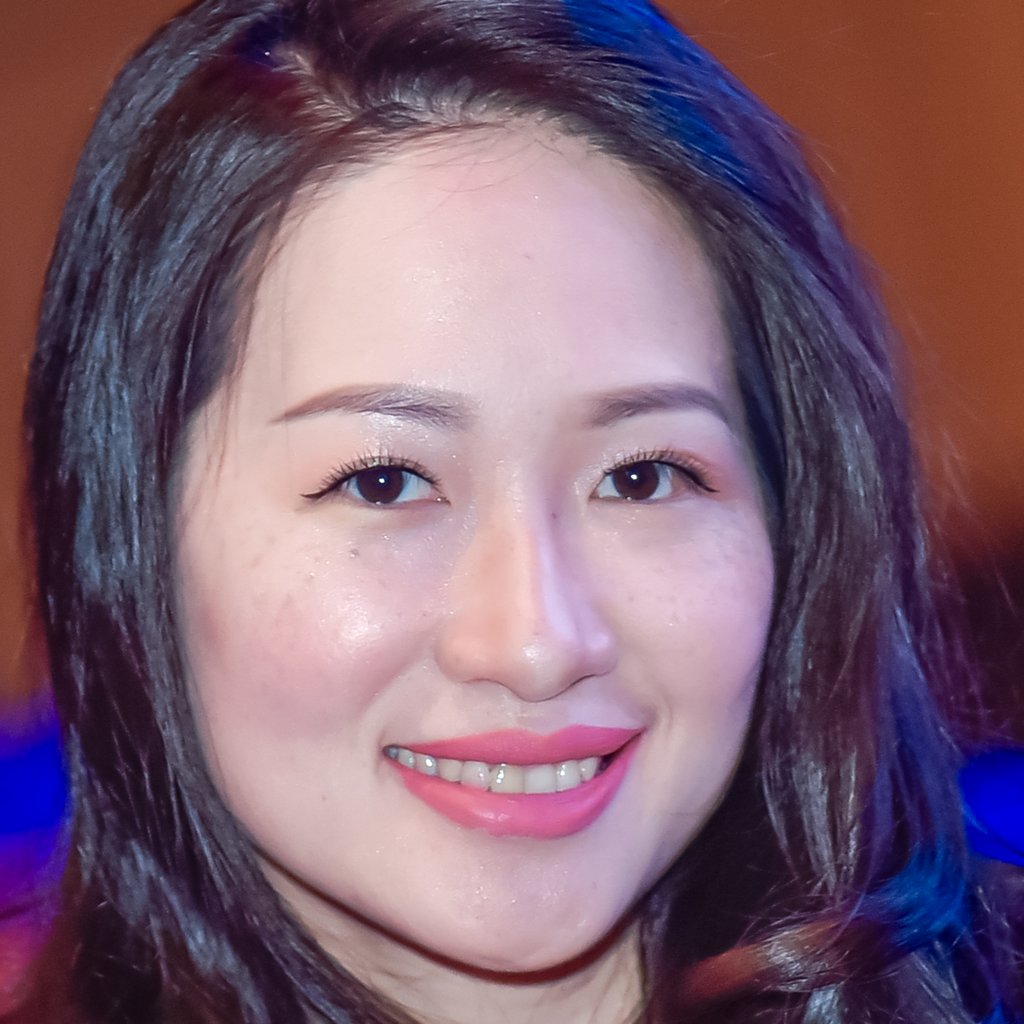}};
        
      \foreach \x [count=\i] in {1, 2, 3} {
        \node[] at ({\figurewidth*\i-1*\figurewidth},{-1.1\figureheight}) {\includegraphics[width=\figurewidth]{./fig/attribute/\x/1.jpg}};
        \node[] at ({\figurewidth*\i-1*\figurewidth},{-2.1\figureheight}) {\includegraphics[width=\figurewidth]{./fig/attribute/\x/B1.jpg}};}

    \foreach \x [count=\i] in {Input, No smile, Fe/male, Glasses} {
      \node[fill=white,inner sep=0, minimum width=4mm] at ({\figurewidth*\i-2*\figurewidth},-0.35\figureheight) {\bf \x};
    }

    \end{tikzpicture}

      \caption{Scale-specific attribute modification}
      \label{fig:attribs}
  \end{subfigure}
  \caption{Examples of controlling individual decoder layer ranges at training time and at evaluation time. (a)~Training with face identity invariance enforcement under azimuth rotation. We generate images with the `non-coarse' styles of source A and the `coarse' ones from each top row image. With `Enforced identity invariance', the top row only drives the face pose while conserving identity. In comparison, the `Regular' training lets the top row also affect other characteristics, including identity. (b)~Modifying an attribute in latent space by using only 4 exemplar images of it. In `regular' all-scales manipulation, the variance in the exemplars causes unwanted changes in, \eg, texture and pose. When the latent vector only drives the relevant scales, the variance in other scales is inconsequential.}

\end{figure*}

\paragraph{Invariances in a weakly supervised setup}
In order to leverage the method of \cref{sec:strong-conservation}, one needs image data that contains pairs or sets that share a scale-specific prominent invariant feature (or, conversely, are identical in every other respect except that feature). To this end, we demonstrate a proof-of-concept experiment that uses the simplest image transformation possible: horizontal flipping. For \CelebAHQ, this yields pairs of images that share every other property except the azimuth rotation angle of the face, making the face identity invariant amongst each pair. Since the original rotation of faces in the set varies, the flip-augmented data set contains faces rotated across a wide continuum of angles.
For further simplicity, we make an artificially strong hypothesis that the 2D projected face shape is the only relevant feature at $4{\times}4$ scale and does not need to affect scales finer than $8{\times}8$. This lets us enforce the $\mathcal{L}_\mathrm{inv}$ loss for layers~\#1--2. Since we do not want to restrict the scale $8{\times}8$ for the shape features alone, we add an extra $8{\times}8$ layer after layer~\#2 of the regular stack, so that layers~\#2--3 both operate at $8{\times}8$, layer~\#4 only at $16{\times}16$, \etc\ 
Now, with $\vz_2$ that corresponds to the horizontally flipped counterpart of $\vz_1$, we have $\vtheta_{3:N}(\vxi^{(2)}, \vz_1) = \vtheta_{3:N}(\vxi^{(2)}, \vz_2)$. Our choices amount to $j=3, k=N$, allowing us to drop the outermost part of \cref{eq:loss-inv}. Hence, our additional encoder loss terms are
\begin{align}
\mathcal{L}_\mathrm{inv} &= d(\vx_2, \vtheta_{3:N}(\vtheta_{1:2}(\vxi^{(0)}, \vz_2), \vz_1)) \quad \text{and} \\
\mathcal{L}_\mathrm{inv}' &= d(\vx_1, \vtheta_{3:N}(\vtheta_{1:2}(\vxi^{(0)}, \vz_1), \vz_2)).
\end{align}
\cref{fig:invariances} shows the results after training with the new loss (50\% of the training samples flipped in each minibatch). With the invariance enforcement, the model forces decoder layers~\#1--2 to only affect the pose. We generate images by driving those layers with faces at different poses, while modulating the rest of the layers with the face whose identity we seek to conserve. The resulting face variations now only differ in terms of pose, unlike in regular automodulator training.

\paragraph{Scale-specific attribute editing}
Consider the mean difference in latent codes of images that display or do not display an attribute of interest (\eg, smile). Appropriately scaled, such codes can added to any latent code to modify that attribute. Here, one can restrict the effect of the latent code only to the layers driving the expected scale of the attribute (\eg, $16{\times}16 - 32{\times}32$), yielding precise manipulation (\cref{fig:attribs}, comparisons in Supplement) with only a few exemplars (\eg, \citep{heljakka2019towards} used 32).

\section{Discussion and Conclusion}
In this paper, we proposed a new generative autoencoder model with a latent representation that independently modulates each decoder layer. The model supports reconstruction and style-mixing of real images, scale-specific editing and sampling.
Despite the extra skill, the model still largely outperforms or matches other generative autoencoders in terms of latent space disentanglement, faithfulness of reconstructions, and sample quality. We use the term {\em automodulator} to denote any autoencoder that uses the latent code only to modulate the statistics of the information flow through the layers of the decoder. This could also include, \eg, 3D or graph convolutions.

Various improvements to the model are possible. The mixture outputs still show occasional artifacts, indicating that the factors of variation have not been perfectly disentangled. Also, while the layer-induced noise helps training, using it in evaluation to add texture details would often reduce output quality. Also, to enable even more general utility of the model, the performance could be measured on auxiliary downstream tasks such as classification.

Potential future applications include introducing completely interchangeable `plugin' layers or modules in the decoder, trained afterwards on top of the pretrained base automodulator, leveraging the mutual independence of the layers. The affine maps themselves could also be re-used across domains, potentially offering mixing of different domains. Such examples highlight that the range of applications of our model is far wider than the initial ones shown here, making the automodulators a viable alternative to state-of-the-art autoencoders and GANs.

Our source code is available at \url{https://github.com/AaltoVision/automodulator}.

  \section*{Broader Impact}
  The presented line of work intends to shift the focus of generative models from random sample generation towards controlled semantic editing of existing inputs. In essence, the ultimate goal is to offer `knobs' that allow content editing based on high-level features, and retrieving and combining desired characteristics based on examples. While we only consider images, the techniques can be extended to other data domains such as graphs and 3D structures.

  Ultimately, such research could reduce very complex design tasks into approachable ones and thus reduce dependency on experts. For instance, contrast an expert user of a photo editor or design software, carefully tuning details, with a layperson who simply finds images or designs with the desired characteristics and guiding the smart editor to selectively combine them.

  Leveling the playing field in such tasks will empower larger numbers of people to contribute to design, engineering and science, while also multiplying the effectiveness of the experts. The downside of such empowerment will, of course, include the threats of deepfakes and spread of misinformation. Fortunately, public awareness of these abuses has been increasing rapidly.
  We attempt to convey the productive prospects of these technologies by also including image data sets with cars and bedrooms, while comparison with prior work motivates the focus on face images.

\begin{ack}
The authors wish to acknowledge the Aalto Science-IT project and CSC -- IT Center for Science, Finland, for computational resources. Authors acknowledge funding from GenMind Ltd. This research was supported by the Academy of Finland grants 308640, 324345, 277685, 295081, and 309902. We thank Jaakko Lehtinen and Janne Hellsten (NVIDIA) for the StyleGAN latent space projection script (for the baseline, only) and advice on its usage. We also thank Christabella Irwanto, Tuomas Kynk{\"a}{\"a}nniemi, and Paul Chang for comments on the manuscript.
\end{ack}

\begingroup
\small
\bibliographystyle{abbrvnat}
\bibliography{bibliography}

\begin{thebibliography}{50}
\providecommand{\natexlab}[1]{#1}
\providecommand{\url}[1]{\texttt{#1}}
\expandafter\ifx\csname urlstyle\endcsname\relax
  \providecommand{\doi}[1]{doi: #1}\else
  \providecommand{\doi}{doi: \begingroup \urlstyle{rm}\Url}\fi

\bibitem[Abdal et~al.(2019)Abdal, Qin, and Wonka]{image2stylegan}
R.~Abdal, Y.~Qin, and P.~Wonka.
\newblock {Image2StyleGAN}: {H}ow to embed images into the {StyleGAN} latent
  space?
\newblock In \emph{International Conference on Computer Vision (ICCV)}, 2019.

\bibitem[Barron(2019)]{Barron17}
J.~T. Barron.
\newblock A general and adaptive robust loss function.
\newblock In \emph{Proceedings of the IEEE Conference on Computer Vision and
  Pattern Recognition (CVPR)}, pages 4331--4339, 2019.

\bibitem[Berthelot et~al.(2017)Berthelot, Schumm, and Metz]{berthelot2017}
D.~Berthelot, T.~Schumm, and L.~Metz.
\newblock {BEGAN}: {B}oundary equilibrium generative adversarial networks.
\newblock \emph{arXiv preprint arXiv:1703.10717}, 2017.

\bibitem[Binkowski et~al.(2018)Binkowski, Sutherland, Arbel, and
  Gretton]{binkowski2018}
M.~Binkowski, D.~J. Sutherland, M.~Arbel, and A.~Gretton.
\newblock Demystifying {MMD} {GAN}s.
\newblock In \emph{International Conference on Learning Representations
  (ICLR)}, 2018.

\bibitem[Brock et~al.(2019)Brock, Donahue, and Simonyan]{brock2018}
A.~Brock, J.~Donahue, and K.~Simonyan.
\newblock Large scale {GAN} training for high fidelity natural image synthesis.
\newblock In \emph{International Conference on Learning Representations
  (ICLR)}, 2019.

\bibitem[Chen et~al.(2019)Chen, Lucic, Houlsby, and Gelly]{chen2018self}
T.~Chen, M.~Lucic, N.~Houlsby, and S.~Gelly.
\newblock On self modulation for generative adversarial networks.
\newblock In \emph{International Conference on Learning Representations
  (ICLR)}, 2019.

\bibitem[Chen et~al.(2016)Chen, Duan, Houthooft, Schulman, Sutskever, and
  Abbeel]{NIPS2016_6399}
X.~Chen, Y.~Duan, R.~Houthooft, J.~Schulman, I.~Sutskever, and P.~Abbeel.
\newblock {InfoGAN}: {I}nterpretable representation learning by information
  maximizing generative adversarial nets.
\newblock In \emph{Advances in Neural Information Processing Systems 29
  (NIPS)}, pages 2172--2180. Curran Associates, Inc., 2016.

\bibitem[Creswell and Bharath(2019)]{creswell18}
A.~Creswell and A.~A. Bharath.
\newblock Inverting the generator of a generative adversarial network.
\newblock \emph{{IEEE} Transactions on Neural Networks and Learning Systems},
  30\penalty0 (7):\penalty0 1967--1974, 2019.

\bibitem[Donahue and Simonyan(2019)]{DBLP:journals/corr/abs-1907-02544}
J.~Donahue and K.~Simonyan.
\newblock Large scale adversarial representation learning.
\newblock In \emph{Advances in Neural Information Processing Systems
  (NeurIPS)}, pages 10541--10551. Curran Associates, Inc., 2019.

\bibitem[Donahue et~al.(2017)Donahue, {Kr\"ahenb\"uhl}, and
  Darrell]{Donahue2017}
J.~Donahue, P.~{Kr\"ahenb\"uhl}, and T.~Darrell.
\newblock Adversarial feature learning.
\newblock In \emph{International Conference on Learning Representations
  (ICLR)}, 2017.

\bibitem[Dumoulin et~al.(2017{\natexlab{a}})Dumoulin, Belghazi, Poole,
  Mastropietro, Lamb, Arjovsky, and Courville]{dumoulin2016D}
V.~Dumoulin, I.~Belghazi, B.~Poole, O.~Mastropietro, A.~Lamb, M.~Arjovsky, and
  A.~Courville.
\newblock Adversarially learned inference.
\newblock In \emph{International Conference on Learning Representations
  (ICLR)}, 2017{\natexlab{a}}.

\bibitem[Dumoulin et~al.(2017{\natexlab{b}})Dumoulin, Shlens, and
  Kudlur]{dumoulin2016learned}
V.~Dumoulin, J.~Shlens, and M.~Kudlur.
\newblock A learned representation for artistic style.
\newblock In \emph{International Conference on Learning Representations
  (ICLR)}, 2017{\natexlab{b}}.

\bibitem[Gatys et~al.(2016)Gatys, Ecker, and Bethge]{gatys2016image}
L.~A. Gatys, A.~S. Ecker, and M.~Bethge.
\newblock Image style transfer using convolutional neural networks.
\newblock In \emph{Proceedings of the IEEE Conference on Computer Vision and
  Pattern Recognition (CVPR)}, pages 2414--2423, 2016.

\bibitem[Goodfellow et~al.(2014)Goodfellow, Pouget-Abadie, Mirza, Xu,
  Warde-Farley, Ozair, Courville, and Bengio]{goodfellow2014}
I.~Goodfellow, J.~Pouget-Abadie, M.~Mirza, B.~Xu, D.~Warde-Farley, S.~Ozair,
  A.~Courville, and Y.~Bengio.
\newblock Generative adversarial nets.
\newblock In \emph{Advances in Neural Information Processing Systems (NIPS)},
  volume~27, pages 2672--2680. Curran Associates, Inc., 2014.

\bibitem[Gregor et~al.(2015)Gregor, Danihelka, Graves, Rezende, and
  Wierstra]{gregor2015draw}
K.~Gregor, I.~Danihelka, A.~Graves, D.~Rezende, and D.~Wierstra.
\newblock {DRAW}: {A} recurrent neural network for image generation.
\newblock In \emph{Proceedings of the 32nd International Conference on Machine
  Learning (ICML)}, volume~37 of \emph{PMLR}, pages 1462--1471, 2015.

\bibitem[Gregor et~al.(2016)Gregor, Besse, Rezende, Danihelka, and
  Wierstra]{gregor2016towards}
K.~Gregor, F.~Besse, D.~J. Rezende, I.~Danihelka, and D.~Wierstra.
\newblock Towards conceptual compression.
\newblock In \emph{Advances in Neural Information Processing Systems (NIPS)},
  volume~29, pages 3549--3557. Curran Associates, Inc., 2016.

\bibitem[Heljakka et~al.(2018)Heljakka, Solin, and Kannala]{heljakka2018}
A.~Heljakka, A.~Solin, and J.~Kannala.
\newblock Pioneer networks: {P}rogressively growing generative autoencoder.
\newblock In \emph{Asian Conference on Computer Vision (ACCV)}, pages 22--38,
  2018.

\bibitem[Heljakka et~al.(2020)Heljakka, Solin, and
  Kannala]{heljakka2019towards}
A.~Heljakka, A.~Solin, and J.~Kannala.
\newblock Towards photographic image manipulation with balanced growing of
  generative autoencoders.
\newblock In \emph{IEEE Winter Conference on Applications of Computer Vision
  (WACV)}, 2020.

\bibitem[Heusel et~al.(2017)Heusel, Ramsauer, Unterthiner, Nessler, and
  Hochreiter]{heusel2017}
M.~Heusel, H.~Ramsauer, T.~Unterthiner, B.~Nessler, and S.~Hochreiter.
\newblock {GAN}s trained by a two time-scale update rule converge to a local
  {N}ash equilibrium.
\newblock In \emph{Advances in Neural Information Processing Systems (NIPS)},
  volume~30, pages 6626--6637. Curran Associates, Inc., 2017.

\bibitem[Hinton and McClelland(1988)]{hinton1988learning}
G.~E. Hinton and J.~L. McClelland.
\newblock Learning representations by recirculation.
\newblock In \emph{Neural Information Processing Systems (NIPS)}, pages
  358--366. American Institute of Physics, 1988.

\bibitem[Hou et~al.(2019)Hou, Heljakka, and Solin]{hou2019}
Y.~Hou, A.~Heljakka, and A.~Solin.
\newblock {Gaussian Process Priors for View-Aware Inference}.
\newblock \emph{arXiv preprint arXiv:1912.03249}, 2019.

\bibitem[Huang et~al.(2018)Huang, Li, He, Sun, and Tan]{huang2018}
H.~Huang, Z.~Li, R.~He, Z.~Sun, and T.~Tan.
\newblock {IntroVAE}: {I}ntrospective variational autoencoders for photographic
  image synthesis.
\newblock In \emph{Neural Information Processing Systems (NeurIPS)}, volume~31,
  pages 52--63. Curran Associates, Inc., 2018.

\bibitem[Huang and Belongie(2017)]{HuangB17}
X.~Huang and S.~Belongie.
\newblock Arbitrary style transfer in real-time with adaptive instance
  normalization.
\newblock In \emph{Proceedings of the IEEE International Conference on Computer
  Vision (ICCV)}, pages 1501--1510, 2017.

\bibitem[Karras et~al.(2018)Karras, Aila, Laine, and Lehtinen]{karras2017}
T.~Karras, T.~Aila, S.~Laine, and J.~Lehtinen.
\newblock Progressive growing of {GAN}s for improved quality, stability, and
  variation.
\newblock In \emph{International Conference on Learning Representations
  (ICLR)}, 2018.

\bibitem[Karras et~al.(2019)Karras, Laine, and Aila]{karras2018style}
T.~Karras, S.~Laine, and T.~Aila.
\newblock A style-based generator architecture for generative adversarial
  networks.
\newblock In \emph{Proceedings of the IEEE Conference on Computer Vision and
  Pattern Recognition (CVPR)}, pages 4401--4410, 2019.

\bibitem[Karras et~al.(2020)Karras, Laine, Aittala, Hellsten, Lehtinen, and
  Aila]{2019arXiv191204958K}
T.~Karras, S.~Laine, M.~Aittala, J.~Hellsten, J.~Lehtinen, and T.~Aila.
\newblock Analyzing and improving the image quality of {StyleGAN}.
\newblock In \emph{Proceedings of the IEEE Conference on Computer Vision and
  Pattern Recognition (CVPR)}, pages 8110--8119, 2020.

\bibitem[Kingma and Dhariwal(2018)]{Kingma+Dhariwal:2018}
D.~P. Kingma and P.~Dhariwal.
\newblock {Glow}: {G}enerative flow with invertible 1x1 convolutions.
\newblock In \emph{Advances in Neural Information Processing Systems
  (NeurIPS)}, volume~31, pages 10236--10245. Curran Associates, Inc., 2018.

\bibitem[Kingma and Welling(2014)]{kingma2014}
D.~P. Kingma and M.~Welling.
\newblock Auto-encoding variational {B}ayes.
\newblock In \emph{International Conference on Learning Representations
  (ICLR)}, 2014.

\bibitem[Kynk{\"a}{\"a}nniemi et~al.(2019)Kynk{\"a}{\"a}nniemi, Karras, Laine,
  Lehtinen, and Aila]{kynkaanniemi2019improved}
T.~Kynk{\"a}{\"a}nniemi, T.~Karras, S.~Laine, J.~Lehtinen, and T.~Aila.
\newblock Improved precision and recall metric for assessing generative models.
\newblock In \emph{Advances in Neural Information Processing Systems
  (NeurIPS)}, pages 3929--3938. Curran Associates, Inc., 2019.

\bibitem[Larsen et~al.(2016)Larsen, Kaae~S{\o}nderby, Larochelle, and
  Winther]{larsen2015}
A.~Larsen, S.~Kaae~S{\o}nderby, H.~Larochelle, and O.~Winther.
\newblock Autoencoding beyond pixels using a learned similarity metric.
\newblock In \emph{International Conference on Machine Learning (ICML)}, pages
  1558--1566, 2016.

\bibitem[Li et~al.(2018)Li, Xiao, and Ouyang]{li2018}
Y.~Li, N.~Xiao, and W.~Ouyang.
\newblock Improved boundary equilibrium generative adversarial networks.
\newblock \emph{IEEE Access}, 6:\penalty0 11342--11348, 2018.

\bibitem[Lin et~al.(2019)Lin, Thekumparampil, Fanti, and
  Oh]{DBLP:journals/corr/abs-1906-06034}
Z.~Lin, K.~K. Thekumparampil, G.~C. Fanti, and S.~Oh.
\newblock {InfoGAN-CR}: {D}isentangling generative adversarial networks with
  contrastive regularizers.
\newblock \emph{arXiv preprint arXiv:1906.06034}, 2019.

\bibitem[Locatello et~al.(2019)Locatello, Bauer, Lucic, R\"atsch, Gelly, and
  Sch\"olkopf]{locatello2018}
F.~Locatello, S.~Bauer, M.~Lucic, G.~R\"atsch, S.~Gelly, and B.~O. Sch\"olkopf,
  B.
\newblock Challenging common assumptions in the unsupervised learning of
  disentangled representations.
\newblock In \emph{Proceedings of the 36th International Conference on
  International Conference on Machine Learning (ICML)}, volume~97 of
  \emph{PMLR}, pages 4114--4124, 2019.

\bibitem[Makhzani(2018)]{DBLP:journals/corr/abs-1805-09804}
A.~Makhzani.
\newblock Implicit autoencoders.
\newblock \emph{arXiv preprint arXiv:1805.09804}, 2018.

\bibitem[Makhzani et~al.(2016)Makhzani, Shlens, Jaitly, and
  Goodfellow]{makhzani2015}
A.~Makhzani, J.~Shlens, N.~Jaitly, and I.~Goodfellow.
\newblock Adversarial autoencoders.
\newblock In \emph{International Conference on Learning Representations
  (ICLR)}, 2016.

\bibitem[Mescheder et~al.(2017)Mescheder, Nowozin, and Geiger]{mescheder2017}
L.~Mescheder, S.~Nowozin, and A.~Geiger.
\newblock Adversarial variational {B}ayes: {U}nifying variational autoencoders
  and generative adversarial networks.
\newblock In \emph{Proceedings of the 34th International Conference on Machine
  Learning (ICML)}, volume~70 of \emph{PMLR}, pages 2391--2400, 2017.

\bibitem[Miyato et~al.(2018)Miyato, Kataoka, Koyama, and Yoshida]{miyato2018}
T.~Miyato, T.~Kataoka, M.~Koyama, and Y.~Yoshida.
\newblock Spectral normalization for generative adversarial networks.
\newblock In \emph{International Conference on Learning Representations
  (ICLR)}, 2018.

\bibitem[{Puzer (GitHub user)}(2019)]{puzer:2019}
{Puzer (GitHub user)}.
\newblock {StyleGAN} encoder -- {C}onverts real images to latent space.
\newblock \url{https://github.com/Puzer/stylegan-encoder}, 2019.
\newblock GitHub repository.

\bibitem[Razavi et~al.(2019)Razavi, van~den Oord, and Vinyals]{VQVAE2}
A.~Razavi, A.~van~den Oord, and O.~Vinyals.
\newblock Generating diverse high-fidelity images with {VQ-VAE-2}.
\newblock In \emph{Advances in Neural Information Processing Systems
  (NeurIPS)}, volume~32, pages 14837--14847. Curran Associates, Inc., 2019.

\bibitem[Rezende et~al.(2014)Rezende, Mohamed, and Wierstra]{rezende2014}
D.~J. Rezende, S.~Mohamed, and D.~Wierstra.
\newblock Stochastic backpropagation and approximate inference in deep
  generative models.
\newblock In \emph{Proceedings of the 31st International Conference on Machine
  Learning (ICML)}, volume~32 of \emph{PMLR}, pages 1278--1286, 2014.

\bibitem[Rezende et~al.(2016)Rezende, Mohamed, Danihelka, Gregor, and
  Wierstra]{rezende2016one}
D.~J. Rezende, S.~Mohamed, I.~Danihelka, K.~Gregor, and D.~Wierstra.
\newblock One-shot generalization in deep generative models.
\newblock In \emph{Proceedings of the 33rd International Conference on
  International Conference on Machine Learning (ICML)}, volume~48 of
  \emph{PMLR}, pages 1521--1529, 2016.

\bibitem[Simonyan and Zisserman(2015)]{simonyan:2015}
K.~Simonyan and A.~Zisserman.
\newblock Very deep convolutional networks for large-scale image recognition.
\newblock In \emph{International Conference on Learning Representations
  (ICLR)}, 2015.

\bibitem[Tolstikhin et~al.(2018)Tolstikhin, Bousquet, Gelly, and
  Sch\"olkopf]{tolstikhin2018}
I.~Tolstikhin, O.~Bousquet, S.~Gelly, and B.~Sch\"olkopf.
\newblock Wasserstein auto-encoders.
\newblock In \emph{International Conference on Learning Representations
  (ICLR)}, 2018.

\bibitem[Ulyanov et~al.(2016)Ulyanov, Vedaldi, and Lempitsky]{UlyanovVL16}
D.~Ulyanov, A.~Vedaldi, and V.~S. Lempitsky.
\newblock Instance normalization: The missing ingredient for fast stylization.
\newblock \emph{arXiv preprint arXiv:1607.08022}, 2016.

\bibitem[Ulyanov et~al.(2018{\natexlab{a}})Ulyanov, Vedaldi, and
  Lempitsky]{ulyanov2017}
D.~Ulyanov, A.~Vedaldi, and V.~Lempitsky.
\newblock It takes (only) two: Adversarial generator-encoder networks.
\newblock In \emph{Proceedings of the Thirty-Second AAAI Conference on
  Artificial Intelligence (AAAI)}, pages 1250--1257, 2018{\natexlab{a}}.

\bibitem[Ulyanov et~al.(2018{\natexlab{b}})Ulyanov, Vedaldi, and
  Lempitsky]{ulyanovGithub}
D.~Ulyanov, A.~Vedaldi, and V.~Lempitsky.
\newblock Adversarial generator-encoder networks.
\newblock \url{https://github.com/DmitryUlyanov/AGE}, 2018{\natexlab{b}}.
\newblock GitHub repository.

\bibitem[{van den Oord} et~al.(2017){van den Oord}, Vinyals, and
  kavukcuoglu]{NIPS2017_7210}
A.~{van den Oord}, O.~Vinyals, and k.~kavukcuoglu.
\newblock Neural discrete representation learning.
\newblock In \emph{Advances in Neural Information Processing Systems (NIPS)},
  volume~30, pages 6306--6315. Curran Associates, Inc., 2017.

\bibitem[Yu et~al.(2015)Yu, Seff, Zhang, Song, Funkhouser, and Xiao]{Yu:2015}
F.~Yu, A.~Seff, Y.~Zhang, S.~Song, T.~Funkhouser, and J.~Xiao.
\newblock {LSUN}: {C}onstruction of a large-scale image dataset using deep
  learning with humans in the loop.
\newblock \emph{arXiv preprint arXiv:1506.03365}, 2015.

\bibitem[Zamir et~al.(2017)Zamir, Wu, Sun, Shen, Shi, Malik, and
  Savarese]{zamir2017}
A.~R. Zamir, T.~Wu, L.~Sun, W.~B. Shen, B.~E. Shi, J.~Malik, and S.~Savarese.
\newblock Feedback networks.
\newblock In \emph{Proceedings of the IEEE Conference on Computer Vision and
  Pattern Recognition (CVPR)}, pages 1808--1817, 2017.

\bibitem[Zhang et~al.(2018)Zhang, Isola, Efros, Shechtman, and Wang]{zhang2018}
R.~Zhang, P.~Isola, A.~A. Efros, E.~Shechtman, and O.~Wang.
\newblock The unreasonable effectiveness of deep features as a perceptual
  metric.
\newblock In \emph{Proceedings of the IEEE Conference on Computer Vision and
  Pattern Recognition (CVPR)}, pages 586--595, 2018.

\end{thebibliography}
\endgroup

\clearpage\appendix
\setcounter{section}{0}
\nipstitle{{Supplementary Material for} \\ Deep Automodulators}
\pagestyle{empty}

In the appendix, we include further details underlying the model and the experiments and complement the results in the main paper with examples and more comprehensive results. We start with the details of training and evaluation (\cref{sec:trainingdetails}), complemented by detailed description of architecture and data flow in the network (\cref{sec:archdetail}). We then show a comparison of scale-specific attribute modification with the regular one, providing more context to the quick qualitative experiment in \cref{sec:exper}. We proceed with showing more random samples (\cref{sec:rands}) and reconstructions (\cref{sec:recos2}). Note that there are reconstructions in the diagonals of all style-mixture images, too. Importantly, we show systematic style-mixture examples in \cref{sec:stylemixint}, corresponding to \cref{fig:stylegan-mixing} but with {\textit{real}} (unseen) input images from the FFHQ test set. We follow with showing latent space interpolations at all scales between real input images (which also could be done on a scale-specific basis). We then continue with an experiment regarding the inversion of StyleGAN \citet{karras2018style} with an optimization process, and finish with an experiment focused on conditional sampling, in which certain scales of an input image are fixed in the reconstruction images but other scales are randomly sampled over, creating variations of the same input face.

\begin{figure*}[!b]
  \centering

    \tikzstyle{block} = [rounded corners=0.5pt,minimum width=1.5mm,minimum height=1.5mm,inner sep=0,draw=black!50!black,fill=white]

    \newcommand{\encoder}[4]{%
      \foreach \j [count=\i] in {4,8,16,32,64,128,256} 
        \node[block,minimum width={0.25cm+0.2cm*\i}] (#4-\i) at (#1,#2-.3*\i) {\scalebox{.5}{$\j{\times}\j$}};
    }   
  \hspace{\fill}
  \begin{subfigure}[b]{.28\textwidth}
    \centering
    \scalebox{0.9}{%
    \begin{tikzpicture}[inner sep=0]

    \node [minimum width=3mm,minimum height=3mm] (zA) at (1.5,.25) {$\vz$};

    \node [minimum width=3mm] (xi0) at (3,.25) {$\vxi^{(0)}$};

    \node at (0,-3.2) (in) {\includegraphics[width=1.5cm]{./fig/sysmix8/42420N2/source-5.jpg}};
    \node[minimum width=7mm,minimum height=3.5mm,fill=white,opacity=1,draw=black,rounded corners=0.5pt] at (0,-3.2-.75) {};
    \node at (0,-3.2-.75) {$\vphantom{\hat{\vx}}\vx$};

    \node (out) at (3,-3.2) {\includegraphics[width=1.5cm]{./fig/sysmix8/42420N2/mix-5_5_0--1.jpg}};
    \node[minimum width=7mm,minimum height=3.5mm,fill=white,opacity=1,draw=black,rounded corners=0.5pt] at (3,-3.2-.75) {};
    \node at (3,-3.2-.75) {$\hat{\vx}$};

    \draw [->, rounded corners=3mm, thick] (in) |- (zA);
    \draw [->, rounded corners=3mm, thick] (xi0) -- (out);

    \encoder{0}{0}{4}{enc}
    \encoder{3}{0}{4}{dec}

    \draw [->, rounded corners=3mm, thick] (zA) |- (dec-1);
    \draw [->, rounded corners=3mm, thick] (zA) |- (dec-2);
    \draw [->, rounded corners=3mm, thick] (zA) |- (dec-3);
    \draw [->, rounded corners=3mm, thick] (zA) |- (dec-4);
    \draw [->, rounded corners=3mm, thick] (zA) |- (dec-5);
    \draw [->, rounded corners=3mm, thick] (zA) |- (dec-6);
    \draw [->, rounded corners=3mm, thick] (zA) |- (dec-7);

    \node[rotate=90] at (-.75,-.5) {\scriptsize Encoder, $\vphi$};
    \node[rotate=-90] at (3.75,-.5) {\scriptsize Decoder, $\vtheta$};
    \node[above of=zA,yshift=-3mm,text width=1cm, align=center] {\scriptsize Latent\\[-3pt] encoding};        
    \node[above of=xi0,yshift=-3mm,text width=1cm, align=center] {\scriptsize Canvas\\[-3pt] variable};

    \foreach \i in {1,2,3,4,5,6,7} 
      \node[block,minimum width={0.2},shape=circle,draw=black,fill=white] at (1.9,-0.3*\i) {};            
            
    \end{tikzpicture}}
    \caption{Top-level}   
    \label{fig:archd}     
  \end{subfigure}
  \hspace{\fill}
  \begin{subfigure}[b]{.35\textwidth}
    \centering  
    \hspace*{0cm}
    \scalebox{0.9}{%
    \includegraphics[width=6cm]{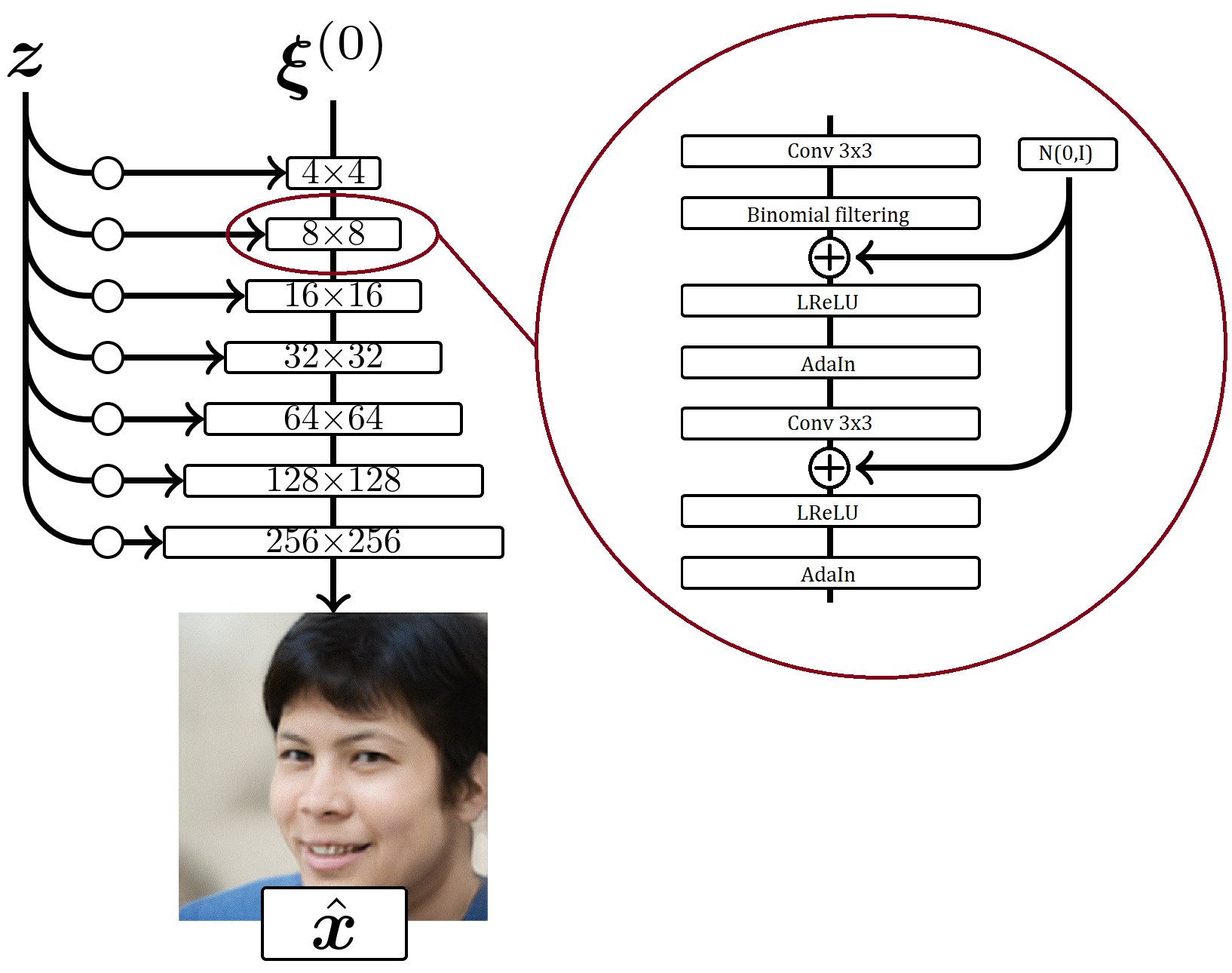}
    }
    \caption{Decoder block structure}
    \label{fig:decoderd}     
  \end{subfigure}
  \hspace{\fill}
  \begin{subfigure}[b]{.25\textwidth}
    \centering  
    \hspace*{0cm}
    \scalebox{0.9}{%
    \includegraphics[width=\textwidth]{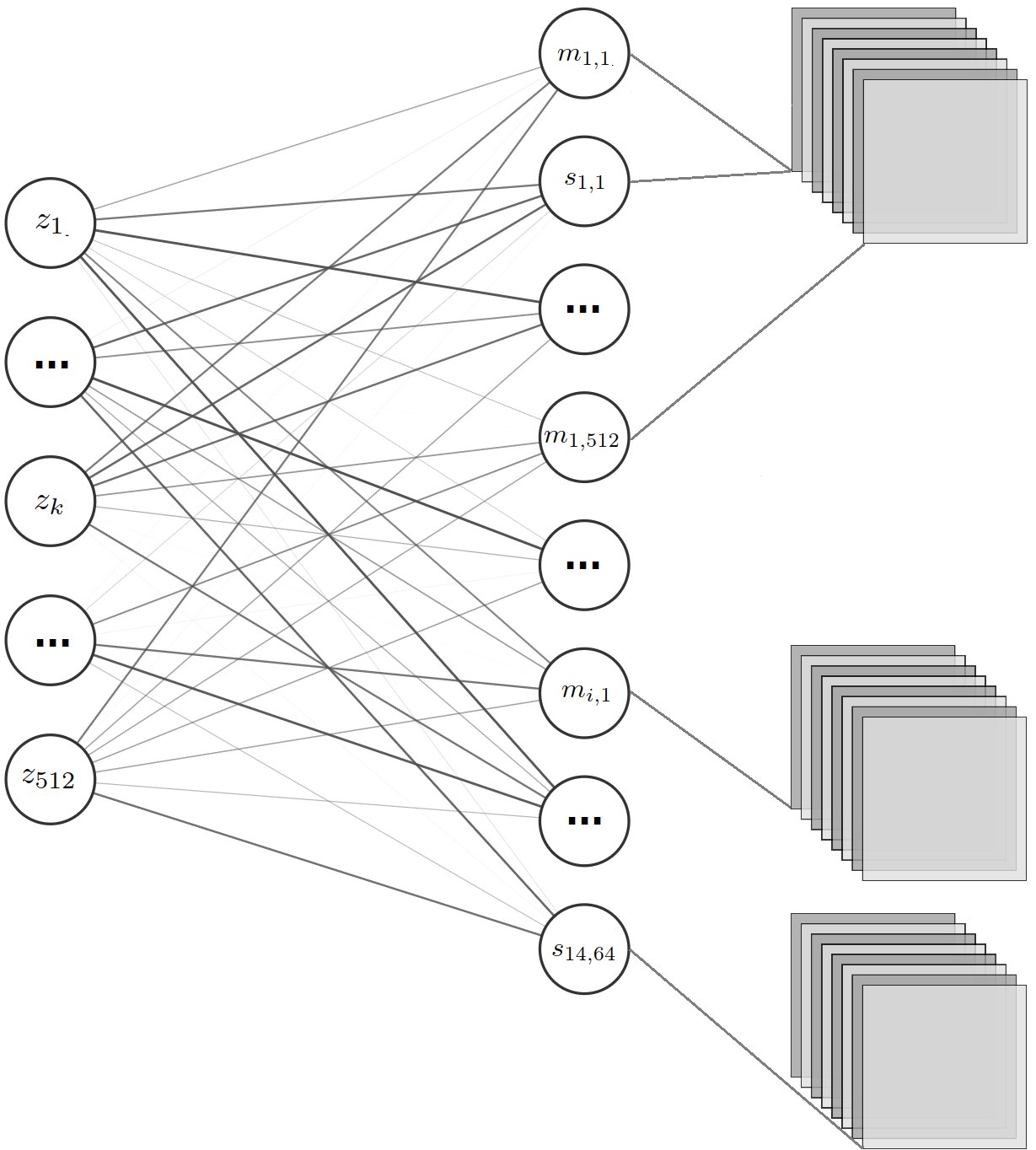}
    }
    \caption{Latent vector to modulation}
    \label{fig:latentmapd}
  \end{subfigure}
  \hspace{\fill}  
  \caption{(a)~Top-level view, where a single decoder block corresponds to a specific resolution. (b)~A single decoder block contains two convolutional layers and other repeating components. The noise is added to each channel of the layer using a single scale parameter per layer, \ie, a different random value is added across each activation map, but the scale is the same for all maps in the same layer. (c)~The latent codes are connected to the modulation scalar ($m$,$s$) pair of every activation map of each of the $7 {\times} 2$ convolutional layers. In the encoder, the number of channels in the convolutional blocks follows \citep{heljakka2019towards}
  as 64,128,256,512,512,512,512. In $256{\times}256$ \CelebAHQ, the decoder channels are the symmetric inverse: 512, 512, 512, 512, 256, 128, 64. In other datasets, it was beneficial to double the number of feature maps in high resolutions of the decoder, with the number of channels as: 512, 512, 512, 512, 512, 256, 128 for $256{\times}256$ datasets. For $512{\times}512$ FFHQ a final 64-channel block was added to this, resulting in 512, 512, 512, 512, 256, 128, 64.}

\end{figure*}

\section{Training Details}
\label{sec:trainingdetails}
\begin{figure*}[t!]
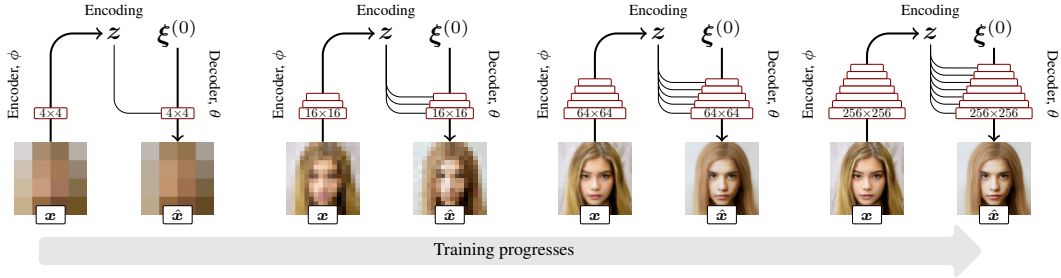

  \centering
  \resizebox{\textwidth}{!}{%
  \begin{tikzpicture}[inner sep=0]

    \newcommand{\encoder}[5]{%
      \tikzstyle{block} = [rounded corners=0.5pt,minimum width=1.5mm,minimum height=1.5mm,inner sep=0,draw=red!50!black,fill=white]
      \foreach \jj in {1,...,#3} {
        \ifnum \jj<#3
        \foreach \ii in {1,2}
          \node[block,minimum height=1mm,minimum width={0.25cm+0.2cm*\jj-0.1cm+0.1cm*\ii}] (#5-#4-\jj-\ii) at (#1,#2-2mm*\jj-1mm*\ii+1.25mm+#3*2mm-1mm) {};
        \fi}

      \node[block,minimum width={0.25cm+0.2cm*#3}] (#5-#4) at (#1,#2-1mm) {\scalebox{.5}{$#4{\times}#4$}};}

    \foreach \res [count=\i] in {4,16,64,256} {

      \node [minimum width=3mm,minimum height=3mm] (z-\res) at (3.75*\i+0.8750,1) {$\vz$};
      \node [minimum width=3mm,minimum height=3mm] (xi-\res) at (3.75*\i+1.75,1) {$\vxi^{(0)}$};

      \node[minimum width=1cm,minimum height=1cm,fill=gray!20] at (3.75*\i,-1) {\includegraphics[width=1cm]{fig/architecture/input-\res.jpg}};
      \node[minimum width=1cm,minimum height=1cm,fill=gray!20] at (3.75*\i+1.75,-1) {\includegraphics[width=1cm]{fig/architecture/output-\res.jpg}};

      \node[minimum width=1cm,minimum height=1cm] (in-\res) at (3.75*\i,-1) {};
      \node[minimum width=1cm,minimum height=1cm] (out-\res) at (3.75*\i+1.75,-1) {};

      \node[minimum width=4mm,minimum height=2.5mm,fill=white,opacity=1,draw=black,rounded corners=0.5pt] at (3.75*\i,-1.5) {};
      \node at (3.75*\i,-1.5) {\tiny $\vphantom{\hat{\vx}}\vx$};

      \node[minimum width=4mm,minimum height=2.5mm,fill=white,opacity=1,draw=black,rounded corners=0.5pt] at (3.75*\i+1.75,-1.5) {};
      \node at (3.75*\i+1.75,-1.5) {\tiny $\hat{\vx}$};

      \draw [->, rounded corners=3mm, thick] (in-\res) |- (z-\res);
      \draw [->, rounded corners=3mm, thick] (xi-\res) -- (out-\res);

      \encoder{3.75*\i}{0}{\i}{\res}{enc}
      \encoder{3.75*\i+1.75}{0}{\i}{\res}{dec}

      \foreach \j in {1,...,\i} {
        \ifnum \j<\i
          \draw [-, rounded corners=2mm] (z-\res) |- (dec-\res-\j-1);
          \draw [-, rounded corners=2mm] (z-\res) |- (dec-\res-\j-2);
        \fi        
      }
      \draw [-, rounded corners=2mm] (z-\res) |- (dec-\res);

      \node [rotate=90] at (3.75*\i-0.4-0.1*\i,0.3) {\tiny Encoder, $\phi$};
      \node [rotate=-90] at (3.75*\i+1.75+0.4+0.1*\i,0.3) {\tiny Decoder, $\theta$};   
      \node [rotate=0] at (3.75*\i+0.8750,1.3) {\tiny Encoding};               
    }

    \tikzstyle{arrow} = [draw=black!10, single arrow, minimum height=130mm, minimum width=5mm, single arrow head extend=2mm, fill=black!10, anchor=center, rotate=0, inner sep=2pt, rounded corners=2pt]    
    \node[arrow] at (10,-2) {\scriptsize Training progresses};
    
  \end{tikzpicture}}   
  \vspace*{-16pt}
  \caption{The model grows step-wise during training; the resolution doubles on every step. Input $\vx$ is encoded into a latent encoding $\vz$ (a dimensionality of 512 used throughout this paper). The decoder acts by modulating an empty canvas $\vxi^{(0)}$ by the latent encoding and produces the output $\hat{\vx}$. Further explanation of the model architecture is provided in \cref{fig:arch}.}
  \label{fig:architecture}
  \vspace*{-1em}
\end{figure*}

The training method largely follows the Balanced \PIONEER \citep{heljakka2019towards}, with progressively growing encoder and decoder with symmetric high-level structure (\cref{fig:archd}), and decreasing the batch size when moving to higher resolutions (\cref{fig:architecture}). The encoder and decoder consist of 7 blocks each containing two residual blocks with a $3 {\times} 3$ filter. In both blocks of the encoder, the convolutions are followed by a spectral normalization operation \citep{miyato2018} and activation (conv - spectral norm - act). In the first block of the decoder, they are followed by binomial filtering, layer noise, activation and AdaIn normalization (conv - binomial - noise - act - AdaIn); in the second block of decoder, by layer noise, activation and AdaIn normalization (conv - noise - act - AdaIn). A leaky ReLU ($p=0.2$) is used as the activation. Equalized learning rate \citep{karras2017} is used for decoder convolutions. In the encoder, each block halves the resolution of the convolution map, while in the decoder, each block doubles it. The output of the final encoder layers is flattened into a 512-dimensional latent block. As in StyleGAN \citep{karras2018style}, the block is mapped by affine mapping layers so that each convolutional layer $C$ in the decoder block is preceded by its own fully connected layer that maps the latent to two vectors each of length $N$, when $N$ equals the number of channels in $C$.

Each resolution phase until $32 {\times} 32$, for all data sets, uses a learning rate $\alpha=0.0005$ and thereafter $0.001$. Optimization is done with ADAM ($\beta_1=0, \beta_2=0.99, \epsilon=10^{-8}$). KL margin is 2.0 for the first two resolution steps, and therafter fixed to 0.5, except for \CelebAHQ, for which it is switched to 0.2 at $128 {\times} 128$ and 0.4 at $256 {\times} 256$, and for \LSUN Bedrooms, for which the margin was 0.2 from $128 {\times} 128$ upwards. We believe that 0.5 for low resolutions and 0.2 thereafter would work sufficiently across all these data sets. Note that unlike in \citep{heljakka2019towards}, we use only one generator training step for each individual encoder training step.
The length of each training phase amounts to 2.4M training samples until $64 {\times} 64$ resolution phase, which lasts for 10.4M samples (totaling in 20.0M). For FFHQ, the $128 {\times} 128$ phase uses 10.6M samples while \CelebAHQ and \LSUN Cars use 7.5M samples, and \LSUN Bedrooms uses 4.8M samples. For FFHQ, the $256 {\times} 256$ phase uses 5.0M samples, \CelebAHQ uses 4.5M, \LSUN Bedrooms 2.9M samples and \LSUN Cars 2M samples. Then training with FFHQ up to $512 {\times} 512$, this final phase uses 6.7M samples. The training of the final stage was generally cut off when reasonable FID results had been obtained. More training and learning rate optimization would likely improve results. With two NVIDIA Titan V100 GPUs, the training times were 10~days for \CelebAHQ, 10.5~days for FFHQ $256 {\times} 256$ and (total) 22.5~days for FFHQ $512 {\times} 512$, for \LSUN Bedrooms 18.5~days, and for Cars 18~days. 3 evaluation runs with different seeds were done for \CelebAHQ (separately for each configuration of the ablation study, including the full loss with and without layer noise), 3 for FFHQ, 3 for \LSUN Bedrooms and 3 for \LSUN Cars (1 with and 2 without layer noise). Some runs shared the same pretrained network up to $64 {\times} 64$ resolution (except in Ablation study, where each run was done from scratch).

For evaluating the model after training, a moving exponential running average of generator weights \citep{karras2017,heljakka2019towards} was used. For visual evaluation, the layer noise can be turned off, often yielding slightly more polished-looking results.
For all data sets, training/test set splits were as follows: 60k/10k for FFHQ (download at \url{https://github.com/NVlabs/ffhq-dataset}), 27k/3k split for \CelebAHQ (download with instructions at \url{https://github.com/tkarras/progressive_growing_of_gans}), 4,968,695/552,061 for \LSUN Cars (download at \url{https://github.com/fyu/lsun}), and 3033042/300 for \LSUN Bedrooms (download at \url{https://github.com/fyu/lsun}). Note that in regular GAN training, complete data sets are often used without train/test split, yielding larger effective training sets. For instance, in FFHQ, we train on the actual 60k training images only, whereas StyleGAN trained on all 70k. For FFHQ and \CelebAHQ, cropping and alignment of the faces should be performed exactly as described by the authors of the data sets as referred to above, which also direct to the readily available scripts for the alignments (based on facial keypoint detection). For \LSUN images, there was no preprocessing except cropping the Cars to $256 {\times} 256$. Mirror augmentation was used in training the face data sets, but not for training the \LSUN data sets (for comparison with prior work).

For baselines in \cref{tbl:results} and \cref{tbl:results2}, we used pre-trained models for StyleGAN, PGGAN, \PIONEER, and GLOW with default settings provided by the authors, except Balanced \PIONEER for FFHQ which we trained. FID of PGGAN for Cars and Bedrooms is from \citet{karras2017}, whereas FID of FFHQ is from \citet{karras2018style} and FID of \CelebAHQ we computed for $256 {\times} 256$ separately.
 We trained the VAE and WAE models manually. StyleGAN FID for \LSUN Bedrooms is from \citet{karras2018style} whereas the other FIDs were calculated for $256 {\times} 256$ separately. PPLs for StyleGAN and PGGAN for FFHQ come from \citet{karras2018style} while the PPL for StyleGAN v2 is from \citet{2019arXiv191204958K}, PPL for PGGAN \CelebAHQ from \citet{heljakka2019towards} and PPL for \CelebAHQ of StyleGAN was computed from the pretrained model.
 For all VAE baselines the weight for KL divergence loss term was 0.005. For all WAE baseline, we used the WAE-MMD algorithm. The weight of the MMD loss term with automodular architecture (WAE-AdaIn) was four and with Balanced \PIONEER (WAE-classic) architecture it was two. For VAEs, the learning rate for the encoder was 0.0001, and for the generator 0.0005. For WAEs, the learning rate for both was 0.0002. We trained Balanced \PIONEER for FFHQ by otherwise using the \CelebAHQ hyperparameters, but increasing the length of the $64 {\times} 64$ and $128 {\times} 128$ pretraining stages proportionally to the larger training set size (60k vs. 27k), training the former up to 20.04M samples and the latter to 27.86M samples, followed by the $256 {\times} 256$ stage, which was trained up to 35.4M samples, after which we observed no further improvement. (With shorter pre-training stages, the model training did not converge properly.) Note: Some apparent discrepancies between reported FID results between papers are often explained by different resolutions. In \cref{tbl:results2} we have used $256 {\times} 256$ resolution.

For evaluating the encoding and decoding performance, we used 10k unseen test images from the FFHQ data set, cropped the input and reconstruction to $128{\times}128$ as in \citet{karras2018style} and evaluated the LPIPS distance between the inputs and reconstructions. We evaluated 50k random samples in all data sets and compare against the provided training set. The GLOW model has not been shown to work with $256 {\times} 256$ resolution on \LSUN Bedrooms nor Cars (the authors show qualitative result only for $128 {\times} 128$ for Bedrooms).

For Perceptual Path Length (PPL), we repeatedly take a random vector of length $\varepsilon = 10^{-4}$ in the latent space, generate images at its endpoints, crop them around mid-face to $128{\times}128$ or $64{\times}64$, and measure the LPIPS between them \citep{karras2018style}. PPL equals the scaled expectation of this value (for a sample of 100k vectors).

\paragraph{Hyperparameter selection} The driving principle to select hyperparameters in this paper was to use the same values as \citet{heljakka2019towards} whenever necessary, and minimize variation across data sets, so as to show generalization rather than tuning for maximum performance in each data set. The learning rate was attempted at the same rate as in \citep{heljakka2019towards} ($\alpha = 0.001$) for the whole length of training. However, the pre-training stages up to $32 {\times} 32$ appeared unstable, hence $\alpha = 0.0005$ was attempted and found more stable for those stages. Margin values 0.2, 0.4 and 0.5 were attempted for training stages from $128 {\times} 128$ upwards for FFHQ, \CelebAHQ and \LSUN Bedrooms. However, we did not systematically try out all possible combinations, but rather started from the values used in \citet{heljakka2019towards} and only tried other values if performance seemed insufficient. For the length of the $128 {\times} 128$ training stage, separately for each data set, we first tried a long training session (up to 10M seen samples) and observed whether FID values were improving. We selected the cutoff point to $256 {\times} 256$ for each data set based on approximately when the FID no longer seemed to improve for the lower resolution. The $256 {\times} 256$ phase was then trained until FID no longer seemed to improve, or, in the ablation study, we decided to run for the fixed 40M seen samples. For the $\lambda_\mathcal{X}$ in the image space reconstruction loss, we tried values 1, 0.5 and 0.2, of which 0.2 appeared to best retain the same training dynamics as the L1 loss in \CelebAHQ, and was hence used for all experiments. Other hyperparameters not mentioned here follow the values and reasoning of \citet{heljakka2019towards} by default.

\section{Detailed Explanation of Automodulation Architecture}
\label{sec:archdetail}

As the structure of the proposed decoder is rather unorthodox (though very similar to \citet{karras2018style}) and the modulation step especially is easy to misunderstand, we now explain the workings of the decoder step-by-step in detail.

\paragraph{Encoder} The encoder works in the same way as any convolutional image encoder, where the last convolution block maps the highest-level features of the input image $\vx$ into a single 512-dimensional vector $\vz$. To understand the concept of latent mixing, we can immediately consider having two samples $\vx_1$ and $\vx_2$ which are mapped into $\vz_1$ and $\vz_2$. In reality, we will use minibatches in the regular way when we train the decoder, but for the purposes of this explanation, let us assume that our batch has only these 2 invidiual samples. Thus our whole latent vector is of size [2, 512]. Each latent vector is independently normalized to unit hypersphere, \ie~to reside within [-1, 1].

\paragraph{Decoder high-level structure} Corresponding to the 7 levels of image resolutions (from 4x4 to 256x256), the decoder comprises of 7 high-level blocks (\cref{fig:archd}). Each such block has an internal structure as depicted in \cref{fig:decoderd}.
In order to understand the modulation itself, the individual activation map of a convolutional layer is the relevant level of abstraction.

\paragraph{Activation maps} As usual, each deconvolutional operation produces a single activation map per channel, hence for a single deconvolutional layer (of which there are 2 per block), there can be \eg~512 such maps (\ie, 1024 per block). We now proceed to modulate the mean and variance of each of those 512 maps separately, so that each map has two scalar values for that purpose. In other words, there will be 512 + 512 scalars to express the new channel-wise mean and variance for the single decoder layer. As in \cref{eq:adain}, the activations of each map are separately scaled to exactly amount to the new mean and variance. Note that those statistics pertain only within each map, {\textit{not}} across all the maps in that layer.

\paragraph{Connecting the latent to the scaling factors} In order to drive the modulating scalars with the original 512-dimensional latent vector $z$, we take add a fully connected linear layer that connects the latent vector to each and every modulating scalar, yielding $512 {\times} 2 {\times} N$ connections where N is the total number of activation maps in the full decoder (\cref{fig:latentmapd}). Note that this linear layer is not affected at all by the way in which the decoder is structured; it only looks at the latent vector and each convolutional activation map.

\paragraph{Initiating the data flow with the constant inputs} Hence, given a latent vector, one can start decoding. The inputs to the first deconvolutional operations at the top of the decoder are constant values of 1. This apparently counter-intuitive approach is actually nothing special. Consider that were the latent code simply connected to the first deconvolutional layers with weight $1$ to create the mean and with weight $0$ for the variance, this would essentially be the same as driving the first layer directly with the latent code, as in the traditional image decoder architecture.

\paragraph{The data flow} Now, as the image content flows through the decoder blocks, each operation occurs as in regular decoders, except for the modulation step. Whenever an activation function is computed, a separate modulation operation will follow, with unique scaling factors. For the downstream operations, this modulation step is invisible, since it merely scaled the statistics of those activations. Then, the data flow continues in the same way, through each block, until at the last step, the content is mapped to a 3-dimensional grid with the size of the image, \ie, our final generated image.

\begin{figure*}[t!]
  \centering\scriptsize
  \setlength{\figurewidth}{.1\textwidth}
  \setlength{\figureheight}{\figurewidth}  
  \begin{subfigure}[t]{0.5\textwidth}
    \centering

    \begin{tikzpicture}[inner sep=0]
      \node[] at ({-1*\figurewidth},{-1.1\figureheight}) {\includegraphics[width=\figurewidth]{./fig/attribute/68522.jpg}};
      \node[] at ({-1*\figurewidth},{-2.1\figureheight}) {\includegraphics[width=\figurewidth]{./fig/attribute/68522.jpg}};
      
    \foreach \x [count=\i] in {1, 2, 3} {
      \node[] at ({\figurewidth*\i-1*\figurewidth},{-1.1\figureheight}) {\includegraphics[width=\figurewidth]{./fig/attribute/\x/B1.jpg}};
      \node[] at ({\figurewidth*\i-1*\figurewidth},{-2.1\figureheight}) {\includegraphics[width=\figurewidth]{./fig/attribute/\x/B2.jpg}};}

  \foreach \x [count=\i] in {Input, No smile, Fe/male, Glasses} {
    \node[fill=white,inner sep=0, minimum width=4mm] at ({\figurewidth*\i-2*\figurewidth},-0.35\figureheight) {\bf \x};
  }

     \node[align=center] at (-2.15*\figurewidth,-1.1*\figureheight) {\bf Scale- \\ \bf specific $\rightarrow$};
     \node[align=center] at (-2.15*\figurewidth,-2.1*\figureheight) {\bf Regular $\rightarrow$};

  \end{tikzpicture}

    \label{fig:attribs}
\end{subfigure}
\begin{subfigure}[t]{0.45\textwidth}
      \centering

      \begin{tikzpicture}[inner sep=0]
        \node[] at ({-1*\figurewidth},{-1.1\figureheight}) {\includegraphics[width=\figurewidth]{./fig/fig_rotation_invariance/128/source-2B.jpg}};
        \node[] at ({-1*\figurewidth},{-2.1\figureheight}) {\includegraphics[width=\figurewidth]{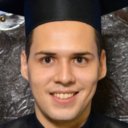}};
        
      \foreach \x [count=\i] in {1, 2, 3} {
        \node[] at ({\figurewidth*\i-1*\figurewidth},{-1.1\figureheight}) {\includegraphics[width=\figurewidth]{./fig/attribute/\x/1.jpg}};
        \node[] at ({\figurewidth*\i-1*\figurewidth},{-2.1\figureheight}) {\includegraphics[width=\figurewidth]{./fig/attribute/\x/2.jpg}};}

    \foreach \x [count=\i] in {Input, No smile, Fe/male, Glasses} {
      \node[fill=white,inner sep=0, minimum width=4mm] at ({\figurewidth*\i-2*\figurewidth},-0.35\figureheight) {\bf \x};
    }

    \end{tikzpicture}

      \label{fig:attribs2}
  \end{subfigure}
  \caption{Modifying an attribute in latent space by using only 4 exemplar images of it, for two unseen test images. In `regular' all-scales manipulation (bottom row), the variance in the exemplars causes unwanted changes in, \eg, texture, face size and pose. When the latent vector only drives the relevant scales, the variance in other scales is inconsequential (top row).}
  \vspace*{-1em}
\end{figure*}

\section{Scale-specific attribute modification}
\label{sec:scalemod1}

Interesting attributes of a previously unseen input image can be modified by altering its latent code in the direction of the attribute. The direction can be found by taking N image samples with the attribute and N without it, encoding the images, and taking the difference between the mean encodings. Scaled as desired, the resulting attribute latent vector can then be added to the latent code of a new unseen input image

The quality of the attribute vector depends on the selected exemplars (and, obviously, on the encoder). Given that all the exemplars have (or, for the opposite case, lack) the attribute A and are randomly drawn from a balanced distribution, then, as N increases (\eg, to $N=64$), all other feature variation embedded in the latent vector except for the attribute should cancel out. However, for small N (\eg, $N=4$), this does not happen, and the latent vector will be noisy. For the example in \cref{fig:attribs}, we now show the difference between applying such a vector on all scales as usually done (\eg, in \citep{heljakka2019towards}) in architectures that do not allow latent modulation, and applying it only on the layers that correspond to the scales where we expect the attribute to have an effect (\cref{fig:attribs2}). Here, we simply determine the range of layers manually, as the $16{\times}16 - 64{\times}64$ for the smile on/off transform, $8 {\times} 8 - 64{\times}64$ layers for male-to-female, and $4{\times}4 - 8{\times}8$ for glasses. The effect of the noise in the attribute-coding vector is greatly reduced, since most of the scales simply are not touched by it.

Note that while autoencoder-like models can directly infer the latents from real exemplar images, in GANs without an encoder, you must take the reverse and more tedious route: the formation of latent vectors needs to take place by picking the desired attribute from randomly generated samples, presuming that it eventually appears in a sufficient number.

\section{Random Samples}
\label{sec:rands}
\begin{figure*}[!t]
  \centering\scriptsize
  \setlength{\figurewidth}{.142\textwidth}
  \setlength{\figureheight}{\figurewidth}
  \begin{subfigure}{\textwidth}
  \begin{tikzpicture}[inner sep=0]

    \foreach \x [count=\i] in {12,...,25}
      \node[draw=white,fill=black!20,minimum size=\figurewidth,inner sep=0pt]
        (\i) at ({\figurewidth*mod(\i-1,7)},{-\figureheight*int((\i-1)/7)})
        {\includegraphics[width=\figurewidth]{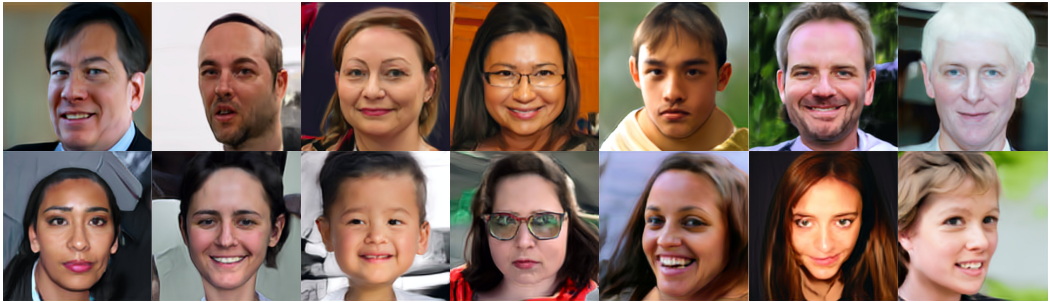}};
  \end{tikzpicture}
  \caption{FFHQ $512 {\times} 512$}
  \label{fig:randomffhq}  
  \end{subfigure}\\[1em]
  \begin{subfigure}{\textwidth}
  \centering\scriptsize
  \setlength{\figurewidth}{.142\textwidth}
  \setlength{\figureheight}{\figurewidth}
  \begin{tikzpicture}[inner sep=0]

    \foreach \x [count=\i] in {0,...,13}
      \node[draw=white,fill=black!20,minimum size=\figurewidth,inner sep=0pt]
        (\i) at ({\figurewidth*mod(\i-1,7)},{-\figureheight*int((\i-1)/7)})
        {\includegraphics[width=\figurewidth]{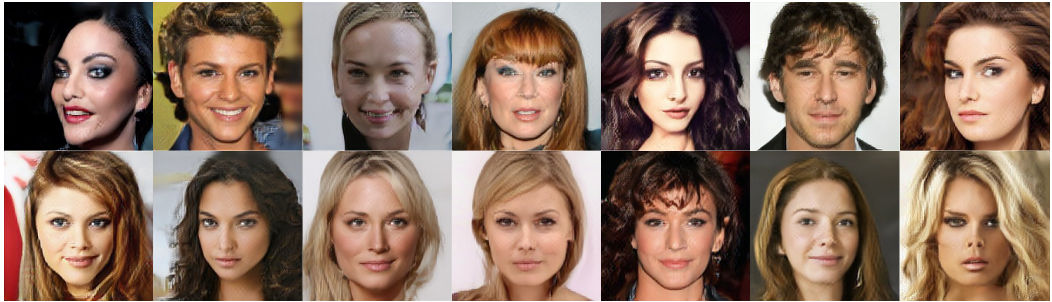}};
  \end{tikzpicture}
  \caption{\CelebAHQ $256 {\times} 256$}
  \label{fig:randomcahq}  
  \end{subfigure}  
  \caption{Uncurated random samples for an automodulator trained on FFHQ and \CelebAHQ, respectively.}
\end{figure*}

\begin{figure*}[!t]
  \centering\scriptsize
  \setlength{\figurewidth}{.2\columnwidth}
  \setlength{\figureheight}{\figurewidth}
  \begin{subfigure}{\columnwidth}
    \centering
    \begin{tikzpicture}[inner sep=0]
      \foreach \x [count=\i] in {15,...,29}
        \node[draw=white,fill=black!20,minimum size=\figurewidth,inner sep=0pt]
          (\i) at ({\figurewidth*mod(\i-1,5)},{-\figureheight*int((\i-1)/5)})
          {\includegraphics[width=\figurewidth]{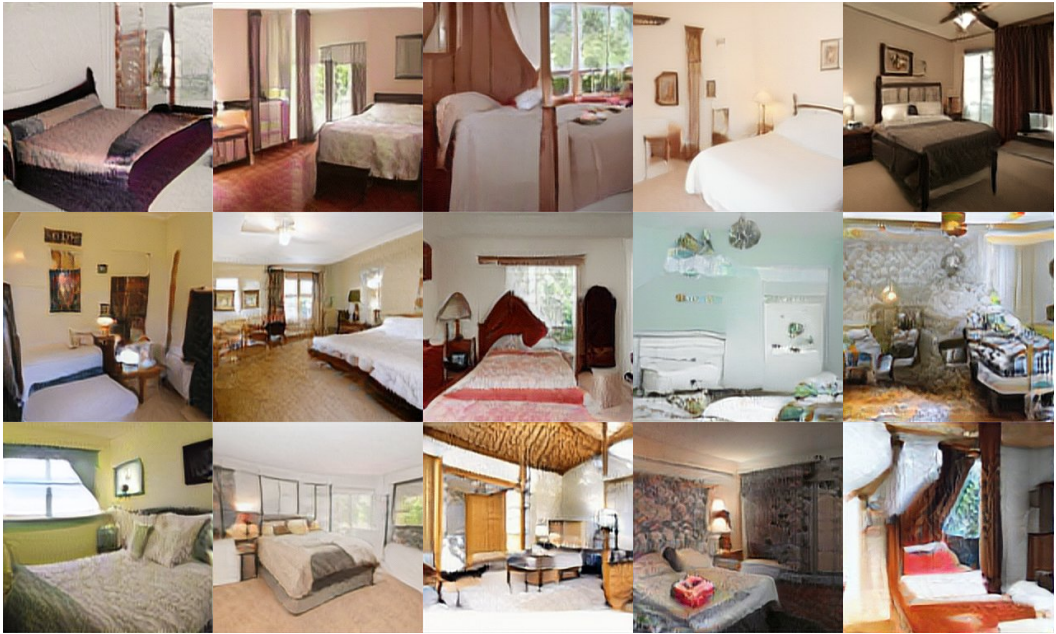}};
    \end{tikzpicture}
    \caption{{\LSUN Bedrooms}}
  \end{subfigure}
  \hfill
  \begin{subfigure}{\columnwidth}
    \centering
    \begin{tikzpicture}[inner sep=0]
      \foreach \x [count=\i] in {12,...,26}
        \node[draw=white,fill=black!20,minimum size=\figurewidth,inner sep=0pt]
          (\i) at ({\figurewidth*mod(\i-1,5)},{-\figureheight*int((\i-1)/5)})
          {\includegraphics[width=\figurewidth]{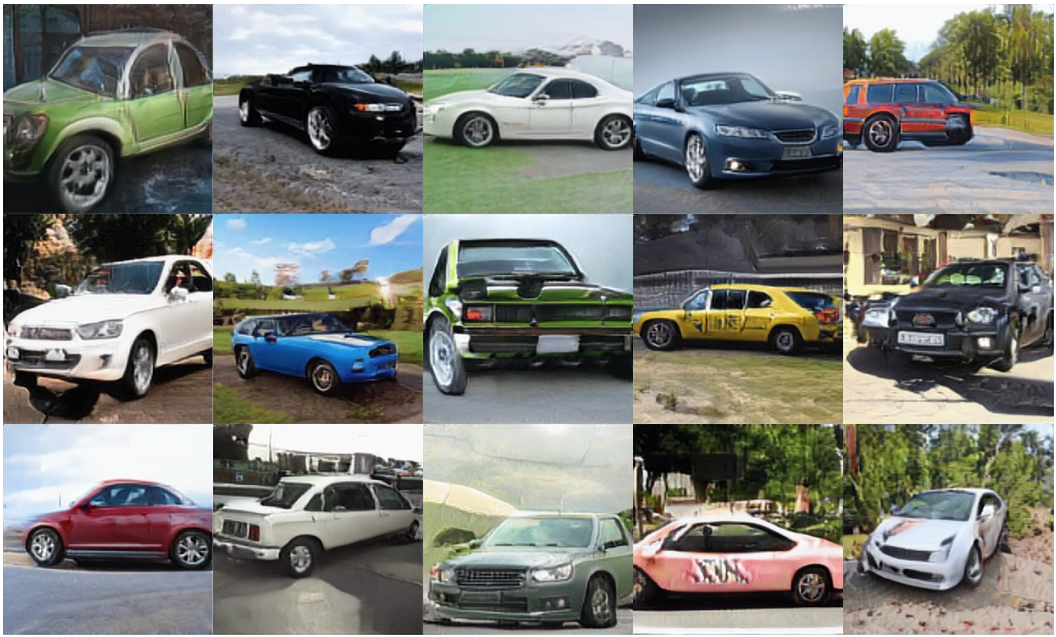}};
    \end{tikzpicture}
    \caption{{\LSUN Cars}}
  \end{subfigure}  
  \caption{Additional samples from an automodulator trained on \LSUN Bedrooms and Cars a resolution of at $256 {\times} 256$.}
  \label{fig:randomsamples}  
\end{figure*}

Our model is capable of fully random sampling by specifying $\vz \sim \mathrm{N}(\vzero,\MI)$ to be drawn from a unit Gaussian. \cref{fig:randomffhq,fig:randomcahq,fig:randomsamples} show samples from an automodulator trained with the FFHQ/\CelebAHQ/\LSUN data sets up to resolution $256 {\times} 256$.

\section{Reconstructions}
\label{sec:recos2}
\begin{figure*}[!t]
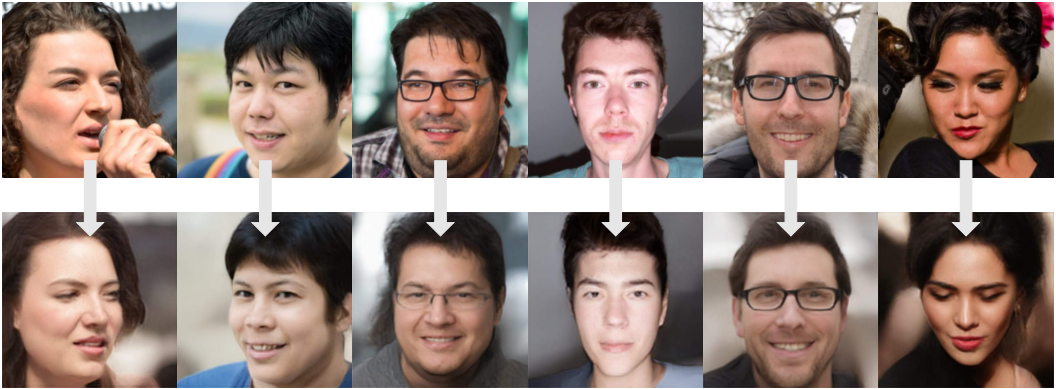

  \centering\footnotesize
  \setlength{\figurewidth}{.1667\textwidth}
  \setlength{\figureheight}{\figurewidth}
  \begin{tikzpicture}[inner sep=0]

  \tikzstyle{arrow} = [draw=black!10, single arrow, minimum height=10mm, minimum width=3mm, single arrow head extend=1mm, fill=black!10, anchor=center, rotate=-90, inner sep=2pt]

  \foreach \x [count=\i] in {0,...,5} {
     \node[] at ({\figurewidth*\i},{0*\figureheight}) {\includegraphics[width=\figurewidth]{./fig/reconstructions/ffhq_42420N/\x_orig.jpg}};
     \node[] at ({\figurewidth*\i},{-1.2\figureheight}) {\includegraphics[width=\figurewidth]{./fig/reconstructions/ffhq_42420N/\x_pine.jpg}};
     \node[arrow] at ({\figurewidth*\i},{-0.6\figureheight}) {};}

  \end{tikzpicture}
  \caption{Uncurated examples of reconstruction quality in $512{\times}512$ resolution with unseen images from the FFHQ test set (top row: inputs, bottom row: reconstructions).}
  \label{fig:reconstructions-ffhq}
\end{figure*}

\begin{figure*}[!t]
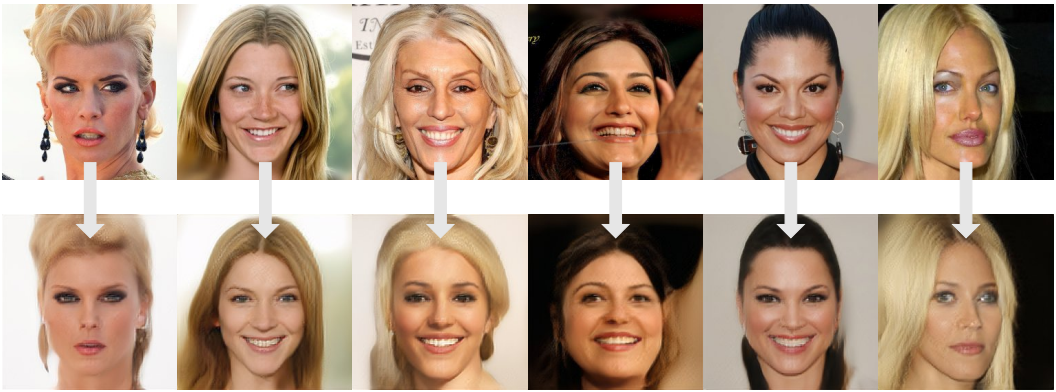

  \centering\footnotesize
  \setlength{\figurewidth}{.1667\textwidth}
  \setlength{\figureheight}{\figurewidth}
  \begin{tikzpicture}[inner sep=0]

  \tikzstyle{arrow} = [draw=black!10, single arrow, minimum height=10mm, minimum width=3mm, single arrow head extend=1mm, fill=black!10, anchor=center, rotate=-90, inner sep=2pt]

  \foreach \x [count=\i] in {0,...,5} {
     \node[] at ({\figurewidth*\i},{0*\figureheight}) {\includegraphics[width=\figurewidth]{./fig/reconstructions/celebahq/31877/\x_orig.jpg}};
     \node[] at ({\figurewidth*\i},{-1.2\figureheight}) {\includegraphics[width=\figurewidth]{./fig/reconstructions/celebahq/31877/\x_pine.jpg}};
     \node[arrow] at ({\figurewidth*\i},{-0.6\figureheight}) {};}

  \end{tikzpicture}
  \caption{Uncurated examples of reconstruction quality in $256{\times}256$ resolution with unseen images from the \CelebAHQ test set (top row: inputs, bottom row: reconstructions).}
  \label{fig:reconstructions-celeba}
\end{figure*}

We include examples of the reconstruction capabilities of the automodulator at $256{\times}256$ in for uncurated test set samples from the FFHQ and \CelebAHQ data sets. These examples are provided in \cref{fig:reconstructions-ffhq,fig:reconstructions-celeba}. 

\section{Style Mixing and Interpolation}
\label{sec:stylemixint}
\begin{figure*}[!t]
  \centering
  \resizebox{\textwidth}{!}{%
  \begin{tikzpicture}[inner sep=0]

    \newcommand{\figg}[2]{\includegraphics[width=1cm]{./fig/ffhq_interpolate/interpolations_6_32180002_1_#1x#2.jpg}}

    \foreach \i in {0,...,7} {
      \foreach \j in {0,...,7} {
        \node (\i-\j) [] at (\j,-\i) {\figg{\i}{\j}};
      }
    }

    \newcommand{\insquare}[3]{\node [minimum width=1.5cm,minimum height=1.5cm, rounded corners=3pt,path picture={\node at (path picture bounding box.center){\includegraphics[width=1.5cm]{#3.jpg}};}] at (#1,#2) {};};

    \insquare{-1.5}{0}{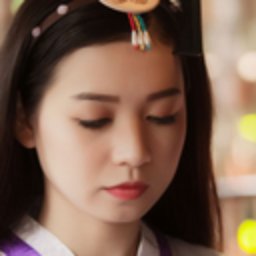}
    \insquare{8.5}{0}{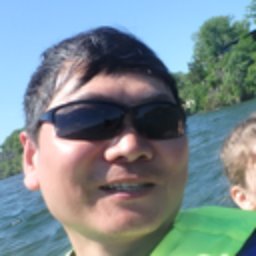}
    \insquare{-1.5}{-7}{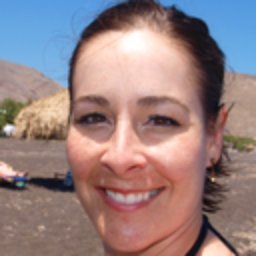}
    \insquare{8.5}{-7}{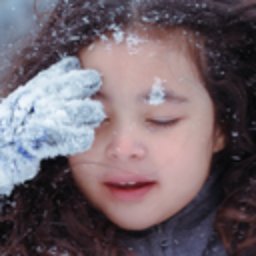}
    
  \end{tikzpicture}}
  \caption{Interpolation between random input images from FFHQ test set in $256{\times}256$ (originals in the corners) which the model has not seen during training. The model captures most of the salient features in the reconstructions and produces smooth interpolations at all points in the traversed space.}
  \label{fig:interpolation}

\end{figure*}

The well disentangled latent space allows for interpolations between encoded images. We show regular latent space interpolations between the reconstructions of new input images (\cref{fig:interpolation}).

As two more systematic style mixing examples, we include style mixing results based on both FFHQ and \LSUN Cars. The source images are unseen real test images, not self-generated images. In \cref{fig:face-mixing,fig:car-mixing} we show a matrix of cross-mixing either `coarse' (latent resolutions $4 {\times} 4$ -- $8 {\times} 8$) or `intermediate' ($16 {\times} 16$ -- $32 {\times} 32$) latent features. Mixing coarse features results in large-scale changes, such as pose, while the intermediate features drive finer details, such as color.

\begingroup
  \setlength{\figurewidth}{.13\textwidth}
  \setlength{\figureheight}{\figurewidth}

  \newcommand{\stylemix}[2]{%
  \begin{tikzpicture}[inner sep=0]

  \foreach \y [count=\j] in {0,2,11,9,5,1}
    \node[] at ({-.2*\figurewidth},{-\figureheight*\j}) {\includegraphics[width=\figurewidth]{#1/source-\y.jpg}};
  \foreach \x [count=\i] in {0,2,11,9,5,1}%
    \node[] at ({\figurewidth*\i},{.2*\figureheight}) {\includegraphics[width=\figurewidth]{#1/source-\x.jpg}};
  
  \foreach \y [count=\j] in {0,2,11,9,5,1} {
    \foreach \x [count=\i] in {0,2,11,9,5,1} {
      \ifnum \x>\y
        \node[] at ({\figurewidth*\i},{-\figureheight*\j}) {\includegraphics[width=\figurewidth]{#1/mix-\y_\x_#2.jpg}};
      \fi
      \ifnum \x<\y
        \node[] at ({\figurewidth*\i},{-\figureheight*\j}) {\includegraphics[width=\figurewidth]{#1/mix-\y_\x_#2.jpg}};
      \fi      
      \ifnum \x=\y
        \node[] at ({\figurewidth*\i},{-\figureheight*\j}) {\includegraphics[width=\figurewidth]{#1/mix-\x_\y_0--1.jpg}};
      \fi      
    }
  }
  \node at (3.5*\figurewidth,.9*\figureheight) {\bf Source B (real)};
  \node[rotate=90] at (-.9*\figurewidth,-3.5*\figureheight) {\bf Source A (real)};
    
  \end{tikzpicture}}

\begin{figure*}[!p]
  \centering
  \begin{subfigure}[b]{\textwidth}
    \centering
    \scalebox{1}{
      \stylemix{./fig/sysmix8/42420N2}{0-2}
    }
    \caption{Using `coarse' (latent resolutions $4 {\times} 4$ -- $8 {\times} 8$) latent features from B and the rest from A.}
    \label{fig:face-mixing-a}
  \end{subfigure}
  \caption{Style mixing of $512 {\times} 512$ FFHQ face images. The source images are unseen real test images, not self-generated images. The reconstructions of the input images are shown on the diagonal.}  

\end{figure*}%
\begin{figure*}[p]\ContinuedFloat
  \begin{subfigure}[b]{\textwidth}
    \centering
    \scalebox{1}{
      \stylemix{./fig/sysmix8/42420N2}{2-5}
    }
    \caption{Using the `intermediate' ($16 {\times} 16$ -- $64 {\times} 64$) latent features from B and the rest from A.}
    \label{fig:face-mixing-b}
  \end{subfigure}  
  \caption{Style mixing of $512 {\times} 512$ FFHQ face images. The source images are unseen real test images, not self-generated images. The reconstructions of the input images are shown on the diagonal.}
  \label{fig:face-mixing}
\end{figure*}

\endgroup

\begin{figure*}[!t]
  \centering
  \setlength{\figurewidth}{.075\textwidth}
  \setlength{\figureheight}{\figurewidth}

  \newcommand{\stylemix}[2]{%
  \begin{tikzpicture}[inner sep=0]

  \foreach \y [count=\j] in {0,1,2,3,4,5,6,7}
    \node[] at ({-.2*\figurewidth},{-\figureheight*\j}) {\includegraphics[width=\figurewidth]{#1/source-\y.jpg}};
  \foreach \x [count=\i] in {0,1,2,3,4,5,6,7}
    \node[] at ({\figurewidth*\i},{.2*\figureheight}) {\includegraphics[width=\figurewidth]{#1/source-\x.jpg}};
  
  \foreach \y [count=\j] in {0,1,2,3,4,5,6,7} {
    \foreach \x [count=\i] in {0,1,2,3,4,5,6,7} {
      \ifnum \x>\y
        \node[] at ({\figurewidth*\i},{-\figureheight*\j}) {\includegraphics[width=\figurewidth]{#1/mix_\x_\y_#2.jpg}};
      \fi
      \ifnum \x<\y
        \node[] at ({\figurewidth*\i},{-\figureheight*\j}) {\includegraphics[width=\figurewidth]{#1/mix_\x_\y_#2.jpg}};
      \fi      
      \ifnum \x=\y
        \node[] at ({\figurewidth*\i},{-\figureheight*\j}) {\includegraphics[width=\figurewidth]{#1/mix_\x_\y_0--1.jpg}};
      \fi      
    }
  }
  \node at (4.5*\figurewidth,.9*\figureheight) {\bf Source B (real)};
  \node[rotate=90] at (-.9*\figurewidth,-4.5*\figureheight) {\bf Source A (real)};
    
  \end{tikzpicture}}
  \begin{subfigure}[b]{\textwidth}
    \centering
    \scalebox{1}{
      \stylemix{./fig/sysmix8/31test}{0-2}
    }
    \caption{Using `coarse' (latent resolutions $4 {\times} 4$ -- $8 {\times} 8$) latent features from B and the rest from A. Most notably, the B cars drive the car pose.}
    \label{fig:car-mixing-a}
  \end{subfigure}
  \\[3em]
  \begin{subfigure}[b]{\textwidth}
    \centering
    \scalebox{1}{
      \stylemix{./fig/sysmix8/31test}{2-4}
    }
    \caption{Using the `intermediate' ($16 {\times} 16$ -- $32 {\times} 32$) latent features from B and the rest from A.}
    \label{fig:car-mixing-b}
  \end{subfigure}  
  \caption{Style mixing of $256 {\times} 256$ \LSUN Cars. The source images are unseen real test images, not self-generated images. The reconstructions of the input images are shown on the diagonal.}
  \label{fig:car-mixing}
\end{figure*}

\section{Comparison to GAN Inversion}
\label{app:comparison-to-gan-inv}
Although a GAN trained without an encoder cannot take inputs directly, it is possible to fit images into its latent space by training an encoder after regular GAN training, or by using a separate optimization process. One may wonder how well such image reconstruction would compare to our results here, and we will focus on a readily available method using the latter approach - optimization.

Specifically, we can find the latent codes for StyleGAN \citep{karras2018style} with an optimizer, leveraging VGG16 feature projections \citep{simonyan:2015,puzer:2019}. The optimization takes place in the large $18 {\times} 512$ latent W space, and the resulting latent codes are decoded back to $1024 {\times} 1024$ image space in the regular way by the GAN generator network. It should be noted that the latent space of our automodulator is more compact -- $1 {\times} 512$ -- and hence the two approaches are not directly comparable. However, according to \citet{image2stylegan}, the StyleGAN inversion does not work well if the corresponding original latent Z space of StyleGAN is used instead of the large W space.

Besides the higher dimensionality of the latent space, there are other issues that hamper straighforward comparison. First, the GAN inversion now hinges on a third (ad hoc) deep network, in addition to the GAN generator and discriminator. It is unclear whether inverting a model trained on one specific data set (faces) will work equally well with other data sets. Consider, \eg, the case of microscope imaging. Even though one could apply both the Automodulator and StyleGAN to learn such images in a straight-forward manner as long as they can be approached with convolutions, one is faced with a more complex question about which optimizer should now be chosen for StyleGAN inversion, given the potentially poorer performance of VGG16 features on such images. In any case, we now have a separate optimization problem to solve. This brings us to the second issue, the very slow convergence, which calls for evaluation as a function of optimization time. Third, the relationship of the projected latent coordinates of input images to their hypothetical optimal coordinates is an interesting open question, which we will tentatively address by evaluating the interpolated points between the projected latent coordinates.

First, for the convergence evaluation, we run the projector for a varying number of iterations, up to 200, or 68 seconds per image on average. We use the StyleGAN network pretrained on FFHQ, and compare to Automodulator also trained on FFHQ. We test the results on 1000 \CelebAHQ test images, on a single NVIDIA Titan V gpu. The script is based on the implementation of \citet{puzer:2019}. To measure the similarity of reconstructed images to the originals, we use the same LPIPS measure as before, with images cropped to $128 {\times} 128$ in the middle of the face region. Note that StyleGAN images are matched at $1024 {\times} 1024$ scale and then scaled down to $256 {\times} 256$ before the cropping. (Note: concurrently to the publishing of this version of the manuscript, an improved version of StyleGAN with possibly better projection capabilities has been released in \citet{2019arXiv191204958K}.)

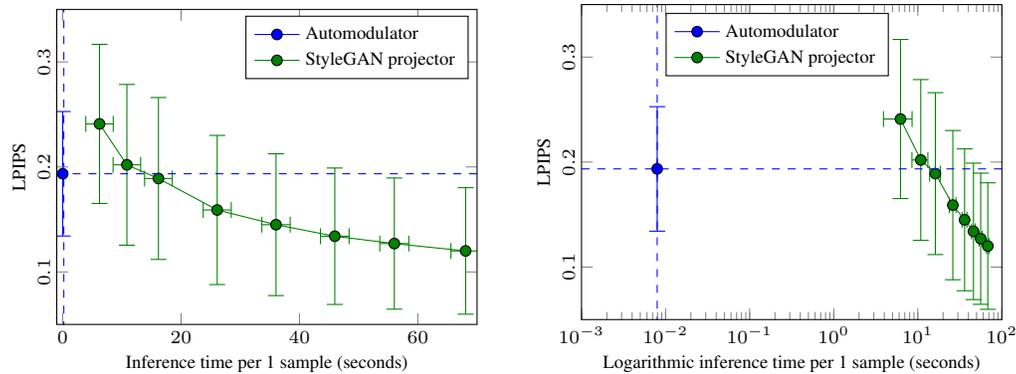
\begin{figure*}[!t]
  \begin{subfigure}[b]{.5\textwidth}
  \centering\scriptsize
  \setlength{\figurewidth}{.8\textwidth}
  \setlength{\figureheight}{.75\figurewidth}
  \pgfplotsset{yticklabel style={rotate=90}, ylabel near ticks, clip=true,scale only axis,axis on top,clip marker paths,legend style={row sep=0pt},xlabel near ticks,legend style={fill=white}}
\begin{tikzpicture}

\begin{axis}[
axis on top,
height=\figureheight,
legend cell align={left},
legend entries={{Automodulator},{StyleGAN projector}},
tick pos=both,
width=\figurewidth,
xlabel={Inference time per 1 sample (seconds)},
xmin=-1, xmax=70,
ylabel={LPIPS},
ymin=0.05, ymax=0.35
]
\addlegendimage{blue, mark=*, mark size=2, mark options={solid,draw=black}}
\addlegendimage{green!50.0!black, mark=*, mark size=2, mark options={solid,draw=black}}
\path [draw=blue] (axis cs:0.00790764649183123,0.1935)
--(axis cs:0.00797721822300662,0.1935);

\path [draw=blue] (axis cs:0.00794243235741892,0.1342207854092)
--(axis cs:0.00794243235741892,0.2527792145908);

\path [draw=green!50.0!black] (axis cs:3.90868932001272,0.241)
--(axis cs:8.54257497606722,0.241);

\path [draw=green!50.0!black] (axis cs:8.51667900958499,0.202)
--(axis cs:13.1734931607529,0.202);

\path [draw=green!50.0!black] (axis cs:13.8345696579301,0.189)
--(axis cs:18.5213089913298,0.189);

\path [draw=green!50.0!black] (axis cs:23.7380604360322,0.159)
--(axis cs:28.4727453665641,0.159);

\path [draw=green!50.0!black] (axis cs:33.6437190011083,0.145)
--(axis cs:38.4320894286097,0.145);

\path [draw=green!50.0!black] (axis cs:43.5472427618655,0.134)
--(axis cs:48.3900272169087,0.134);

\path [draw=green!50.0!black] (axis cs:53.5680917491614,0.127)
--(axis cs:58.4712768100235,0.127);

\path [draw=green!50.0!black] (axis cs:65.583537692533,0.12)
--(axis cs:70.5528791743718,0.12);

\path [draw=green!50.0!black] (axis cs:6.22563214803997,0.1652424294353)
--(axis cs:6.22563214803997,0.3167575705647);

\path [draw=green!50.0!black] (axis cs:10.8450860851689,0.1254459325671)
--(axis cs:10.8450860851689,0.2785540674329);

\path [draw=green!50.0!black] (axis cs:16.1779393246299,0.1118927195072)
--(axis cs:16.1779393246299,0.2661072804928);

\path [draw=green!50.0!black] (axis cs:26.1054029012981,0.0879608299732)
--(axis cs:26.1054029012981,0.2300391700268);

\path [draw=green!50.0!black] (axis cs:36.037904214859,0.0774670135975)
--(axis cs:36.037904214859,0.2125329864025);

\path [draw=green!50.0!black] (axis cs:45.9686349893871,0.0690407198071)
--(axis cs:45.9686349893871,0.1989592801929);

\path [draw=green!50.0!black] (axis cs:56.0196842795924,0.0645259764493)
--(axis cs:56.0196842795924,0.1894740235507);

\path [draw=green!50.0!black] (axis cs:68.0682084334524,0.0597588893771)
--(axis cs:68.0682084334524,0.1802411106229);

\addplot [blue, mark=|, mark size=3, mark options={solid}, only marks]
table [row sep=\\]{%
0.00790764649183123	0.1935 \\
};
\addplot [blue, mark=|, mark size=3, mark options={solid}, only marks]
table [row sep=\\]{%
0.00797721822300662	0.1935 \\
};
\addplot [blue, mark=-, mark size=3, mark options={solid}, only marks]
table [row sep=\\]{%
0.00794243235741892	0.1342207854092 \\
};
\addplot [blue, mark=-, mark size=3, mark options={solid}, only marks]
table [row sep=\\]{%
0.00794243235741892	0.2527792145908 \\
};
\addplot [blue, dashed, mark=*, mark size=2, mark options={solid,draw=black}]
table [row sep=\\]{%
0.00794243235741892	0.1935 \\
};
\addplot [green!50.0!black, mark=|, mark size=3, mark options={solid}, only marks]
table [row sep=\\]{%
3.90868932001272	0.241 \\
8.51667900958499	0.202 \\
13.8345696579301	0.189 \\
23.7380604360322	0.159 \\
33.6437190011083	0.145 \\
43.5472427618655	0.134 \\
53.5680917491614	0.127 \\
65.583537692533	0.12 \\
};
\addplot [green!50.0!black, mark=|, mark size=3, mark options={solid}, only marks]
table [row sep=\\]{%
8.54257497606722	0.241 \\
13.1734931607529	0.202 \\
18.5213089913298	0.189 \\
28.4727453665641	0.159 \\
38.4320894286097	0.145 \\
48.3900272169087	0.134 \\
58.4712768100235	0.127 \\
70.5528791743718	0.12 \\
};
\addplot [green!50.0!black, mark=-, mark size=3, mark options={solid}, only marks]
table [row sep=\\]{%
6.22563214803997	0.1652424294353 \\
10.8450860851689	0.1254459325671 \\
16.1779393246299	0.1118927195072 \\
26.1054029012981	0.0879608299732 \\
36.037904214859	0.0774670135975 \\
45.9686349893871	0.0690407198071 \\
56.0196842795924	0.0645259764493 \\
68.0682084334524	0.0597588893771 \\
};
\addplot [green!50.0!black, mark=-, mark size=3, mark options={solid}, only marks]
table [row sep=\\]{%
6.22563214803997	0.3167575705647 \\
10.8450860851689	0.2785540674329 \\
16.1779393246299	0.2661072804928 \\
26.1054029012981	0.2300391700268 \\
36.037904214859	0.2125329864025 \\
45.9686349893871	0.1989592801929 \\
56.0196842795924	0.1894740235507 \\
68.0682084334524	0.1802411106229 \\
};
\addplot [green!50.0!black, mark=*, mark size=2, mark options={solid,draw=black}]
table [row sep=\\]{%
6.22563214803997	0.241 \\
10.8450860851689	0.202 \\
16.1779393246299	0.189 \\
26.1054029012981	0.159 \\
36.037904214859	0.145 \\
45.9686349893871	0.134 \\
56.0196842795924	0.127 \\
68.0682084334524	0.12 \\
};
\addplot [blue, dashed, forget plot]
table [row sep=\\]{%
	0.1935 0 \\
	0.1935 1 \\
};
\addplot[blue, dashed, samples=5, domain=-1:70] {0.1935};
\end{axis}

\end{tikzpicture}
  \end{subfigure}
  \hspace{\fill}
  \begin{subfigure}[b]{.5\textwidth}
    \centering\scriptsize
  \setlength{\figurewidth}{.8\textwidth}
  \setlength{\figureheight}{.75\figurewidth}
  \pgfplotsset{yticklabel style={rotate=90}, ylabel near ticks, clip=true,scale only axis,axis on top,clip marker paths,legend style={row sep=0pt},xlabel near ticks,legend style={fill=white}}
\begin{tikzpicture}

\begin{axis}[
axis on top,
height=\figureheight,
legend cell align=left,
legend style={at={(0.20,0.873)},anchor=west},
legend entries={{Automodulator},{StyleGAN projector}},
tick pos=both,
width=\figurewidth,
xlabel={Logarithmic inference time per 1 sample (seconds)},
xmin=0.001, xmax=100,
xmode=log,
ylabel={LPIPS},
ymin=0.05, ymax=0.35
]
\addlegendimage{blue, mark=*, mark size=2, mark options={solid,draw=black}}
\addlegendimage{green!50.0!black, mark=*, mark size=2, mark options={solid,draw=black}}
\path [draw=blue] (axis cs:0.00790764649183123,0.1935)
--(axis cs:0.00797721822300662,0.1935);

\path [draw=blue] (axis cs:0.00794243235741892,0.1342207854092)
--(axis cs:0.00794243235741892,0.2527792145908);

\path [draw=green!50.0!black] (axis cs:3.90868932001272,0.241)
--(axis cs:8.54257497606722,0.241);

\path [draw=green!50.0!black] (axis cs:8.51667900958499,0.202)
--(axis cs:13.1734931607529,0.202);

\path [draw=green!50.0!black] (axis cs:13.8345696579301,0.189)
--(axis cs:18.5213089913298,0.189);

\path [draw=green!50.0!black] (axis cs:23.7380604360322,0.159)
--(axis cs:28.4727453665641,0.159);

\path [draw=green!50.0!black] (axis cs:33.6437190011083,0.145)
--(axis cs:38.4320894286097,0.145);

\path [draw=green!50.0!black] (axis cs:43.5472427618655,0.134)
--(axis cs:48.3900272169087,0.134);

\path [draw=green!50.0!black] (axis cs:53.5680917491614,0.127)
--(axis cs:58.4712768100235,0.127);

\path [draw=green!50.0!black] (axis cs:65.583537692533,0.12)
--(axis cs:70.5528791743718,0.12);

\path [draw=green!50.0!black] (axis cs:6.22563214803997,0.1652424294353)
--(axis cs:6.22563214803997,0.3167575705647);

\path [draw=green!50.0!black] (axis cs:10.8450860851689,0.1254459325671)
--(axis cs:10.8450860851689,0.2785540674329);

\path [draw=green!50.0!black] (axis cs:16.1779393246299,0.1118927195072)
--(axis cs:16.1779393246299,0.2661072804928);

\path [draw=green!50.0!black] (axis cs:26.1054029012981,0.0879608299732)
--(axis cs:26.1054029012981,0.2300391700268);

\path [draw=green!50.0!black] (axis cs:36.037904214859,0.0774670135975)
--(axis cs:36.037904214859,0.2125329864025);

\path [draw=green!50.0!black] (axis cs:45.9686349893871,0.0690407198071)
--(axis cs:45.9686349893871,0.1989592801929);

\path [draw=green!50.0!black] (axis cs:56.0196842795924,0.0645259764493)
--(axis cs:56.0196842795924,0.1894740235507);

\path [draw=green!50.0!black] (axis cs:68.0682084334524,0.0597588893771)
--(axis cs:68.0682084334524,0.1802411106229);

\addplot [blue, mark=|, mark size=3, mark options={solid}, only marks]
table [row sep=\\]{%
0.00790764649183123	0.1935 \\
};
\addplot [blue, mark=|, mark size=3, mark options={solid}, only marks]
table [row sep=\\]{%
0.00797721822300662	0.1935 \\
};
\addplot [blue, mark=-, mark size=3, mark options={solid}, only marks]
table [row sep=\\]{%
0.00794243235741892	0.1342207854092 \\
};
\addplot [blue, mark=-, mark size=3, mark options={solid}, only marks]
table [row sep=\\]{%
0.00794243235741892	0.2527792145908 \\
};
\addplot [blue, dashed, mark=*, mark size=2, mark options={solid,draw=black}]
table [row sep=\\]{%
0.00794243235741892	0.1935 \\
};
\addplot [green!50.0!black, mark=|, mark size=3, mark options={solid}, only marks]
table [row sep=\\]{%
3.90868932001272	0.241 \\
8.51667900958499	0.202 \\
13.8345696579301	0.189 \\
23.7380604360322	0.159 \\
33.6437190011083	0.145 \\
43.5472427618655	0.134 \\
53.5680917491614	0.127 \\
65.583537692533	0.12 \\
};
\addplot [green!50.0!black, mark=|, mark size=3, mark options={solid}, only marks]
table [row sep=\\]{%
8.54257497606722	0.241 \\
13.1734931607529	0.202 \\
18.5213089913298	0.189 \\
28.4727453665641	0.159 \\
38.4320894286097	0.145 \\
48.3900272169087	0.134 \\
58.4712768100235	0.127 \\
70.5528791743718	0.12 \\
};
\addplot [green!50.0!black, mark=-, mark size=3, mark options={solid}, only marks]
table [row sep=\\]{%
6.22563214803997	0.1652424294353 \\
10.8450860851689	0.1254459325671 \\
16.1779393246299	0.1118927195072 \\
26.1054029012981	0.0879608299732 \\
36.037904214859	0.0774670135975 \\
45.9686349893871	0.0690407198071 \\
56.0196842795924	0.0645259764493 \\
68.0682084334524	0.0597588893771 \\
};
\addplot [green!50.0!black, mark=-, mark size=3, mark options={solid}, only marks]
table [row sep=\\]{%
6.22563214803997	0.3167575705647 \\
10.8450860851689	0.2785540674329 \\
16.1779393246299	0.2661072804928 \\
26.1054029012981	0.2300391700268 \\
36.037904214859	0.2125329864025 \\
45.9686349893871	0.1989592801929 \\
56.0196842795924	0.1894740235507 \\
68.0682084334524	0.1802411106229 \\
};
\addplot [green!50.0!black, mark=*, mark size=2, mark options={solid,draw=black}]
table [row sep=\\]{%
6.22563214803997	0.241 \\
10.8450860851689	0.202 \\
16.1779393246299	0.189 \\
26.1054029012981	0.159 \\
36.037904214859	0.145 \\
45.9686349893871	0.134 \\
56.0196842795924	0.127 \\
68.0682084334524	0.12 \\
};
\addplot [blue, dashed, forget plot]
table [row sep=\\]{%
0.00794243235741892	0 \\
0.00794243235741892	1 \\
};
\addplot[blue, dashed, samples=5, domain=0.001:100] {0.1935};
\end{axis}

\end{tikzpicture}

  \end{subfigure}

  \caption{Comparison of LPIPS similarity of image reconstructions in Automodulator (ours) and StyleGAN (left: linear scale, right: log xscale). The error bars indicate standard deviations across evaluation runs. We show that optimization to StyleGAN latent space takes over 3 orders of magnitude more time to match the Automodulator (up to 16~s), but will continue improving thereafter. Here, the Automodulator encodes 1 image in 0.008~s, with the LPIPS shown as the constant horizontal line.}
  \label{fig:stylegan-comp}
\end{figure*}

The results (\cref{fig:stylegan-comp}) are calculated for various stages of the optimization for StyleGAN, against the single direct encoding result of Automodulator. The Automodulator uses no separate optimization processes. The results indicate that on this hardware setup, it takes over 10~seconds for the optimization process to reconstruct a single image to match the LPIPS of Automodulator, whereas a single images takes only 0.0079 s for the Automodulator encoder inference (or 0.1271~s for a batch of 16~images). The performance difference is almost at the considerable four orders of magnitude. StyleGAN projection does, however, continue improving to produce significantly better LPIPS, given more optimization. Moreover, to get the best results, we used the $1024 {\times} 1024$ resolution, which makes the optimization somewhat slower, and has not yet been matched by Automodulator. However, it is clear that a performance difference of  $10000{\times}$ limits the use cases of the GAN projection approach. For instance, in cases where the projected latent codes can be complemented by fast inference, such as \citet{hou2019}, the optimization speed is not limiting.

Second, in order to evaluate the properties of the projected latent coordinates, we again projected 1000 \CelebAHQ test images into the (FFHQ-trained) StyleGAN latent space and then sampled 10000 random points linearly interpolated between them (in the extended W latent space), with each point converted back into a generated image. For comparison, the similar procedure was done for the (FFHQ-trained) Automodulator, using the built-in encoder. We then evaluated the quality and diversity of the results in terms of FID, measured against 10000 \CelebAHQ training set images. Although such a measure is not ideal when one has only used 1000 (test set) images to begin with, it can be reasonably justified on the basis of the fact that, due to combinatorial explosion, interpolations should cover a relatively diverse set of images that goes far beyond the original images. The results of this experiment yielded FID of $52.88 \pm 0.71$ for StyleGAN and FID of $48.83 \pm 0.95$ for Automodulator. Hence, in this specific measure, StyleGAN performed slightly worse (despite the fact that StyleGAN projection still used nearly 10000x more time).

\begin{figure*}[!t]
  \begin{subfigure}[b]{.5\textwidth}
  \centering\scriptsize
  \setlength{\figurewidth}{.8\textwidth}
  \setlength{\figureheight}{.75\figurewidth}
  \pgfplotsset{yticklabel style={rotate=90}, ylabel near ticks, clip=true,scale only axis,axis on top,clip marker paths,legend style={row sep=0pt},xlabel near ticks,legend style={fill=white}}
\begin{tikzpicture}

\definecolor{color0}{rgb}{0.12156862745098,0.466666666666667,0.705882352941177}
\definecolor{color1}{rgb}{1,0.498039215686275,0.0549019607843137}

\begin{axis}[
height=\figureheight,
width=\figurewidth,
legend cell align={left},
legend style={fill opacity=0.8, draw opacity=1, text opacity=1, draw=white!80!black},
tick align=outside,
tick pos=left,
x grid style={white!69.0196078431373!black},
xlabel={Inference time per 1 input image (seconds)},
xmin=-1, xmax=70,
xtick style={color=black},
y grid style={white!69.0196078431373!black},
ylabel={FID (10k)},
ymin=37.0383124045463, ymax=67.5870595851764,
ytick style={color=black}
]
\path [draw=blue, semithick]
(axis cs:0.00790764649183123,52.1917995250683)
--(axis cs:0.00797721822300662,52.1917995250683);

\path [draw=blue, semithick]
(axis cs:0.00794243235741892,51.9675219114737)
--(axis cs:0.00794243235741892,52.4160771386629);

\path [draw=green!50.0!black, semithick]
(axis cs:3.305686406,65.9899521174613)
--(axis cs:3.505686406,65.9899521174613);

\path [draw=green!50.0!black, semithick]
(axis cs:8.314216015,47.6958129152746)
--(axis cs:8.714216015,47.6958129152746);

\path [draw=green!50.0!black, semithick]
(axis cs:16.72843203,38.7546144619782)
--(axis cs:17.32843203,38.7546144619782);

\path [draw=green!50.0!black, semithick]
(axis cs:33.65686406,43.3108189778656)
--(axis cs:34.45686406,43.3108189778656);

\path [draw=green!50.0!black, semithick]
(axis cs:67.61372812,45.4980318314881)
--(axis cs:68.61372812,45.4980318314881);

\path [draw=green!50.0!black, semithick]
(axis cs:3.405686406,65.7814240670476)
--(axis cs:3.405686406,66.198480167875);

\path [draw=green!50.0!black, semithick]
(axis cs:8.514216015,47.511196882085)
--(axis cs:8.514216015,47.8804289484642);

\path [draw=green!50.0!black, semithick]
(axis cs:17.02843203,38.4268918218477)
--(axis cs:17.02843203,39.0823371021088);

\path [draw=green!50.0!black, semithick]
(axis cs:34.05686406,43.0483593293836)
--(axis cs:34.05686406,43.5732786263475);

\path [draw=green!50.0!black, semithick]
(axis cs:68.11372812,45.1176350730273)
--(axis cs:68.11372812,45.8784285899489);

\addplot [semithick, blue, dashed, forget plot]
table {%
0.0079424323574207 37.0383124045463
0.0079424323574207 67.5870595851764
};
\addplot [semithick, blue, dashed, forget plot]
table {%
-1 52.1917995250683
70 52.1917995250683
};
\addplot [semithick, blue, dashed, mark=*, mark size=2, mark options={solid,draw=black}]
table {%
0.00794243235741892 52.1917995250683
};
\addlegendentry{Automodulator}
\addplot [semithick, green!50.0!black, mark=*, mark size=2, mark options={solid,draw=black}]
table {%
3.405686406 65.9899521174613
8.514216015 47.6958129152746
17.02843203 38.7546144619782
34.05686406 43.3108189778656
68.11372812 45.4980318314881
};
\addlegendentry{StyleGAN projector}
\end{axis}

\end{tikzpicture}
  \end{subfigure}
  \hspace{\fill}
  \begin{subfigure}[b]{.5\textwidth}
    \centering\scriptsize
  \setlength{\figurewidth}{.8\textwidth}
  \setlength{\figureheight}{.75\figurewidth}
  \pgfplotsset{yticklabel style={rotate=90}, ylabel near ticks, clip=true,scale only axis,axis on top,clip marker paths,legend style={row sep=0pt},xlabel near ticks,legend style={fill=white}}
\begin{tikzpicture}

\definecolor{color0}{rgb}{0.12156862745098,0.466666666666667,0.705882352941177}
\definecolor{color1}{rgb}{1,0.498039215686275,0.0549019607843137}

\begin{axis}[
height=\figureheight,
width=\figurewidth,
legend cell align={left},
legend style={fill opacity=0.8, draw opacity=1, text opacity=1, draw=white!80!black,at={(0.10,0.873)},anchor=west},
log basis x={10},
tick align=outside,
tick pos=left,
x grid style={white!69.0196078431373!black},
xlabel={Inference time per 1 input image (seconds)},
xmin=0.00502491890843758, xmax=107.976490038181,
xmode=log,
xtick style={color=black},
y grid style={white!69.0196078431373!black},
ylabel={FID (10k)},
ymin=37.0383124045463, ymax=67.5870595851764,
ytick style={color=black}
]
\path [draw=blue, semithick]
(axis cs:0.00790764649183123,52.1917995250683)
--(axis cs:0.00797721822300662,52.1917995250683);

\path [draw=blue, semithick]
(axis cs:0.00794243235741892,51.9675219114737)
--(axis cs:0.00794243235741892,52.4160771386629);

\path [draw=green!50.0!black, semithick]
(axis cs:3.305686406,65.9899521174613)
--(axis cs:3.505686406,65.9899521174613);

\path [draw=green!50.0!black, semithick]
(axis cs:8.314216015,47.6958129152746)
--(axis cs:8.714216015,47.6958129152746);

\path [draw=green!50.0!black, semithick]
(axis cs:16.72843203,38.7546144619782)
--(axis cs:17.32843203,38.7546144619782);

\path [draw=green!50.0!black, semithick]
(axis cs:33.65686406,43.3108189778656)
--(axis cs:34.45686406,43.3108189778656);

\path [draw=green!50.0!black, semithick]
(axis cs:67.61372812,45.4980318314881)
--(axis cs:68.61372812,45.4980318314881);

\path [draw=green!50.0!black, semithick]
(axis cs:3.405686406,65.7814240670476)
--(axis cs:3.405686406,66.198480167875);

\path [draw=green!50.0!black, semithick]
(axis cs:8.514216015,47.511196882085)
--(axis cs:8.514216015,47.8804289484642);

\path [draw=green!50.0!black, semithick]
(axis cs:17.02843203,38.4268918218477)
--(axis cs:17.02843203,39.0823371021088);

\path [draw=green!50.0!black, semithick]
(axis cs:34.05686406,43.0483593293836)
--(axis cs:34.05686406,43.5732786263475);

\path [draw=green!50.0!black, semithick]
(axis cs:68.11372812,45.1176350730273)
--(axis cs:68.11372812,45.8784285899489);

\addplot [semithick, blue, dashed, forget plot]
table {%
0.00794243235741892 37.0383124045463
0.00794243235741892 67.5870595851764
};
\addplot [semithick, blue, dashed, forget plot]
table {%
0.00502491890843757 52.1917995250683
107.976490038181 52.1917995250683
};
\addplot [semithick, blue, dashed, mark=*, mark size=2, mark options={solid,draw=black}]
table {%
0.00794243235741892 52.1917995250683
};

\addlegendentry{Automodulator}
\addplot [semithick, green!50.0!black, mark=*, mark size=2, mark options={solid,draw=black}]
table {%
3.405686406 65.9899521174613
8.514216015 47.6958129152746
17.02843203 38.7546144619782
34.05686406 43.3108189778656
68.11372812 45.4980318314881
};
\addlegendentry{StyleGAN projector}
\end{axis}

\end{tikzpicture}

  \end{subfigure}

  \caption{FID comparison of the results of images produced by mixing two reconstructions in Automodulator (ours) and StyleGAN (left: linear scale, right: log xscale), based on three random combination runs. Standard deviations (not visualized) are 0.38 at maximum for StyleGAN and 0.22 for Automodulator. The optimization to StyleGAN latent space takes about 3 orders of magnitude more time to match the Automodulator, continues to improve thereafter, but further optimization of single images leads to worse FID of their mixtures. Here, the Automodulator encodes 1 input image in 0.008~s, with the FID shown as the constant horizontal line. }
  \label{fig:stylegan-fidcomp}
\end{figure*}
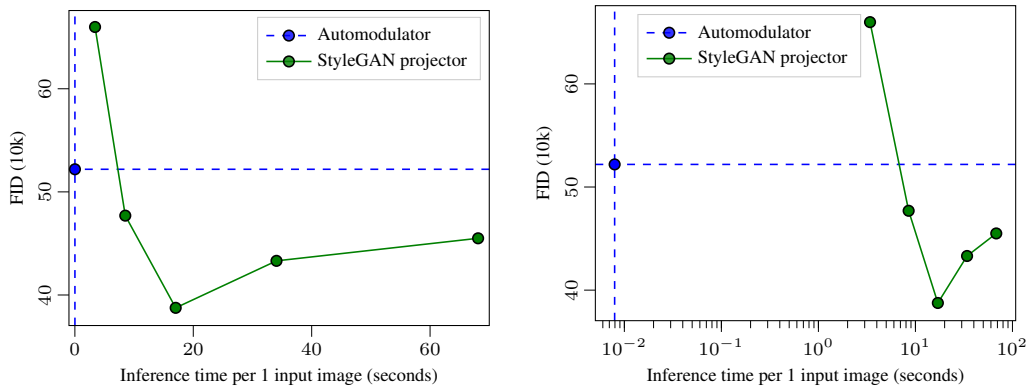

Third, in order to evaluate the {\textit{mixture}} properties of the projected latent coordinates, we once again projected 1000 \CelebAHQ test images into the (FFHQ-trained) StyleGAN latent space, but now take 10000 random pairs of those encodings, and mix each pair in StyleGAN so that the first code drives the first two layers of the decoder, while the second code drives the rest. We then look at the FID against 10000 \CelebAHQ training set images. We run this for a varying number of iterations (i.e. varying amounts of optimization time) for StyleGAN, and for a direct encoding-and-mixing result of Automodulator. The result (\cref{fig:stylegan-fidcomp}) indicates that the initial StyleGAN projections are inferior to the Automodulator results, but then improve with a larger iteration budget, reaching $26.5\%$ lower FID, but thereafter begin to deteriorate again. Our hypothesis is that by increasing the number of iterations, one finds a StyleGAN latent code that produces a better local projection fit than the earlier iterations (corresponding to a better LPIPS)but resides in a more pathological neighborhood, yielding worse mixing results when combined with another similarly projected latent. More research is needed to investigate this.

Although more research is called for, the FID results suggest that only a fraction of the fidelity and diversity of StyleGAN random samples is retained during projection. More subtle evaluation methods, and \eg, the effect of layer noise, are a topic for future research. For an additional comparison, one could also run similar optimization in Automodulator latent space.

\section{Conditional Sampling}
\label{sec:condsamp}
\begin{figure*}[!h]
  \centering\scriptsize
  \setlength{\figurewidth}{.11\textwidth}
  \setlength{\figureheight}{\figurewidth}
  \begin{tikzpicture}[inner sep=0]

  \node[] at ({0.9*\figurewidth},{-1.0*\figureheight*0}) {\includegraphics[width=\figurewidth]{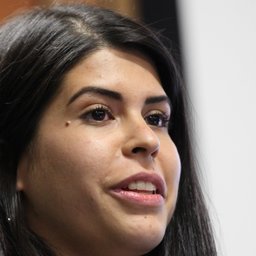}};
  \node[] at ({0.9*\figurewidth},{-1.0*\figureheight*1}) {\includegraphics[width=\figurewidth]{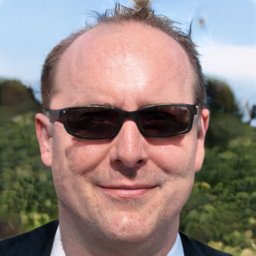}};
  \node[] at ({0.9*\figurewidth},{-1*\figureheight*2-.25*\figureheight}) {\includegraphics[width=\figurewidth]{./fig/conditional_sampling/38B/source-A-7.jpg}};
  \node[] at ({0.9*\figurewidth},{-1*\figureheight*3-.25*\figureheight}) {\includegraphics[width=\figurewidth]{./fig/conditional_sampling/38/source-A-10.jpg}};

  \foreach \x [count=\i] in {0,1,2,3,4,5,6,7} {
    \node[] at ({\figurewidth*\i+1*\figurewidth},{0*\figureheight}) {\includegraphics[width=\figurewidth]{./fig/conditional_sampling/38B/mix-7-\x_0-2.jpg}};
    \node[] at ({\figurewidth*\i+1*\figurewidth},{-1\figureheight}) {\includegraphics[width=\figurewidth]{./fig/conditional_sampling/38/mix-10-\x_0-2.jpg}};
    \node[] at ({\figurewidth*\i+1*\figurewidth},{-2\figureheight-.25*\figureheight}) {\includegraphics[width=\figurewidth]{./fig/conditional_sampling/38B/mix-7-\x_2-4.jpg}};
    \node[] at ({\figurewidth*\i+1*\figurewidth},{-3\figureheight-.25*\figureheight}) {\includegraphics[width=\figurewidth]{./fig/conditional_sampling/38/mix-10-\x_2-4.jpg}};}

  \node[] at (0.9*\figurewidth,.6*\figurewidth) {\bf Input};
  \node[align=left,text width=5*\figurewidth] at (4.5*\figurewidth,.6*\figurewidth) {\bf Samples with coarse features from input};
  \node[] at (0.9*\figurewidth,-1.65*\figurewidth) {\bf Input};
  \node[align=left,text width=5*\figurewidth] at (4.5*\figurewidth,-1.65*\figurewidth) {\bf Samples with intermediate features from input};

  \tikzstyle{arrow} = [draw=black!50, single arrow, minimum height=7mm, minimum width=2mm, single arrow head extend=1mm, fill=black!10, anchor=center, rotate=0, inner sep=2pt]

  \foreach \y [count=\j] in {0,1}
    \node[arrow] at ({1.45\figurewidth},{-\figureheight*\y}) {};
  \foreach \y [count=\j] in {2,3}
    \node[arrow] at ({1.45\figurewidth},{-\figureheight*\y-.25*\figureheight}) {};
        
  \end{tikzpicture}
  \caption{Conditional sampling of $256 {\times} 256$ random face images based on `coarse' (latent resolutions $4 {\times} 4$ -- $8 {\times} 8$) and `intermediate' ($16 {\times} 16$ -- $32 {\times} 32$) latent features of the fixed unseen test input. The input image controls the coarse features (such as head shape, pose, gender) on the top and more fine features (expressions, accessories, eyebrows) on the bottom.}
  \label{fig:conditional-sampling}
\end{figure*}

The automodulator directly allows for conditional sampling in the sense of fixing a latent encoding $\vz_A$, but allowing some of the modulations come from a random encoding $\vz_B \sim \mathrm{N}(\vzero,\MI)$. In \cref{fig:conditional-sampling}, we show conditional sampling of $128 {\times} 128$ random face images based on `coarse' (latent resolutions $4 {\times} 4$ -- $8 {\times} 8$) and `intermediate' ($16 {\times} 16$ -- $32 {\times} 32$) latent features of the fixed input. The input image controls the coarse features (such as head shape, pose, gender) on the top and more fine features (expressions, accessories, eyebrows) on the bottom.

\end{document}